\documentclass{ieeeaccess}
\usepackage{cite}
\usepackage{amsmath,amssymb,amsfonts}
\usepackage{graphicx}
\usepackage{textcomp}
\usepackage{csquotes}
\usepackage{multicol}
\usepackage[hyphens]{url}
\usepackage[export]{adjustbox}
\usepackage{hhline,colortbl}
\usepackage{algpseudocode}

\definecolor{grannysmithapple}{rgb}{0.66, 0.89, 0.63}
\definecolor{lightsalmonpink}{rgb}{1.0, 0.6, 0.6}

\usepackage{placeins}
\usepackage{algorithm}
\usepackage{multirow}
\usepackage{array}
\usepackage{bibentry}

\usepackage{bm}
\makeatletter
\AtBeginDocument{\DeclareMathVersion{bold}
\SetSymbolFont{operators}{bold}{T1}{times}{b}{n}
\SetSymbolFont{NewLetters}{bold}{T1}{times}{b}{it}
\SetMathAlphabet{\mathrm}{bold}{T1}{times}{b}{n}
\SetMathAlphabet{\mathit}{bold}{T1}{times}{b}{it}
\SetMathAlphabet{\mathbf}{bold}{T1}{times}{b}{n}
\SetMathAlphabet{\mathtt}{bold}{OT1}{pcr}{b}{n}
\SetSymbolFont{symbols}{bold}{OMS}{cmsy}{b}{n}
\renewcommand\boldmath{\@nomath\boldmath\mathversion{bold}}}
\makeatother

\def\BibTeX{{\rm B\kern-.05em{\sc i\kern-.025em b}\kern-.08em
    T\kern-.1667em\lower.7ex\hbox{E}\kern-.125emX}}

\begin{document}
\doi{10.48550/arXiv.2506.17001}

\title{PersonalAI 2.0: Enhancing knowledge graph traversal/retrieval with planning mechanism for Personalized LLM Agents}
\author{\uppercase{Mikhail Menschikov}\authorrefmark{1}, \uppercase{Matvey Iskornev}\authorrefmark{1}, Alexander Kharitonov\authorrefmark{2}, Alina Bogdanova\authorrefmark{3}, Mikhail Belkin\authorrefmark{4}, Ekaterina Lisitsyna\authorrefmark{4}, Artyom Sosedka\authorrefmark{4,5}, Victoria Dochkina\authorrefmark{6}, Ruslan Kostoev\authorrefmark{6}, Ilia Perepechkin\authorrefmark{6} and Evgeny Burnaev\authorrefmark{1,7}}

\address[1]{Skoltech, Moscow, Russia}
\address[2]{SberAI, Moscow, Russia}
\address[3]{Huawei, Moscow, Russia}
\address[4]{Sber, Moscow, Russia}
\address[5]{National University of Science and Technology MISIS, Moscow, Russia}
\address[6]{Public joint stock company "Sberbank of Russia" , Moscow, Russia}
\address[7]{AIRI, Moscow, Russia}


\corresp{Corresponding author: Mikhail Menschikov (e-mail: m.menschikov@ skoltech.ru).}

\tfootnote{The work was supported by the grant for research centers in the field of AI provided by the Ministry of Economic Development of the Russian Federation in accordance with the agreement 000000C313925P4F0002 and the agreement with Skoltech №139-10-2025-033}

\begin{abstract}
We introduce PersonalAI 2.0 (PAI-2), a novel framework designed to enhance LLM-based systems through integration of external knowledge graphs (KGs). The proposed approach addresses key limitations of existing Graph Retrieval-Augmented Generation (GraphRAG) methods by incorporating a dynamic, multistage pipeline. The central point of the PAI-2 design is its ability to perform adaptive, iterative information search, guided by extracted entities, matched graph vertices and generated clue-queries. Conducted evaluation on five benchmarks (Natural Questions, TriviaQA, HotpotQA, 2WikiMultihopQA and MuSiQue) demonstrates an improvement in the factual correctness of generated answers compared to baseline methods (LightRAG, RAPTOR, HippoRAG 2 and PAI-1). PAI-2 achieves a 9 percentage points (pp) average gain by LLM-as-a-Judge on 2WikiMultihopQA and MuSiQue benchmarks, and attains comparable accuracy to HippoRAG 2 on the TriviaQA and HotpotQA benchmarks with Qwen2.5 7B and multilingual-e5-large reflecting its effectiveness in reducing hallucination rates and increasing recall. We show that enabled search plan enhancement mechanism gain in average a 18 pp boost compared to disabled one by LLM-as-a-Judge across five benchmarks. In addition, an ablation study reveals that PAI-2 achieves superior result on the MINE-1 benchmark, obtaining an 89\% information-retention score with Qwen2.5 7B and multilingual-e5-large. Collectively, these findings underscore the potential of PAI-2 to serve as a reusable component for personalized AI applications, which require context-aware knowledge-representation and reasoning capabilities. The source code of PAI-2 is available at the following link: \url{https://github.com/Dzigen/PersonalAI}.
\end{abstract}


\titlepgskip=-21pt

\maketitle

\section{Introduction}

Large Language Models (LLMs) have revolutionized the field of artificial intelligence (AI), providing powerful tools for automated reasoning and conversational interactions~\cite{yang2025qwen3technicalreport,deepseekai2025deepseekv3technicalreport,5team2025glm45agenticreasoningcoding}. Their strengths lie in generative fluency and contextual understanding. However, these models face fundamental challenges when dealing with fact-rich domains, where knowledge consistency, scalability, and groundedness are crucial. Integration of external knowledge graphs (KGs) into LLM-driven systems offers a promising opportunity to bridge the gaps between reasoning and factuality~\cite{chepurova-etal-2026-wikontic,bai2025autoschemakgautonomousknowledgegraph,10.1145/3746027.3755628}. Yet, the complexity of scaling KG-based methods for closed-domain QA tasks and maintaining high retrieval recall remains a bottleneck.

\begin{figure*}[th!]
	\centering
	\includegraphics[width=.85\textwidth]{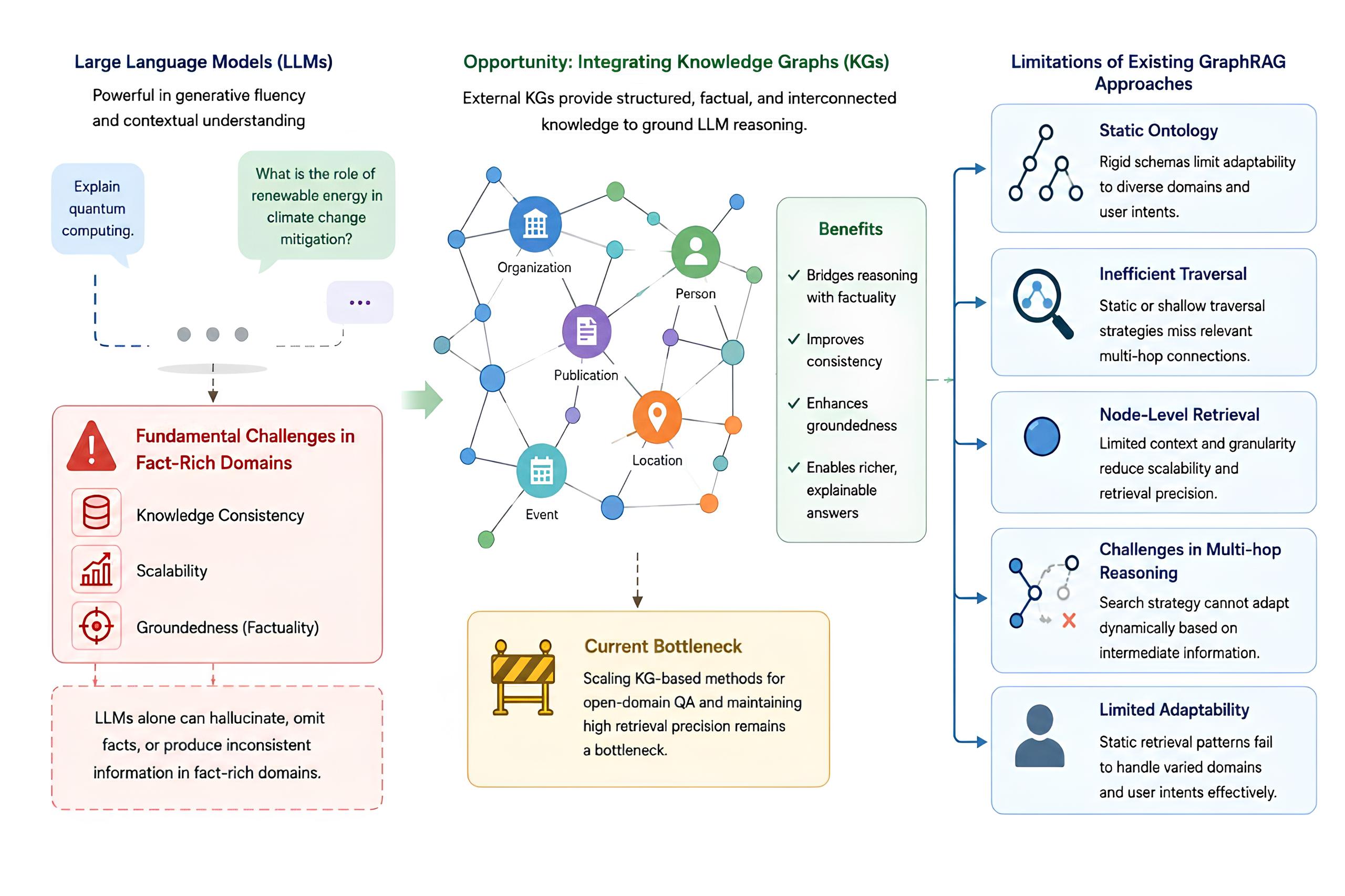}
	\caption{Illustration of existing GraphRAG challenges and limitations}
	\label{fig:problem visualisation}
\end{figure*}

Graph-based Retrieval-Augmented Generation (GraphRAG)~\cite{Gao2023RetrievalAugmentedGF, 10.1145/3777378} frameworks have gained prominence by augmenting prompts with retrieved information, yet they remain restricted by static ontologies and inefficient graph traversal algorithms. Thus, dynamic, tailored algorithms for knowledge retrieval and reasoning are crucial for maximizing the utility of KGs in combination with LLMs. Traditional GraphRAG systems rely predominantly on node-level retrievals, limiting their scalability~\cite{hu-etal-2025-grag,mavromatis-karypis-2025-gnn,luo2024graph}. They face difficulties in handling multi-hop reasoning tasks, where search strategy must evolve based on information received from intermediate steps. Further, static retrieval patterns limit their adaptability to varied domains and user intents: see Figure \ref{fig:problem visualisation}.

To address these challenges, we propose PersonalAI 2.0 (PAI-2), a GraphRAG method that incorporates graph-based external memory model to store unstructured textual knowledge alongside LLM-driven reasoning. By introducing a dynamic, multistage pipeline, PAI-2 guides graph traversal through explicit search planning and iterative query refinement. Its contributions lie in dynamically planned, iterative information searches, guided by entity extraction and vertex matching. By systematically decomposing complex queries into manageable subqueries, PAI-2 ensures focused retrieval of only relevant graph segments. Ultimately, this modification holds promise for improving factuality and reducing hallucinations across multi-hop reasoning tasks. The proposed method can be applied in a wide range of fields: from personalized education platforms to customer service chatbots, where contextual awareness and recall are highly important. 

In summary, our main contributions are as follows:
\begin{enumerate}
    \item We propose PersonalAI 2.0 (PAI-2), a GraphRAG method that effectively integrates graph-based external memory to store unstructured knowledge from texts and LLM reasoning abilities to plan information search and manage graph traversal.
    \item We evaluate PAI-2 on the Natural Questions, TriviaQA, HotpotQA, 2WikiMultihopQA, MuSiQue benchmarks and compare it with LightRAG, RAPTOR, HippoRAG 2, PAI-1 methods. Our method with Qwen2.5 7B and multilingual-e5-large shows 9 pp average gain by LLM-as-a-Judge on 2WikiMultihopQA and MuSiQue benchmarks. On TriviaQA and HotpotQA benchmarks PAI-2 shows comparable results with HippoRAG 2: difference in LLM-as-a-Judge scores is not statistically significant.
    \item We show that enabled plan enhancing mechanism for information search in PAI-2 increases answer accuracy by an average of 18 pp by LLM-as-a-Judge across five benchmarks.
    \item PAI-2 achieves superior result on MINE-1 benchmark, reaching 89\% information-retention score with Qwen2.5 7B and multilingual-e5-large. We show that PAI's memory construction algorithm is more stable (less LLM parsing errors), compared to KGGen and Wikontic methods in 7-14B LLM setting. 
    \item We conduct a systematic ablation of the Question Preprocessing stage (decomposition, denoising, enhancing), revealing that leaving questions unprocessed frequently yields higher LLM-as-a-Judge scores than applying the full preprocessing pipeline, and that dataset-specific configurations (denoising for Natural Questions, decomposition for 2WikiMultihopQA) achieve the best results. We derive the following practical deployment guidelines: use no preprocessing or denoising alone for general‑purpose use, enable decomposition for multi‑hop reasoning tasks and avoid activating all three preprocessing operations simultaneously.
\end{enumerate}
\section{Related Work}

The combination of LLMs and KGs has recently received considerable attention, aiming to address their respective limitations: LLMs sensitivity to hallucinations and incomplete reasoning versus KGs fragmentary coverage and static ontologies. In this section, we will briefly review several representative methods to enhance reasoning over KGs that illustrate distinct pathways toward modeling external memory for personalized LLM agents and implement planning procedure for information search.

PersonalAI 1.0 (PAI-1)~\cite{11479299} represents a systematic exploration of KG storage and retrieval approaches for personalized LLMs. By presenting a flexible graph-based memory framework, it bridges the gap between dense vector similarity retrieval and structured memory representations. This study underscores the necessity of dynamic retrieval methods, emphasizing multiple traversal mechanisms such as BeamSearch and WaterCircles. However, its focus on memory representation leaves room for improvement concerning scalability and applicability to closed-domain tasks. Think-on-Graph (ToG)~\cite{DBLP:conf/iclr/SunXTW0GNSG24} introduces a tight coupling (LLM $\bigotimes$ KG) paradigm, enabling direct participation of LLMs in graph reasoning processes. ToG exploits the advantages of multi-hop reasoning paths and improves the responsiveness and interpretability of reasoning outcomes. Nonetheless, its dependence on the KG's integrity and relevance limits its adaptability to evolving domains and dynamic user requirements. Reasoning on Graphs (RoG)~\cite{luo2024rog} addresses the hallucination problem by employing a planning-retrieval-reasoning framework. It grounds LLM-generated reasoning steps onto verified KG-derived paths, ensuring faithfulness and interpretability. Though successful in certain KGQA settings, its reliance on manual annotations and fine-tuning strategies restricts broader applicability. Debate on Graph (DoG)~\cite{ma2025debate} proposes an iterative interactive reasoning framework that combines simplified question transformations and debating among multi-role LLMs. This method excels in addressing overly complex and noisy paths, though its computational overhead might impede scalability. Pyramid-Driven Alignment (PDA)~\cite{Li2024AnEP} applies the Pyramid Principle to organize reasoning hierarchies derived from LLMs and KGs. By generating deductive knowledge and recursively unlocking KG reasoning capabilities, PDA achieves high accuracy on multi-hop reasoning tasks. However, its dependency on precise hierarchical organization complicates generalization to diverse contexts. Pseudo-Graph Generation \& Atomic Knowledge Verification (PG\&AKV)~\cite{PGAKV2025} emphasizes generalizability across KGs and open-ended question answering. It constructs pseudo-triples to fill knowledge gaps, followed by verification against actual KG triples. While this resolves certain issues around hallucination, its reliance on additional LLM computation adds latency.

Graph-based inference has also been applied to fact verification. TFGIN  ~\cite{wu2025tfgin} constructs tightly connected graphs from statements, tables, and logical programs, and uses graph inference to verify claims against tabular evidence. In contrast, PAI-2 uses entity-guided graph traversal to retrieve and summarize relevant evidence for answer generation, rather than performing explicit claim verification.

One of the prominent approaches in bridging reasoning and interactive decision-making within LLMs is exemplified by ReAct~\cite{yao2022react}. ReAct adopts a dual-purpose framework that combines verbal reasoning and action generation. Unlike traditional LLMs relying solely on internal representations, ReAct interacts with external sources (such as Wikipedia APIs) to enhance its reasoning capabilities. This hybrid architecture enhances decision-making and problem-solving, especially in dynamic environments requiring external knowledge integration. However, ReAct primarily focuses on isolated reasoning and interaction rather than scalable knowledge graph traversal, leaving room for exploration in broader-scale, graph-centric applications. Another critical area involves addressing hallucination and incomplete information retrieval in LLMs. Works such as FLARE~\cite{jiang-etal-2023-active} emphasize the significance of timing and selectivity in retrieving supplementary information. FLARE proposes an active retrieval strategy, ensuring that retrievals occur only when necessary, thereby minimizing redundancy and increasing efficiency. Nevertheless, this approach does not specifically target the challenge of navigating multi-step reasoning within complex knowledge graphs, focusing instead on improving document-level retrieval strategies. Additionally, Self-RAG~\cite{asai2024selfrag} offers insights into integrating reflective processes into LLMs. Self-RAG utilizes specialized tokens to trigger selective retrieval operations, ensuring better alignment between retrieved passages and final outputs. While Self-RAG improves factual consistency, its primary emphasis lies in fine-tuning individual document retrieval decisions rather than optimizing multi-hop traversals across heterogeneous knowledge structures.

Semantic matching is also central to cross-modal retrieval. Geometric Matching ~\cite{wang2024geometric} represents heterogeneous data using geometric regions to capture one-to-many correspondences between queries and retrieved items. Unlike this approach, PAI-2 performs text-based retrieval within a knowledge graph, combining entity matching with embedding-based filtering of retrieved triples. It does not require matching representations across different modalities.

CosFL ~\cite{qin2026cosfl} formulates fault localization as semantic code search. It uses LLM-generated queries describing suspected faulty functionality and retrieves matching program elements at the module, method and code-chunk levels. Both CosFL and PAI-2 use natural language queries to guide retrieval. However, CosFL targets bug localization in source code, whereas PAI-2 preprocesses user questions and retrieves relevant knowledge-graph evidence for question answering.

Graph-search efficiency has also been studied at the systems level. TianheGraph ~\cite{gan2022tianhegraph} optimizes breadth-first search on Tianhe supercomputers through buffering for high-degree vertices, vectorized searching, and optimized inter-node communication. While such approaches focus on the execution efficiency of graph traversal, PAI-2 addresses a different aspect of graph search: selecting relevant starting vertices, generating clue-queries to guide retrieval, and refining subsequent search steps based on retrieved knowledge. PAI-2 builds on existing graph traversal algorithms from PAI-1 ~\cite{11479299} with its main contribution lying in how these algorithms are coordinated within an adaptive retrieval procedure.

Recent multi-view clustering methods also use graph structure to establish correspondences between data representations. Cross-View Graph Matching  ~\cite{dong2026crossview} uses bipartite graph representations and iteratively refines cross-view sample correspondences. Selective Cross-View Topology ~\cite{dong2025selective} constructs a graph of relationships between views using optimal transport distances, selecting neighboring views to recover missing information. In contrast, PAI-2 operates on a graph of textual entities and their relations. It matches query entities to candidate vertices and uses clue-queries to guide traversal. Multiple starting vertices allow retrieval from potentially disconnected subgraphs. Thus, PAI-2 uses graph topology to retrieve relevant evidence rather than learning cross-view correspondences or completing missing views.

AGREE ~\cite{chen2026bridge} studies attributed graph clustering with mixed numerical and categorical node attributes. It combines attribute alignment with quaternion-based graph convolution to learn representations from node features and topology. In contrast, PAI-2 uses a graph with distinct vertex and relation types, where textual attributes and separately indexed embeddings support retrieval. Rather than learning graph representations for clustering, PAI-2 matches query entities to existing vertices and traverses their relations to retrieve relevant information.

Recent LLM-based applications for personalized services include LLMApp ~\cite{liu2026llmapp}, which compresses app usage histories, incorporates spatiotemporal context, and uses hierarchical LLM reasoning to predict future app usage. Unlike this task-specific prediction framework, PAI-2 focuses on knowledge retrieval for question answering. It grounds query entities in an external knowledge graph, generates clue-queries for graph traversal, and revises its search plan based on retrieved evidence.

PAI-2 builds on the general principle of reasoning-guided interaction with external information, as in ReAct~\cite{yao2022react}, but applies it to retrieval from a structured memory graph. Whereas ReAct interleaves reasoning traces and tool actions, PAI-2 maintains an explicit sequence of graph-search steps, revises the remaining steps using retrieved knowledge, and grounds each step in candidate graph vertices to generate clue-queries for traversal. The contribution lies in integrating these operations into a graph-specific retrieval procedure:
\begin{itemize}
    \item \textbf{Search Plan Generation}. Unlike ToG, which performs reasoning on knowledge graph implicitly, PAI-2 generates an explicit search plan, consisting of isolated natural-language queries.
    \item \textbf{Search Plan Adjustment}. Compared to PDA, which generates one static search plan, PAI-2 dynamically adjusts still-uncompleted steps or adds additional steps, then new information from memory graph is received.
    \item \textbf{Clue-Queries Generation}. In contrast to DoG and PDA, where entities from the search query are matched to single vertices and the same search query is used for alignment of graph traversal, PAI-2 allows match entity with multiple vertices and transforms search query into several clue-queries to retrieve relevant triples.
    \item \textbf{Information Retrieval Strategy}. On each search step, ToG, DoG and PDA can only traverse relations and retrieve triples, which are incident to vertices from the previous steps, while PAI-2 retrieves information without that restriction.
    \item \textbf{No Fine-tuning Requirements}. Like ReAct and FLARE, PAI-2 operates with a frozen LLM at inference time. This contrasts with Self-RAG, which trains an LLM to use specialized reflection tokens.
\end{itemize}





\section{Methods}

\begin{figure*}[th!]
	\centering
	\includegraphics[width=.85\textwidth]{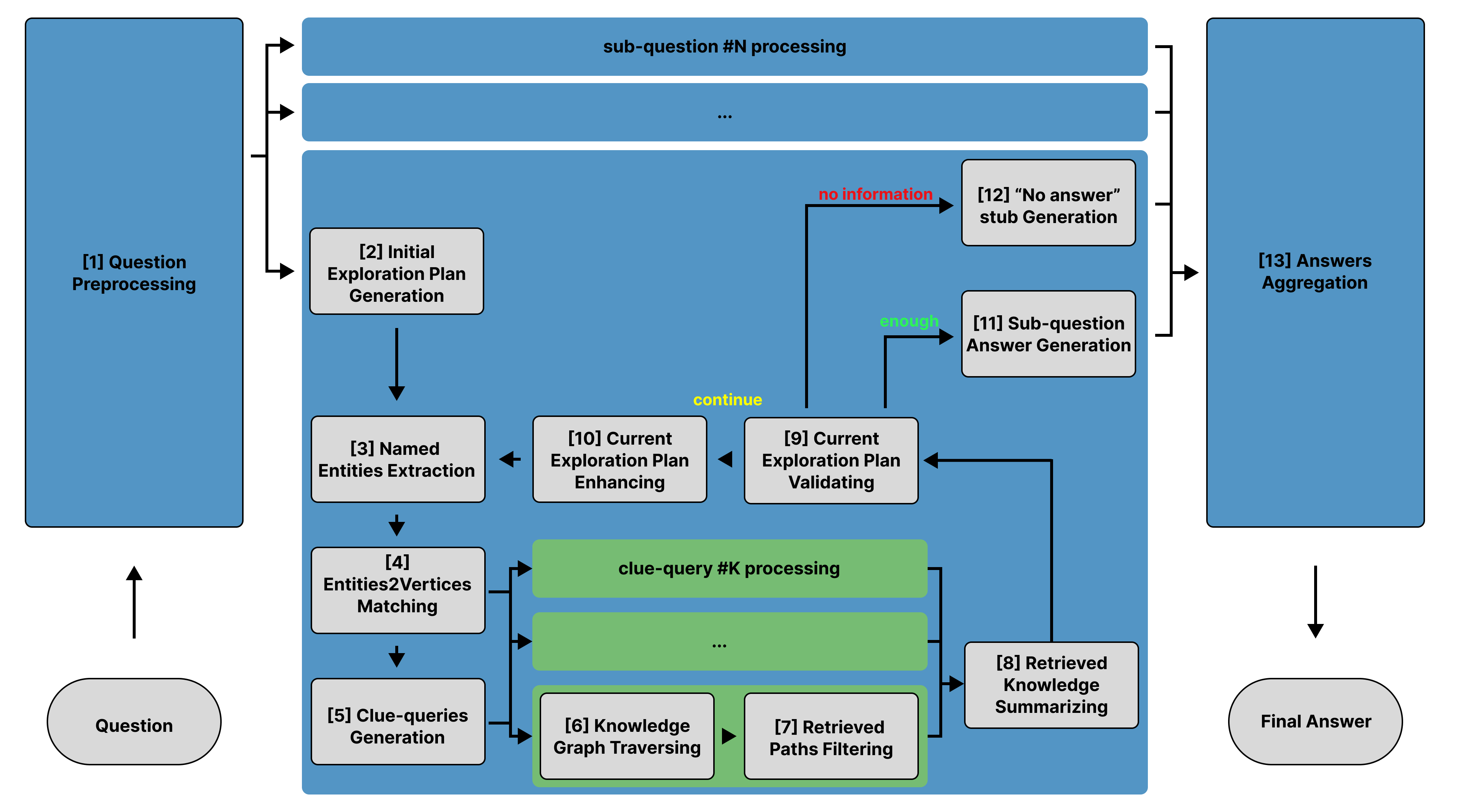}
	\caption{PAI-2's QA pipeline for information search in a memory graph} 				
	\label{fig:mediumqaipeline_diagram}
\end{figure*}

The proposed PAI-2 method draws from PAI-1 approach~\cite{11479299} in structured knowledge retrieval, starting from natural language queries and exploring outward based on a dynamic planning mechanism within a memory graph: see Figure~\ref{fig:mediumqaipeline_diagram}.

Specifically, subdivision of complex questions into manageable sub-questions allows targeted retrieval of only relevant portions of the underlying knowledge in existing memory. Iterative refining ensures gradual accumulation of necessary context until appropriate confidence level is reached for formulating coherent answer. By extracting named entities from search steps and matching them to vertices from memory graph, PAI-2 effectively grounds abstract concepts onto concrete stored representation. Subsequent refinement of entity matches via graph traversal and triplet filtering ensures that only high-relevance knowledge contributes to downstream reasoning processes. In this section, we will explain each step of PAI-2's QA pipeline in detail. 

\subsection{Question Preprocessing}

Given a question $q$, PAI-2 first preprocesses it to ensure clarity and modularity. This step involves a chain of operations: denoising, enhancement and decomposition. Formally, the set of sub-questions $\mathcal{Q} = \{q_1, q_2, \ldots, q_N\}$ is derived as:
\begin{equation}\label{eq:preprocess}
\begin{split}
\mathcal{Q} &= \text{Preprocess}(q) \\ &= \text{Decompose}(\text{Enhance}(\text{Denoise}(q))) \text{.}
\end{split}
\end{equation}

The function $\text{Denoise}(q)$ prompts LLM to correct syntactical errors and remove unnecessary stop words. This yields a cleaned question $q_{d}$:
\begin{equation}
\begin{split}
q_{d} &= \text{PROMPT}_{\text{syntax}}(\text{PROMPT}_{\text{stopwords}}(q)) \text{.}
\end{split}
\end{equation}

The function $\text{Enhance}(q)$ prompts LLM to edit the question according to grammatical rules, rephrase it using precise terminology and expand its meaning for clarity. This yields an enhanced question $q_{e}$:
\begin{equation}
\begin{split}
q_{e} &= \text{PROMPT}_{\text{grammar}}(\text{PROMPT}_{\text{terms}}(\text{PROMPT}_{\text{expand}}(q))) \text{.}
\end{split}
\end{equation}

Finally, the function $\text{Decompose}(q)$ determines whether $q$ contains several independent questions. We define a classification function:
\begin{equation}
\text{PROMPT}_{\text{decompose\_cls}}(q) = \left\{ \begin{array}{rcl}
\text{True}, & \text{if } q \text{ is composite} \text{.} \\ 
\text{False}, & \text{otherwise} \text{.}
\end{array} \right.
\end{equation}
If \text{True}, LLM splits $q$ into independent sub-questions $\mathcal{Q}$. 

\subsection{Memory Graph Exploration}

For each sub-question $q_i \in \mathcal{Q}$, a memory graph exploration operation is performed to search for relevant information and generate an accurate answer $a_i$. This operation consists of an iterative loop over search steps. At the $j$-th search step, the current search plan is represented as
\[
P^j = [s_1, s_2, \ldots, s_M],
\]
where $s_j$ is the current natural language query.

\textbf{Search Plan Initialization and Entity Grounding.} At the beginning of the exploration for $q_i$, an initial exploration plan $P^0$ is generated represented as a collection of search steps:
\begin{equation}
P^0 = \text{PROMPT}_{\text{plan\_init}}(q_i) \text{.}
\end{equation}

Given a search step $s_j$, LLM extracts a set of named entities:
\[
E_j = \text{NER}(s_j) = \{e_1, e_2, \ldots, e_U\}.
\]
Then, we link $E_j$ to object vertices $V^j$ from the memory graph. Let $V_m$ be a hyperparameter representing the maximum number of object vertices linked to one entity. The matching operation is defined as:
\begin{equation}
V^j_{U \times V_m} = \text{Entities2VerticesMatching}(E_j, V_m) \text{.}
\end{equation}
For each entity extracted from the search step, candidate object vertices are retrieved using dense embedding similarity and sparse BM25-based matching. Up to $V_m$ vertices per entity are retained as candidates for graph traversal.

To guide the traversal, a linear combination is performed on $V^j$, selecting the top $C_m$ vertex groups. We then prompt LLM to generate $C_m$ clue-queries, collectively denoted by $CQ_j$, based on $s_j$ and the selected vertices:
\begin{equation}
CQ_j = \text{ClueQueriesGen}(s_j, V^j_{C_m \times U}) \text{.}
\end{equation}
A clue-query represents a reformulation of $s_j$ with respect to a specific group of object vertices. 

\textbf{Graph Traversal and Knowledge Synthesis.} Each clue-query $cq_l \in CQ_j$ guides an independent graph traversal starting from the corresponding vertices $V[l]$. The resulting raw triples are obtained as
\[
T_l^{raw} = \text{KGraphTraverse}(cq_l, V[l]).
\]
In our experiments, KGraphTraverse uses the existing traversal and retrieval algorithms from PAI-1 ~\cite{11479299}, namely BeamSearch, WaterCircles and NaiveRetriever. The proposed planning mechanism determines the clue-queries and starting vertices passed to these algorithms, rather than modifying the underlying traversal algorithms.

The retrieved triples are ranked by embedding similarity to the corresponding clue-query $cq_l$ and the top $F_m$ triples are retained:
\begin{equation}
T_{l} = \text{FilterByRelevance}(cq_l, T^{raw}_{l}) \text{.}
\end{equation}

All filtered triples are summarized in two stages. First, LLM generates an answer for each clue-query using the corresponding filtered triples:
\[
\begin{aligned}
ca_l &= \text{ClueAnswerGen}(cq_l, T_l),\\
CA_j &= \{ca_1, \ldots, ca_{C_m}\}.
\end{aligned}
\]
Second, the clue-answers are summarized with respect to $s_j$ to produce the step knowledge $sa_j$:
\begin{equation}
sa_j = \text{SummarizeClueAnswers}(s_j, CQ_j, CA_j) \text{,}
\end{equation}
where $sa_j$ represents the knowledge retrieved from the memory graph with respect to search step $s_j$.

\textbf{Iterative Refinement and Answering.} The knowledge accumulated before the current search step is represented as
\[
SA^{j-1} = [sa_1, \ldots, sa_{j-1}].
\]
Together with the newly discovered knowledge $sa_j$, it is used by LLM to determine whether an answer $a_i$ can be generated for $q_i$:
\begin{equation}
\text{PROMPT}_{\text{answer\_cls}}(q_i, P^j, SA^j) = \left\{ \begin{array}{rl}
\text{True}, & \text{if relevant } a_i \\ & \text{can be generated} \text{.} \\ 
\text{False}, & \text{otherwise} \text{.}
\end{array} \right.
\end{equation}
If \text{True}, the LLM generates the answer  $a_i$:
\begin{equation}
    a_i = \text{PROMPT}_{\text{answer\_subq}}(q_i, P^j, SA^j) \text{.}   
\end{equation}
If \text{False}, LLM determines whether the current search plan $P^j$ needs modification:
\begin{equation}
\text{PROMPT}_{\text{plan\_enhance\_cls}}(q_i, P^j, SA^j) = \left\{ \begin{array}{rl}
\text{True}, & \text{if } P^j \text{ needs} \\ & \text{enhancement} \text{.} \\ 
\text{False}, & \text{otherwise} \text{.}
\end{array} \right.
\end{equation}
If plan enhancement is required, the current plan is updated using the newly discovered knowledge:
\[
P^{j+1} = \text{SearchPlanEnhance}(q_i, P^j, SA^j).
\]
The procedure then continues with the next search step $s_{j+1}$. If the maximum number of search steps is exceeded without sufficient knowledge, the system returns a "No Answer" stub:
\[
a_i = \text{NoAnswerStubGen}(q_i).
\]

\subsection{Answer Aggregation}

After receiving all answers $\mathcal{A} = [a_1, a_2, \ldots, a_N]$ to sub-questions $\mathcal{Q} = [q_1, q_2, \ldots, q_N]$, we prompt LLM to generate the final answer $a$ to the initial question $q$:
\begin{equation}
a = \text{AggregateSubAnswers}(q, \mathcal{Q}, \mathcal{A}) \text{.}
\end{equation}

This string-formatted output is returned as the result of PAI-2's QA pipeline. Pseudocode of proposed information search algorithm is presented in Appendix~\ref{app:medium_qa_pseudocode}. Example of PAI-2 execution for specific question is presented in Appendix~\ref{app:pai2_example}.
\section{Experiment Set-Up}

This section outlines the research questions, experimental configurations and implementation details used in our study.

\subsection{Research Question Definitions}
\label{subsec:research_questions}

In our study, we aim to answer the following research questions:
\begin{itemize}
    \item \textbf{RQ1}: Can PAI-2 achieve superior results compared to baselines?
    \item \textbf{RQ2}: How does PAI-2's efficiency vary with respect to maximum number of a steps in a search plan?
    \item \textbf{RQ3}: How does PAI-2's efficiency vary with respect to number of generating clue-queries per step of the search plan?  
    \item \textbf{RQ4}: Does the plan enhancing mechanism improves PAI-2's answers accuracy compared to static plan execution setting?
    \item \textbf{RQ5}: How does PAI-2's efficiency vary with enabled Question Preprocessing stage in different settings?
\end{itemize}

To choose LLM backbone for PAI-2 in our main experiments we performed a few-shot evaluation on HotpotQA benchmark. Preprocessing operations and resulted documents subset for memory graphs construction described in Appendix~\ref{app:datasets}. We selected several LLMs from the 7-9B tier: Qwen2.5 7B\footnote{\url{https://ollama.com/library/qwen2.5:7b}}, Llama3.1 8B\footnote{\url{https://ollama.com/library/llama3.1:8b}}, Granite3.3 8B\footnote{\url{https://ollama.com/library/granite3.3:8b}} and Gemma2 9B\footnote{\url{https://ollama.com/library/gemma2:9b}}. Evaluation was performed on first 100 question-answer pairs from the resulted HotpotQA subset. Table~\ref{tab:llm_selection} shows that best LLM across four metrics is Qwen2.5 7B. To create vector representations of knowledge in a memory graph we employed a combination of dense and sparse embeddings: \textit{intfloat/multilingual-e5-large}\footnote{\url{https://huggingface.co/intfloat/multilingual-e5-large}} and BM25.

\begin{table*}[ht!]
    \centering
    \renewcommand{\arraystretch}{1.3}
    \caption{Averaged performance in a few-shot ablation for PAI-1 and PAI-2 on HotpotQA benchmark across four LLMs. Cells contain Context Relevance, Faithfulness, LLM-as-a-Judge and Groundedness scores, respectively, averaged across nine configurations of graph traversal algorithm to identify optimal LLM for the main experiments. Bold values indicate best performance on corresponding metric per method.}
    \begin{tabular}{|c|c|c|c|c||c|}
    \hhline{|-|----||-|}
    \multirow{2}{*}{\textbf{Method}} & \multicolumn{4}{c||}{\textbf{LLM}} & \multirow{2}{*}{Mean} \\ 
    \hhline{|~|----||~|}
     & \textbf{Qwen2.5 7B} & \textbf{Llama3.1 8B} & \textbf{Granite3.3 8B} & \textbf{Gemma2 9B} &  \\ 
     \hhline{=====::=}
    PAI-1 & \textbf{0.60} / 0.59 / 0.41 / 0.13 & 0.52 / 0.46 / 0.38 / 0.11 & 0.59 / \textbf{0.61} / 0.44 / \textbf{0.32} & 0.54 / 0.58 / \textbf{0.48} / 0.13 & 0.56 / 0.56 / 0.43 / 0.17 \\
    \hhline{|-|----||-|}
    PAI-2 & \textbf{0.70} / \textbf{0.82} / \textbf{0.44} / \textbf{0.52} & 0.64 / 0.70 / \textbf{0.44} / 0.42 & 0.54 / 0.72 / 0.39 / 0.51 & 0.65 / 0.66 / \textbf{0.44} / 0.45 & 0.63 / 0.73 / 0.43 / 0.48 \\ 
    \hhline{=====::=}
    Mean & 0.65 / 0.7 / 0.42 / 0.32 & 0.58 / 0.58 / 0.41 / 0.26 & 0.56 / 0.66 / 0.42 / 0.42 & 0.6 / 0.62 / 0.46 / 0.29 & 0.6 / 0.64 / 0.43 / 0.32 \\ 
    \hhline{|-|----||-|}
    \end{tabular}
	\label{tab:llm_selection}
\end{table*}

For PAI-2's knowledge graph traversal procedure we select two combinations of BeamSearch (BS), WaterCircles (WC) and NaiveRetriever (NR) algorithms, presented in PAI-1~\cite{11479299}, as they give superior and comparative performance by our previous research: "BS + WC" and "BS + NR". Settings for base algorithms are fixed (see Appendix~\ref{app:retriev_hyper}). During graph traversal we did not apply constraints on vertex types, but during triples-filtering stage, episode-typed triples are discarded.

\subsection{Summary of evaluated configurations}

For \textbf{RQ1}, best PAI-2 configurations and baselines were evaluated (1) on 1000 question-answer pairs from Natural Questions, HotpotQA, 2WikiMultihopQA and MuSiQue, and (2) on 500 question-answer pairs from TriviaQA. For \textbf{RQ2}, \textbf{RQ3}, \textbf{RQ4}, \textbf{RQ5} each PAI-2 configuration was evaluated on first 100 question-answer pairs from each benchmark. The same LLM was used for both response generation (using the given QA configuration) and the corresponding memory-graph construction. Consequently, for each benchmark 20 distinct PAI-2 configurations were derived. In total, 100 configurations were evaluated, plus 44 configurations for the LLM few-shot ablation study on HotpotQA. 

The few-shot ablation was used solely to select best LLM backbone for PAI-2. It was not used to tune other hyperparameters or configurations reported in the main experiments. The few-shot evaluation was performed by executing PAI-2's QA pipeline on previously constructed memory graphs containing documents from the preprocessed HotpotQA benchmark. We used only the first 100 question-answer pairs from this subset, which is separate from the main RQ1 test sets. The main RQ1 test sets (1000 QA pairs per benchmark, except 500 for TriviaQA) were held out entirely during configuration development. No test set metrics were used, directly or indirectly, to select reported configurations. Other configuration choices (traversal algorithms, embedding models, search parameters) were selected based on our previous research on PAI-1 and not on RQ1 results.

\subsection{Implementation details}

Our memory graph implementation consists of two main parts: a graph part and a vector part. The graph part stores textual representations of object, thesis and episodic vertices together with their properties and relationships (edges). Neo4j is used for this part of the system. The vector part of the memory stores embeddings of graph elements to measure semantic similarity of texts during QA-pipeline execution. Qdrant and OpenSearch are used for this part of the system to store dense and sparse embeddings respectively. The memory graph can be considered heterogeneous, as it contains object, thesis and episodic vertex types linked by different types of relations, including ordinary edges and hyperedges inherited from PAI-1~\cite{11479299}. It is also attributed, since graph elements contain textual properties, while their embeddings are indexed separately in the vector storage. PAI also implements a caching mechanism for storing intermediate results of QA pipeline steps that reduces the overall time required to process incoming questions. It utilizes two non-relational databases: Redis and MongoDB. During our experiments cache was enabled. Efficiency and performance scores were obtained on release version 2.3.7\footnote{\url{https://github.com/Dzigen/Personal-AI/releases/tag/v2.3.7}} of the PersonalAI library. All databases ran on a single machine, each in its own Docker container. Medium-sized LLMs (7-14B) were hosted in a local Ollama Docker container. LLM generation strategy during memory construction and QA pipeline execution for PAI-2 and baselines were fixed: num\_predict = 2048, seed = 42, top\_k = 1, temperature = 0.0. The maximum number of steps in a PAI-2 search plan is fixed and set to 8. To run our experiments we used machine with the following characteristics:
\begin{itemize}
    \item CPU: Intel(R) Xeon(R) CPU E5-2698 v4 @ 2.20GHz (4 processors, 80 cores in total).
    \item GPU: NVIDIA TITAN RTX 24GB.
    \item RAM: 504 GB.
    \item HDD: Western Digital Ultrastar DC HC550 18 TB WUH721818ALE6L4 SATA 6Gb/s (SATA-III).
    \item OS: Ubuntu 22.04.3 LTS (Jammy Jellyfish).
\end{itemize}

To evaluate PAI-2, we constructed five memory graphs based on five selected and preprocessed benchmarks. An average speed of adding documents (with 492 average length) into the memory per minute is approximately 1.63. Detailed characteristics of constructed memory graphs can be found in Appendix~\ref{app:constr_graphs}. 
\section{Evaluation}

This section describes benchmarks, metrics and baseline methods used to evaluate PAI-2.

\subsection{Benchmarks}

To evaluate the proposed method we conducted experiments on five distinct benchmarks. This selection was designed to cover varying domains, structural complexities and reasoning requirements, thereby mitigating potential bias arising from limited domain diversity:
\begin{itemize}
    \item \textbf{Natural Questions}~\cite{kwiatkowski-etal-2019-natural} is a large-scale corpus developed by Google that consists of over 307K samples, each of which includes a natural-language query paired with relevant Wikipedia pages containing the answer spans. The questions originated from real user searches on Google. Key distinguishing features compared to other benchmarks include: (1) diversity in question types -- NQ encompasses factual, definitional, list-based, comparative, and opinion-oriented queries; (2) complex answer requirements -- answers can be short text snippets or long passages requiring deeper reasoning.
    \item \textbf{TriviaQA}~\cite{joshi-etal-2017-triviaqa} is a large-scale benchmark designed for factoid question answering, featuring over 95K question-answer pairs sourced from Bing. Its distinguishing characteristics include: (1) multi-evidence reasoning -- answers often require synthesizing information across multiple documents rather than relying on single-sentence evidence; (2) contextual complexity -- diverse types of questions from various domains like history, science, literature. Compared with benchmarks such as SQuAD or Natural Questions (both of which focus primarily on extractive QA within structured contexts) TriviaQA emphasizes multi-hop inference and retrieval-based tasks. This makes it particularly suitable for evaluating advanced machine reading comprehension systems that can handle complex queries and require broad contextual understanding.
    \item \textbf{HotpotQA}~\cite{yang-etal-2018-hotpotqa} is a crowdsourced question answering corpus built on English Wikipedia, comprising approximately 113K questions. each question is constructed to require the combination of information from the introductory sections of two Wikipedia articles. Benchmark provides two gold paragraphs per question, along with a list of sentences identified as supporting facts necessary to answer the question. HotpotQA includes various reasoning strategies such as bridge questions (involving missing entities), intersection questions (e.g., "what satisfies both property A and property B?") and comparison questions (comparing two entities through a common attribute).
    \item \textbf{2WikiMultihopQA}~\cite{ho-etal-2020-constructing} is a multi-hop question answering corpus that contains complex questions requiring reasoning over multiple Wikipedia paragraphs. Each question is designed to necessitate logical connections across different pieces of information to arrive at the correct answer.
    \item \textbf{MuSiQue}~\cite{10.1162/tacl_a_00475} is a challenging multi-hop QA benchmark containing approximately 25K 2-4 hop questions, constructed by composing single-hop questions from five existing single-hop QA corpora. It is designed to feature diverse and complex reasoning paths, requiring models to integrate information from multiple hops to generate correct answers.
\end{itemize}

Considering the computational and engineering complexity of constructing and traversing large memory graphs, we created manageable yet representative subsets from the original benchmarks. This step was necessary to enable the iterative experimentation required for tuning multiple retrieval algorithms and LLM configurations within practical resource constraints. The resulting subsets used for memory graph construction and evaluation are summarized in Table~\ref{tab:datasets_summ}. Detailed preprocessing steps and benchmarks statistics are provided in Appendix~\ref{app:datasets}.

\begin{table}[ht!]
\centering
\renewcommand{\arraystretch}{1.3}
\caption{Characteristics of prepared QA corpora for PAI-2 and baseline evaluation}
\resizebox{.35\textwidth}{!}{%
\begin{tabular}{|c|c|c|}
\hline
\textbf{Benchmark} & \textbf{\#qa-pairs} & \textbf{\#documents} \\ \hline \hline
Natural Questions           & 2000                & 3970                \\ \hline
TriviaQA         & 500                 & 4925                \\ \hline
HotpotQA         & 2000                & 3933                \\ \hline
2WikiMultihopQA         & 2000                 & 4596                \\ \hline
MuSiQue         & 1931                & 4185                \\ \hline
\end{tabular}%
}
\label{tab:datasets_summ}
\end{table}

\subsection{Metrics}

Traditional statistical evaluation metrics such as BLEU~\cite{papineni2002bleu}, ROUGE~\cite{lin2004rouge} and Meteor Universal~\cite{denkowski2014meteor} struggle to distinguish syntactically similar, but semantically distinct texts. While semantic methods like BERTScore~\cite{zhang2019bertscore} were introduced to address these limitations, our experiments reveal that BERTScore lacks sufficient differentiability, often failing to capture nuanced distinctions between correct and incorrect answers. Therefore, we adopt \textbf{LLM-as-a-Judge}~\cite{zheng2023judging} framework and choose Qwen2.5 7B. The judge evaluates question-answer pairs using a structured prompt that contains question, ground truth and generated answers. It labels $1$ for a correct answer and $0$ for an incorrect one, and we use accuracy as our primary metric. Corresponding LLM prompts and details are provided in Appendix~\ref{app:judgescore_hyperp}.

To validate reliability of LLM as an evaluative judge, human annotation was conducted for best PAI-2 and HippoRAG 2 experimental setups. The responses generated by the Qwen2.5 7B model were annotated using Overlap-3 metric, with domain experts adhering to the same evaluation criteria as the automated judge. Inter-annotator agreement was quantified using Krippendorff's alpha~\cite{Krippendorff2011ComputingKA}, yielding a mean $\alpha$ = 0.93, which indicates a high degree of assessment reliability. Further evaluation of the alignment between the automated judge and human annotators was conducted by computing the raw agreement, Pearson correlation coefficient and Cohen's kappa~\cite{Cohen1960ACO} between the judge's scores and the majority vote derived from human annotations. They all indicate substantial to near-perfect agreement between automated and human judgments across all GraphRAG-methods and benchmarks: mean raw agreement = 0.95; mean $r$ = 0.88; mean $\kappa$ = 0.88. Details regarding the annotation procedure are provided in Appendix~\ref{app:human_evaluation}.

Additionally, to measure other characteristics we used several LLM-based metrics from RAGAS\footnote{\url{https://docs.ragas.io/en/stable/}} library: 
\begin{itemize}
    \item \textbf{Context Relevance} measures whether the retrieved contexts are pertinent to the user question. This is done via two independent LLM-as-a-Judge prompt calls that each rates the relevance on a scale of $0$, $1$ or $2$. The ratings are then converted to a $[0,1]$ scale and averaged to produce the final score. Higher scores indicate that the contexts are more closely aligned with the user question.
    \item \textbf{Faithfulness} measures how factually consistent an answer is with the retrieved context. It ranges from $0$ to $1$, with higher scores indicating better consistency. An answer is considered faithful if all its claims can be supported by the retrieved context.
    \item \textbf{Groundedness} measures how well an answer is supported or "grounded" by the retrieved contexts. It assesses whether each claim in the answer can be found, either wholly or partially, in the provided contexts: $0$ if the answer is not grounded in the context at all; $1$ if the answer is partially grounded, and $2$ if the answer is fully grounded (every statement can be found or inferred from the retrieved context).
\end{itemize}

\subsection{Baselines}

We compare PAI-2 with the following baseline methods:
\begin{itemize}
    \item \textbf{LightRAG}~\cite{guo-etal-2025-lightrag} is a simple alternative to modern GraphRAG methods that focuses on performance. LightRAG is a graph-structured RAG framework that employs a dual-level retrieval system, combining low-level entity retrieval with high-level knowledge discovery. It integrates graph structures with vector representations for efficient retrieval of related entities and their relationships.
    \item \textbf{RAPTOR}~\cite{sarthi2024raptor} is a RAG framework which enhances retrieval via recursive summarisation and hierarchical clustering into a tree structure. RAPTOR recursively clusters chunks of text based on their vector embeddings and generates text summaries of those clusters, constructing a tree from the bottom up. Nodes clustered together are siblings; a parent node contains the text summary of that cluster.
    \item \textbf{HippoRAG 2}~\cite{gutiérrez2025ragmemorynonparametriccontinual} is a non-parametric continual learning framework that leverages a Personalized PageRank algorithm over an open knowledge graph constructed using LLM-extracted triples. It enhances multi-hop reasoning capabilities through sophisticated graph traversal and passage integration mechanisms.
    \item \textbf{PersonalAI 1.0} (PAI-1)~\cite{11479299} is a flexible framework for creating external graph-based memory for AI Agents. Built upon AriGraph~\cite{anokhin2024arigraphlearningknowledgegraph} architecture, PAI-1 introduces a novel hybrid graph design that supports both standard edges and two types of hyper-edges, enabling rich and dynamic semantic and temporal representations. Also, it supports diverse retrieval mechanisms, including A*, WaterCircles traversal, BeamSearch and hybrid methods, making it adaptable to different datasets and LLM capacities. 
\end{itemize}

We evaluate baselines using Qwen2.5 as the LLM backbone for graph construction and information-search algorithms. For baseline methods we select embedding models that were identified as the best choices in the respective papers by the original authors: LightRAG -- \textit{BAAI/bge-m3}\footnote{\url{https://huggingface.co/BAAI/bge-m3}}; RAPTOR -- \textit{sentence-transformers/multi-qa-mpnet-base-cos-v1}\footnote{\url{https://huggingface.co/sentence-transformers/multi-qa-mpnet-base-cos-v1}}; HippoRAG 2 -- \textit{facebook/contriever}\footnote{\url{https://huggingface.co/facebook/contriever}}; PAI-1 -- a fusion of \textit{intfloat/multilingual-e5-large}\footnote{\url{https://huggingface.co/intfloat/multilingual-e5-large}} and BM25. Additionally we evaluate LightRAG, RAPTOR and HippoRAG 2 with specified \textit{intfloat/multilingual-e5-large} embedder. Also we took implementations of baselines from the original GitHub repositories: LightRAG -- release v1.4.12\footnote{\url{https://github.com/HKUDS/LightRAG/releases/tag/v1.4.12}}; RAPTOR -- commit 7da1d48\footnote{\url{https://github.com/parthsarthi03/raptor/tree/7da1d48}}; HippoRAG 2 -- commit d437bfb\footnote{\url{https://github.com/OSU-NLP-Group/HippoRAG/tree/d437bfb}}; PAI-1 -- release v2.3.7\footnote{\url{https://github.com/Dzigen/PersonalAI/releases/tag/v2.3.7}}.
\section{Experiments and Results}




This section presents the experimental results for the five research questions defined in Section~\ref{subsec:research_questions}, followed by additional evaluations.

\subsection{RQ1: Comparison with Baseline Methods}
\label{subsec:rq1_comparison}

To evaluate PAI-2's effectiveness, we conducted comparative experiments on five established benchmarks. Table~\ref{tab:main_comparison} summarizes the best-performing configurations by LLM-as-a-Judge metric.

\begin{table*}[th!]
    \centering
    \caption{Best LLM-as-a-Judge scores for LightRAG, RAPTOR, HippoRAG 2, PAI-1 and PAI-2 on five benchmarks. Bold values indicate best performance per benchmark.}
	\renewcommand{\arraystretch}{1.3}
	\resizebox{\textwidth}{!}{%
    \begin{tabular}{|cc|c|c|c|c|c|c|c||c|}
    \hhline{---------||-}
    \multicolumn{2}{|c|}{\multirow{2}{*}{\textbf{Method}}} & \multirow{2}{*}{\textbf{LLM}} & \multirow{2}{*}{\textbf{Embedder}} & \multicolumn{5}{c||}{\textbf{Benchmark}} & \multirow{2}{*}{Mean} \\ 
    \hhline{~~~~-----||~}
    \multicolumn{2}{|c|}{} &  &  & \multicolumn{1}{c|}{\textbf{Natural Questions}} & \multicolumn{1}{c|}{\textbf{TriviaQA}} & \multicolumn{1}{c|}{\textbf{HotpotQA}} & \multicolumn{1}{c|}{\textbf{2WikiMultihopQA}} & \multicolumn{1}{c||}{\textbf{MuSiQue}} &  \\ 
    \hhline{=========::=}
    \multicolumn{2}{|c|}{\multirow{2}{*}{LightRAG}} & \multirow{12}{*}{Qwen2.5 7B} & bge-m3 & \multicolumn{1}{c|}{0.77} & \multicolumn{1}{c|}{0.67} & \multicolumn{1}{c|}{0.63} & \multicolumn{1}{c|}{0.37} & 0.32 & 0.55 \\ 
    \hhline{~~~------||-}
    \multicolumn{2}{|c|}{} &  & m-e5-large & \multicolumn{1}{c|}{\textbf{0.83}} & \multicolumn{1}{c|}{0.75} & \multicolumn{1}{c|}{0.69} & \multicolumn{1}{c|}{0.34} & 0.33 & 0.59 \\ 
    \hhline{--~------||-}
    \multicolumn{2}{|c|}{\multirow{2}{*}{RAPTOR}} &  & \begin{tabular}[c]{@{}c@{}}multi-qa-mpnet-\\ base-cos-v1\end{tabular} & \multicolumn{1}{c|}{0.70} & \multicolumn{1}{c|}{0.72} & \multicolumn{1}{c|}{0.49} & \multicolumn{1}{c|}{0.20} & 0.21 & 0.46 \\ 
    \hhline{~~~------||-}
    \multicolumn{2}{|c|}{} &  & m-e5-large & \multicolumn{1}{c|}{0.69} & \multicolumn{1}{c|}{0.73} & \multicolumn{1}{c|}{0.55} & \multicolumn{1}{c|}{0.26} & 0.23 & 0.49 \\ 
    \hhline{--~------||-}
    \multicolumn{2}{|c|}{\multirow{2}{*}{HippoRAG 2}} &  & contriever & \multicolumn{1}{c|}{0.79} & \multicolumn{1}{c|}{\textbf{0.80}} & \multicolumn{1}{c|}{0.70} & \multicolumn{1}{c|}{0.56} & 0.33 & 0.63 \\ 
    \hhline{~~~------||-}
    \multicolumn{2}{|c|}{} &  & m-e5-large & \multicolumn{1}{c|}{0.79} & \multicolumn{1}{c|}{0.79} & \multicolumn{1}{c|}{\textbf{0.72}} & \multicolumn{1}{c|}{0.50} & 0.32 & 0.62 \\ 
    \hhline{--~------||-}
    \multicolumn{2}{|c|}{PAI-1 (Ours)} &  & m-e5-large & \multicolumn{1}{c|}{0.55} & \multicolumn{1}{c|}{0.70} & \multicolumn{1}{c|}{0.53} & \multicolumn{1}{c|}{0.31} & 0.21 & 0.46 \\ 
    \hhline{==|~|======::=}
    \multicolumn{1}{|c|}{\multirow{3}{*}{PAI-2 (Ours)}} & \begin{tabular}[c]{@{}c@{}}w/o question \\ preprocessing\end{tabular} &  & \multirow{3}{*}{m-e5-large} & \multicolumn{1}{c|}{0.72} & \multicolumn{1}{c|}{0.77} & \multicolumn{1}{c|}{0.70} & \multicolumn{1}{c|}{0.61} & \textbf{0.42} & \textbf{0.64} \\ 
    \hhline{~-~~-----||-}
    \multicolumn{1}{|c|}{} & \begin{tabular}[c]{@{}c@{}}w question \\ preprocessing\end{tabular} &  &  & \multicolumn{1}{c|}{0.70} & \multicolumn{1}{c|}{0.75} & \multicolumn{1}{c|}{0.62} & \multicolumn{1}{c|}{\textbf{0.65}} & 0.39 & 0.62 \\ 
    \hhline{---------||-}
    \end{tabular}}
	\label{tab:main_comparison}
\end{table*}

\begin{table*}[th!]
    \centering
    \caption{Impact of search steps amount on PAI-2's QA Pipeline performance. Cells contain Context Relevance, Faithfulness and LLM-as-a-Judge scores. Bold values indicate first (with increase of search steps) best performance on corresponding metric per benchmark. Red cells indicate configurations with lowest LLM-as-a-Judge scores per benchmark. Green cells indicate configurations with first highest LLM-as-a-Judge scores per benchmark.}
	\renewcommand{\arraystretch}{1.3}
	\resizebox{\textwidth}{!}{
    \begin{tabular}{|c|c|c|c|c|c||c|}
    \hhline{|------||-|}
    \multirow{2}{*}{\textbf{Max Search Steps}} & \multicolumn{5}{c||}{\textbf{Benchmark}} & \multirow{2}{*}{Mean} \\
    \hhline{|~-----||~|}
     & \multicolumn{1}{c|}{\textbf{Natural Questions}} & \multicolumn{1}{c|}{\textbf{TriviaQA}} & \multicolumn{1}{c|}{\textbf{HotpotQA}} & \multicolumn{1}{c|}{\textbf{2WikiMultihopQA}} & \multicolumn{1}{c||}{\textbf{MuSiQue}} &  \\ 
     \hhline{======::=}
    \multicolumn{1}{|c|}{2} & \multicolumn{1}{c|}{\cellcolor{lightsalmonpink} 0.92 / 0.89 / 0.70} & \multicolumn{1}{c|}{\cellcolor{grannysmithapple} \textbf{0.96} / 0.85 / \textbf{0.82}} & \multicolumn{1}{c|}{\cellcolor{lightsalmonpink} 0.87 / 0.81 / 0.69} & \multicolumn{1}{c|}{\cellcolor{lightsalmonpink} 0.85 / 0.75 / 0.48} & \multicolumn{1}{c||}{\cellcolor{lightsalmonpink} 0.87 / 0.77 / 0.46} & \multicolumn{1}{c|}{\cellcolor{lightsalmonpink} 0.89 / 0.81 / 0.63} \\
    \hhline{|------||-|}
    \multicolumn{1}{|c|}{4} & \multicolumn{1}{c|}{\cellcolor{grannysmithapple} \textbf{0.94} / 0.89 / \textbf{0.71}} & \multicolumn{1}{c|}{0.95 / \textbf{0.86} / 0.82} & \multicolumn{1}{c|}{0.92 / 0.80 / 0.73} & \multicolumn{1}{c|}{\textbf{0.89} / 0.79 / 0.58} & \multicolumn{1}{c||}{\textbf{0.90} / 0.79 / 0.45} & \multicolumn{1}{c|}{\textbf{0.92} / 0.83 / 0.66} \\
    \hhline{|------||-|}
    \multicolumn{1}{|c|}{8} & \multicolumn{1}{c|}{0.94 / \textbf{0.90} / 0.71} & \multicolumn{1}{c|}{\cellcolor{lightsalmonpink} 0.96 / 0.86 / 0.81} & \multicolumn{1}{c|}{\cellcolor{grannysmithapple} \textbf{0.94} / 0.81 / \textbf{0.75}} & \multicolumn{1}{c|}{\cellcolor{grannysmithapple} 0.89 / \textbf{0.82} / \textbf{0.61}} & \multicolumn{1}{c||}{\cellcolor{grannysmithapple} 0.89 / \textbf{0.80} / \textbf{0.47}} & \multicolumn{1}{c|}{\cellcolor{grannysmithapple} 0.92 / \textbf{0.84} / \textbf{0.67}} \\
    \hhline{|------||-|}
    \multicolumn{1}{|c|}{12} & \multicolumn{1}{c|}{0.94 / 0.89 / 0.71} & \multicolumn{1}{c|}{0.96 / 0.85 / 0.82} & \multicolumn{1}{c|}{0.93 / 0.81 / 0.74} & \multicolumn{1}{c|}{0.89 / 0.82 / 0.61} & \multicolumn{1}{c||}{0.90 / 0.78 / 0.46} & \multicolumn{1}{c|}{0.92 / 0.83 / 0.67} \\
    \hhline{|------||-|}
    \multicolumn{1}{|c|}{16} & \multicolumn{1}{c|}{0.94 / 0.89 / 0.71} & \multicolumn{1}{c|}{0.96 / 0.86 / 0.82} & \multicolumn{1}{c|}{0.93 / \textbf{0.82} / 0.73} & \multicolumn{1}{c|}{0.89 / 0.82 / 0.61} & \multicolumn{1}{c||}{0.90 / 0.79 / 0.46} & \multicolumn{1}{c|}{0.92 / 0.84 / 0.67} \\
    \hhline{======::=}
    Mean & 0.94 / 0.89 / 0.71 & 0.96 / 0.86 / 0.82 & 0.92 / 0.81 / 0.73 & 0.88 / 0.8 / 0.58 & 0.89 / 0.79 / 0.46 & 0.92 / 0.83 / 0.66 \\
    \hhline{|------||-|}
    \end{tabular}}
    \label{tab:max-searchsteps-impact}
\end{table*}

\begin{table*}[th!]
    \centering
    \caption{Impact of clue-queries amount (per search step) on PAI-2's QA Pipeline performance. Cells contain Context Relevance, Faithfulness and LLM-as-a-Judge scores. Bold values indicate best performance on corresponding metric per benchmark. Red cells indicate configurations with lowest LLM-as-a-Judge scores per benchmark. Green cells indicate configurations with highest LLM-as-a-Judge scores per benchmark.}
	\renewcommand{\arraystretch}{1.3}
	\resizebox{\textwidth}{!}{
    \begin{tabular}{|c|c|c|c|c|c||c|}
    \hhline{|-|-----||-|}
    \multicolumn{1}{|c|}{\multirow{2}{*}{\textbf{Max Clue-Queries}}} & \multicolumn{5}{c||}{\textbf{Benchmark}} & \multirow{2}{*}{Mean} \\ 
    \hhline{|~|-----||~|}
    \multicolumn{1}{|c|}{} & \multicolumn{1}{c|}{\textbf{Natural Questions}} & \multicolumn{1}{c|}{\textbf{TriviaQA}} & \multicolumn{1}{c|}{\textbf{HotpotQA}} & \multicolumn{1}{c|}{\textbf{2WikiMultihopQA}} & \multicolumn{1}{c||}{\textbf{MuSiQue}} &  \\ 
    \hhline{======::=}
    \multicolumn{1}{|c|}{1} & \multicolumn{1}{c|}{\cellcolor{lightsalmonpink} - / - / 0.65} & \multicolumn{1}{c|}{\cellcolor{lightsalmonpink} 0.95 / \textbf{0.89} / 0.77} & \multicolumn{1}{c|}{0.91 / 0.79 / 0.75} & \multicolumn{1}{c|}{\cellcolor{grannysmithapple} 0.86 / 0.77 / \textbf{0.62}} & \multicolumn{1}{c||}{\cellcolor{lightsalmonpink} 0.81 / \textbf{0.80} / 0.43} & \cellcolor{lightsalmonpink} 0.88 / 0.81 / 0.64 \\ 
    \hhline{|-|-|-|-|-|-||-|}
    \multicolumn{1}{|c|}{2} & \multicolumn{1}{c|}{ - / - / 0.69} & \multicolumn{1}{c|}{0.94 / 0.87 / 0.79} & \multicolumn{1}{c|}{\cellcolor{lightsalmonpink} 0.93 / \textbf{0.82} / 0.71} & \multicolumn{1}{c|}{0.85 / 0.77 / 0.59} & \multicolumn{1}{c||}{\cellcolor{grannysmithapple} 0.89 / \textbf{0.80} / \textbf{0.48}} & 0.90 / 0.82 / 0.65 \\ 
    \hhline{|-|-|-|-|-|-||-|}
    \multicolumn{1}{|c|}{4}& \multicolumn{1}{c|}{0.94 / 0.88 / 0.70} & \multicolumn{1}{c|}{0.93 / \textbf{0.89} / 0.80} & \multicolumn{1}{c|}{\cellcolor{grannysmithapple} \textbf{0.95} / 0.81 / \textbf{0.76}} & \multicolumn{1}{c|}{\cellcolor{lightsalmonpink} 0.85 / 0.77 / 0.58} & \multicolumn{1}{c||}{\textbf{0.90} / \textbf{0.80} / 0.45} & 0.91 / 0.83 / 0.66 \\ 
    \hhline{|-|-|-|-|-|-||-|}
    \multicolumn{1}{|c|}{6} & \multicolumn{1}{c|}{\cellcolor{grannysmithapple} 0.94 / 0.89 / \textbf{0.73}} & \multicolumn{1}{c|}{\cellcolor{grannysmithapple} \textbf{0.96} / 0.86 / \textbf{0.81}} & \multicolumn{1}{c|}{0.94 / \textbf{0.82} / 0.74} & \multicolumn{1}{c|}{\textbf{0.89} / 0.75 / 0.61} & \multicolumn{1}{c||}{0.86 / 0.78 / 0.47} & \textbf{0.92} / 0.82 / \textbf{0.67} \\ 
    \hhline{|-|-|-|-|-|-||-|}
    \multicolumn{1}{|c|}{8} & \multicolumn{1}{c|}{0.94 / \textbf{0.90} / 0.71} & \multicolumn{1}{c|}{\textbf{0.96} / 0.86 / \textbf{0.81}} & \multicolumn{1}{c|}{0.94 / 0.81 / 0.75} & \multicolumn{1}{c|}{\textbf{0.89} / \textbf{0.82} / 0.61} & \multicolumn{1}{c||}{0.89 / \textbf{0.80} / 0.47} & \cellcolor{grannysmithapple}\textbf{0.92} / \textbf{0.84} / \textbf{0.67} \\ 
    \hhline{======::=}
    \multicolumn{1}{|c|}{Mean} & \multicolumn{1}{c|}{0.94 / 0.90 / 0.71} & \multicolumn{1}{c|}{0.95 / 0.87 / 0.80} & \multicolumn{1}{c|}{0.93 / 0.81 / 0.74} & \multicolumn{1}{c|}{0.87 / 0.78 / 0.60} & \multicolumn{1}{c||}{0.87 / 0.80 / 0.46} & 0.91 / 0.82 / 0.66 \\ 
    \hhline{|-|-|-|-|-|-||-|}
    \end{tabular}}
	\label{tab:main_maxcluequeries_ablation}
\end{table*}

\begin{table*}[th!]
    \centering
    \renewcommand{\arraystretch}{1.3}
    \caption{Impact of plan enhancing mechanism on PAI-2's performance. Cells contain Context Relevance, Faithfulness and LLM-as-a-Judge scores. Bold values indicate best performance on corresponding metric per benchmark. Red cells indicate configurations with lowest LLM-as-a-Judge scores per benchmark. Green cells indicate configurations with highest LLM-as-a-Judge scores per benchmark.}
	\resizebox{\textwidth}{!}{
    \begin{tabular}{|c|c|ccccc||c|}
    \hhline{|-|-|-----||-|}
    \multirow{2}{*}{\begin{tabular}[c]{@{}c@{}}\textbf{Retrieval} \\ \textbf{Algorithm}\end{tabular}} & \multirow{2}{*}{\begin{tabular}[c]{@{}c@{}}\textbf{Plan Enhancing} \\ \textbf{Mechanism}\end{tabular}} & \multicolumn{5}{c||}{\textbf{Benchmark}} & \multirow{2}{*}{Mean} \\
    \hhline{|~|~|-----||~|}
     &  & \multicolumn{1}{c|}{\textbf{Natural Questions}} & \multicolumn{1}{c|}{\textbf{TriviaQA}} & \multicolumn{1}{c|}{\textbf{HotpotQA}} & \multicolumn{1}{c|}{\textbf{2WikiMultihopQA}} & \textbf{MuSiQue} &  \\ 
    \hhline{=======::=}
    NR & \multirow{3}{*}{Disabled} & \multicolumn{1}{c|}{\cellcolor{lightsalmonpink} 0.89 / \textbf{0.94} / 0.64} & \multicolumn{1}{c|}{0.90 / 0.88 / 0.77} & \multicolumn{1}{c|}{\cellcolor{lightsalmonpink} 0.80 / \textbf{0.82} / 0.55} & \multicolumn{1}{c|}{\cellcolor{lightsalmonpink} - / - / 0.26} & \multicolumn{1}{c||}{0.53 / 0.72 / 0.12} & \cellcolor{lightsalmonpink} 0.78 / 0.84 / 0.47 \\
    \hhline{|-|~|-|-|-|-|-||-|}
    BS+NR &  & \multicolumn{1}{c|}{0.92 / 0.88 / 0.68} & \multicolumn{1}{c|}{0.90 / \textbf{0.87} / 0.77} & \multicolumn{1}{c|}{0.80 / 0.77 / 0.58} & \multicolumn{1}{c|}{0.69 / 0.76 / 0.33} & \multicolumn{1}{c||}{0.57 / 0.69 / 0.16} & 0.78 / 0.79 / 0.5 \\ 
    \hhline{|-|~|-|-|-|-|-||-|}
    BS+WC &  & \multicolumn{1}{c|}{0.91 / 0.92 / 0.64} & \multicolumn{1}{c|}{\cellcolor{lightsalmonpink} 0.88 / 0.85 / 0.74} & \multicolumn{1}{c|}{0.80 / 0.80 / 0.59} & \multicolumn{1}{c|}{0.65 / 0.82 / 0.34} & \multicolumn{1}{c||}{\cellcolor{lightsalmonpink} 0.46 / 0.71 / 0.11} & 0.74 / 0.82 / 0.48 \\
    \hhline{==|=====::=}
    NR & \multirow{3}{*}{Enabled} & \multicolumn{1}{c|}{\textbf{0.95} / \textbf{0.94} / 0.68} & \multicolumn{1}{c|}{0.94 / 0.86 / 0.80} & \multicolumn{1}{c|}{0.92 / \textbf{0.82} / \textbf{0.75}} & \multicolumn{1}{c|}{0.88 / 0.75 / 0.58} & \multicolumn{1}{c||}{0.86 / \textbf{0.83} / 0.48} & 0.91 / 0.84 / 0.66 \\
    \hhline{|-|~|-|-|-|-|-||-|}
    BS+NR &  & \multicolumn{1}{c|}{\cellcolor{grannysmithapple} 0.94 / 0.90 / \textbf{0.71}} & \multicolumn{1}{c|}{\cellcolor{grannysmithapple} \textbf{0.96} / 0.86 / \textbf{0.81}} & \multicolumn{1}{c|}{\cellcolor{grannysmithapple} \textbf{0.94} / 0.81 / \textbf{0.75}} & \multicolumn{1}{c|}{0.89 / 0.82 / 0.61} & \multicolumn{1}{c||}{\textbf{0.89} / 0.80 / 0.47} & \cellcolor{grannysmithapple} \textbf{0.92} / 0.84 / \textbf{0.67} \\ 
    \hhline{|-|~|-|-|-|-|-||-|}
    BS+WC &  & \multicolumn{1}{c|}{0.93 / 0.93 / 0.68} & \multicolumn{1}{c|}{0.94 / \textbf{0.87} / 0.78} & \multicolumn{1}{c|}{0.93 / 0.78 / 0.73} & \multicolumn{1}{c|}{\cellcolor{grannysmithapple} \textbf{0.90} / \textbf{0.83} / \textbf{0.65}} & \multicolumn{1}{c||}{\cellcolor{grannysmithapple} 0.86 / \textbf{0.83} / \textbf{0.51}} & 0.91 / \textbf{0.85} / \textbf{0.67} \\
    \hhline{|-|-|-|-|-|-|-||-|}
    \end{tabular}}
	\label{tab:main_planenhancement_ablation}
\end{table*}

\begin{table*}[th!]
    \centering
	\renewcommand{\arraystretch}{1.3}
	\caption{Impact of Question Preprocessing stage configurations on PAI-2's performance. Cells contain Context Relevance, Faithfulness and LLM-as-a-Judge scores. Bold values indicate best performance on corresponding metric per benchmark. Red cells indicate configurations with lowest LLM-as-a-Judge scores per benchmark. Green cells indicate configurations with highest LLM-as-a-Judge scores per benchmark. For Question Preprocessing stage "+" and "-" means that corresponding operation is enabled and disabled, respectively.}
    \resizebox{\textwidth}{!}{
    \begin{tabular}{|ccc|ccccc|c|}
    \hhline{|---|-----||-|}
    \multicolumn{3}{|c|}{\textbf{Question Preprocessing Operations}} & \multicolumn{5}{c||}{\textbf{Benchmark}} & \multirow{2}{*}{Mean} \\
    \hhline{|---|-----||~|}
    \multicolumn{1}{|c|}{\textbf{Decomposition}} & \multicolumn{1}{c|}{\textbf{Denoising}} & \multicolumn{1}{c|}{\textbf{Enhancing}} & \multicolumn{1}{c|}{\textbf{Natural Questions}} & \multicolumn{1}{c|}{\textbf{TriviaQA}} & \multicolumn{1}{c|}{\textbf{HotpotQA}} & \multicolumn{1}{c|}{\textbf{2WikiMultihopQA}} & \multicolumn{1}{c||}{\textbf{MuSiQue}} &  \\
    \hhline{========::=}
    \multicolumn{1}{|c|}{-} & \multicolumn{1}{c|}{-} & \multicolumn{1}{c|}{-} & \multicolumn{1}{c|}{- / - / 0.64} & \multicolumn{1}{c|}{\cellcolor{grannysmithapple} \textbf{0.96} / 0.87 / \textbf{0.81}} & \multicolumn{1}{c|}{\cellcolor{grannysmithapple} \textbf{0.92} / 0.79 / 0.73} & \multicolumn{1}{c|}{0.87 / \textbf{0.81} / 0.65} & \multicolumn{1}{c||}{\cellcolor{grannysmithapple} \textbf{0.83} / \textbf{0.79} / \textbf{0.45}} & \cellcolor{grannysmithapple} \textbf{0.90} / 0.82 / \textbf{0.66} \\
    \hhline{|-|-|-|-|-|-|-|-||-|}
    \multicolumn{1}{|c|}{-} & \multicolumn{1}{c|}{-} & \multicolumn{1}{c|}{+} & \multicolumn{1}{c|}{0.93 / 0.93 / 0.63} & \multicolumn{1}{c|}{0.93 / \textbf{0.93} / 0.77} & \multicolumn{1}{c|}{0.88 / 0.77 / 0.70} & \multicolumn{1}{c|}{0.82 / 0.78 / 0.54} & \multicolumn{1}{c||}{0.78 / 0.77 / 0.38} & 0.87 / \textbf{0.84} / 0.60 \\
    \hhline{|-|-|-|-|-|-|-|-||-|}
    \multicolumn{1}{|c|}{-} & \multicolumn{1}{c|}{+} & \multicolumn{1}{c|}{-} & \multicolumn{1}{c|}{\cellcolor{grannysmithapple} \textbf{0.95} / 0.92 / \textbf{0.67}} & \multicolumn{1}{c|}{0.92 / 0.89 / 0.80} & \multicolumn{1}{c|}{0.91 / 0.75 / 0.64} & \multicolumn{1}{c|}{0.86 / 0.76 / 0.65} & \multicolumn{1}{c||}{0.80 / 0.77 / 0.42} & 0.89 / 0.82 / 0.64 \\
    \hhline{|-|-|-|-|-|-|-|-||-|}
    \multicolumn{1}{|c|}{-} & \multicolumn{1}{c|}{+} & \multicolumn{1}{c|}{+} & \multicolumn{1}{c|}{\cellcolor{lightsalmonpink} 0.90 / \textbf{0.94} / 0.61} & \multicolumn{1}{c|}{0.94 / 0.92 / 0.76} & \multicolumn{1}{c|}{0.87 / 0.81 / 0.66} & \multicolumn{1}{c|}{0.83 / 0.74 / 0.54} & \multicolumn{1}{c||}{0.75 / 0.77 / 0.44} & 0.86 / \textbf{0.84} / 0.60 \\
    \hhline{|-|-|-|-|-|-|-|-||-|}
    \multicolumn{1}{|c|}{+} & \multicolumn{1}{c|}{-} & \multicolumn{1}{c|}{-} & \multicolumn{1}{c|}{- / - / 0.64} & \multicolumn{1}{c|}{0.94 / 0.88 / \textbf{0.81}} & \multicolumn{1}{c|}{0.88 / \textbf{0.82} / 0.60} & \multicolumn{1}{c|}{\cellcolor{grannysmithapple} \textbf{0.88} / 0.78 / \textbf{0.71}} & \multicolumn{1}{c||}{0.82 / \textbf{0.79} / 0.42} & 0.88 / 0.82 / 0.64 \\
    \hhline{|-|-|-|-|-|-|-|-||-|}
    \multicolumn{1}{|c|}{+} & \multicolumn{1}{c|}{-} & \multicolumn{1}{c|}{+} & \multicolumn{1}{c|}{0.92 / 0.93 / 0.62} & \multicolumn{1}{c|}{\cellcolor{lightsalmonpink} 0.93 / 0.90 / 0.70} & \multicolumn{1}{c|}{0.84 / \textbf{0.82} / 0.62} & \multicolumn{1}{c|}{0.82 / 0.77 / 0.58} & \multicolumn{1}{c||}{\cellcolor{lightsalmonpink} 0.77 / 0.77 / 0.31} & \cellcolor{lightsalmonpink} 0.86 / \textbf{0.84} / 0.57 \\ 
    \hhline{|-|-|-|-|-|-|-|-||-|}
    \multicolumn{1}{|c|}{+} & \multicolumn{1}{c|}{+} & \multicolumn{1}{c|}{-} & \multicolumn{1}{c|}{0.93 / \textbf{0.94} / 0.66} & \multicolumn{1}{c|}{0.92 / 0.90 / 0.78} & \multicolumn{1}{c|}{\cellcolor{lightsalmonpink} 0.84 / 0.74 / 0.60} & \multicolumn{1}{c|}{0.87 / 0.80 / 0.70} & \multicolumn{1}{c||}{0.79 / 0.76 / 0.41} & 0.87 / 0.83 / 0.63 \\
    \hhline{|-|-|-|-|-|-|-|-||-|}
    \multicolumn{1}{|c|}{+} & \multicolumn{1}{c|}{+} & \multicolumn{1}{c|}{+} & \multicolumn{1}{c|}{0.90 / 0.92 / 0.62} & \multicolumn{1}{c|}{0.92 / 0.92 / 0.71} & \multicolumn{1}{c|}{0.82 / \textbf{0.82} / 0.61} & \multicolumn{1}{c|}{0.83 / 0.79 / 0.59} & \multicolumn{1}{c||}{0.76 / 0.76 / 0.40} & 0.85 / \textbf{0.84} / 0.59 \\ 
    \hhline{========::=}
    \multicolumn{3}{|c|}{Mean} & \multicolumn{1}{c|}{0.92 / 0.93 / 0.64} & \multicolumn{1}{c|}{0.93 / 0.90 / 0.77} & \multicolumn{1}{c|}{0.87 / 0.79 / 0.64} & \multicolumn{1}{c|}{0.85 / 0.78 / 0.62} & \multicolumn{1}{c||}{0.79 / 0.77 / 0.40} & 0.87 / 0.83 / 0.61 \\
    \hhline{|---|-|-|-|-|-||-|}
    \end{tabular}}
	\label{tab:main_queryprep_ablation}
\end{table*}


As shown in Table~\ref{tab:main_comparison}, PAI-2 achieves superior results with a 9 pp average gain (difference is statistically significant by McNemar's paired test with $p < 0.05$) by LLM-as-a-Judge on 2 out of 5 benchmarks: 2WikiMultihopQA and MuSiQue. These benchmarks are characterized by complex multi-hop reasoning requirements, which demonstrates PAI-2's strength in handling intricate question decomposition and iterative information retrieval. Meanwhile, on TriviaQA and HotpotQA, it achieves comparable (difference is not statistically significant by McNemar's paired test with $p > 0.05$) results to HippoRAG 2: with a 3 pp and 2 pp difference by LLM-as-a-Judge, respectively.

It can also be seen that TriviaQA turned out to be the easiest benchmark (in terms of question difficulty), while MuSiQue took the place of the most difficult one, with average LLM-as-a-Judge scores 0.74 and 0.31, respectively. In turn, it can be noticed a gap (difference is statistically significant by McNemar's paired test with $p < 0.05$) between PAI-2 and HippoRAG 2 scores on Natural Questions benchmark: with a 7 pp difference by LLM-as-a-Judge. This observation may be attributed to the following characteristics of this benchmark:

\begin{itemize}
    \item \textbf{Case sensitivity issues}. The original questions are presented in lowercase, which increases the probability of missing critical named entities, consequently losing essential information required for relevant response generation.
    \item \textbf{Ambiguous answer formats}. Some questions require general or insufficiently specific answers, for example, "how are the American declaration of independence and the French declaration of the rights of man similar". Due to PAI-2's implementation, which does not have an explicit mechanism for detecting question type and expected answer format at the decision-making stage, it often returns "No Answer": uncertainty regarding the completeness of retrieved/summarized knowledge from memory.
    \item \textbf{Single-hop nature}. Natural Questions predominantly contains single-hop questions, which may not fully leverage PAI-2's search planning capabilities designed for multi-hop reasoning scenarios.
\end{itemize}

Compared to our previous version of PAI framework (PAI-1), the proposed QA pipeline in PAI-2 significantly improves answer quality: an average improvement is 18 pp by LLM-as-a-Judge across all benchmarks. This substantial improvement demonstrates that integration of the planning stage with search steps enhancing mechanism and knowledge graph traversal based on a set of detailed/adjusted clue-queries increases the probability of extracting relevant information from a structured document store and improves consistency of the final answer with the existing knowledge base. The most significant improvements are observed on multi-hop benchmarks: 100\% improvement on MuSiQue (0.42 vs. 0.21) and 110\% improvement on 2WikiMultihopQA (0.65 vs. 0.31), confirming that the planning mechanism particularly benefits complex reasoning tasks.

Among baseline methods, LightRAG demonstrates strong performance on Natural Questions (0.83), but struggles with multi-hop benchmarks: 2WikiMultihopQA -- 0.37; MuSiQue -- 0.33. RAPTOR shows the lowest overall performance (0.46 mean), particularly on complex reasoning tasks. HippoRAG 2 emerges as the strongest baseline (0.63 mean), achieving best results on TriviaQA (0.80) and HotpotQA (0.72), which validates its effectiveness for factoid and two-hop reasoning scenarios.

The analysis of triples retrieval performance reveals a fundamental trade-off between PAI-1 and PAI-2 architectures that explains the observed improvements in answer quality. As shown in Table~\ref{tab:recall_vs_precision_triplesretrieval}, PAI-2 achieves higher recall across all three benchmarks: representing a 14 pp improvement (0.35 vs. 0.21 mean recall).

\begin{table}[th!]
    \centering
    \caption{Precision and Recall of relevant triples retrieval from constructed memory graphs for PAI-1 and PAI-2 on three benchmarks}
    \renewcommand{\arraystretch}{1.3}
    \begin{tabular}{|cccc||c|}
    \hhline{----||-}
    \multicolumn{1}{|c|}{\multirow{3}{*}{\textbf{Method}}} & \multicolumn{3}{c||}{\textbf{Benchmark}} & \multirow{3}{*}{\textbf{Mean}} \\ 
    \hhline{~---||~}
    \multicolumn{1}{|c|}{} & \multicolumn{1}{c|}{\begin{tabular}[c]{@{}c@{}}\textbf{Natural} \\ \textbf{Questions}\end{tabular}} & \multicolumn{1}{c|}{\begin{tabular}[c]{@{}c@{}}\textbf{2Wiki} \\ \textbf{MultihopQA}\end{tabular}} & \multicolumn{1}{c||}{\textbf{MuSiQue}} &  \\ 
    \hhline{====::=}
    \multicolumn{5}{|c|}{\textbf{Precision}} \\ 
    \hhline{-----}
    \multicolumn{1}{|c|}{PAI-1} & \multicolumn{1}{c|}{0.21} & \multicolumn{1}{c|}{0.31} & \multicolumn{1}{c||}{0.24} & 0.26 \\ 
    \hhline{----||-}
    \multicolumn{1}{|c|}{PAI-2} & \multicolumn{1}{c|}{0.13} & \multicolumn{1}{c|}{0.14} & \multicolumn{1}{c||}{0.13} & 0.13 \\ 
    \hhline{====::=}
    \multicolumn{5}{|c|}{\textbf{Recall}} \\ 
    \hhline{-----}
    \multicolumn{1}{|c|}{PAI-1} & \multicolumn{1}{c|}{0.29} & \multicolumn{1}{c|}{0.17} & \multicolumn{1}{c||}{0.16} & 0.21 \\ 
    \hhline{----||-}
    \multicolumn{1}{|c|}{PAI-2} & \multicolumn{1}{c|}{0.34} & \multicolumn{1}{c|}{0.38} & \multicolumn{1}{c||}{0.32} & 0.35 \\
    \hhline{----||-}
    \end{tabular}
    \label{tab:recall_vs_precision_triplesretrieval}
\end{table}

This enhanced retrieval coverage directly supports PAI-2's search planning mechanism, which generates multiple clue-queries per step to explore broader knowledge graph vertices and edges. However, this comes at the cost of reduced precision, as the adaptive query generation retrieves a wider set of candidate triples, including some less relevant matches: PAI-1 -- 0.26; PAI-2 -- 0.13. This finding validates PAI-2's core design hypothesis that comprehensive information retrieval, guided by iterative search planning, outweighs the benefits of higher-precision, but narrower retrieval for complex QA scenarios requiring multi-hop inference.

Statistical significance analysis of obtained results with exact p-values can be found in Appendix~\ref{app:main_statsign_analysis}.

\subsection{RQ2: Impact of Search Steps Amount}
\label{subsec:rq2_searchsteps}

To investigate how PAI-2's efficiency varies with respect to the maximum number of steps in a search plan, we conducted experiments varying the maximum search steps from 2 to 16. Results are presented in Table~\ref{tab:max-searchsteps-impact}.

From Table~\ref{tab:max-searchsteps-impact} it can be seen several important patterns regarding the relationship between search steps amount and answer quality. Configurations with only 2 search steps consistently underperform across benchmarks, achieving the lowest LLM-as-a-Judge scores on 4 out of 5 benchmarks with a mean score of 0.63. Increasing the maximum search steps from 2 to 8 yields substantial improvements, particularly on complex multi-hop reasoning benchmarks. For 2WikiMultihopQA, the LLM-as-a-Judge score improves from 0.48 to 0.61 (13 pp gain), while HotpotQA improvement from 0.69 to 0.75 (6 pp gain). These results confirm that multi-hop questions require more iterations to successfully traverse memory graph and aggregate distributed information across multiple entities and relationships.

Configurations with 12 and 16 steps show minimal to no improvement over 8 steps, with mean LLM-as-a-Judge scores remaining stable at 0.67. For TriviaQA, which predominantly contains single-hop factual questions, 2 steps already achieves the best LLM-as-a-Judge score (0.82), and additional steps provide no measurable benefit. This suggests that simpler questions do not require extensive search planning, and the optimal number of search steps should be adapted based on question complexity.

The observed patterns can be attributed to PAI-2's search planning mechanism. With only 2 steps, the system has limited opportunity to refine its search strategy based on intermediate retrieval results, potentially missing critical information scattered across the knowledge graph. At 8 steps, the system can perform sufficient iterations to: (1) identify initial entities from the query, (2) traverse related vertices, (3) generate and execute clue-queries based on retrieved information, (4) refine the search path based on intermediate results, and (5) aggregate comprehensive context for answer generation. Beyond 8 steps, additional iterations provide marginal benefit as the system has already retrieved most relevant information, and further steps may introduce redundant or marginally relevant context.

Based on these results, we recommend setting the maximum search steps to 8 for general-purpose deployments, as this configuration achieves optimal balance between answer quality and computational efficiency. For applications targeting predominantly single-hop questions (similar to TriviaQA), 2-4 steps may be sufficient. For complex multi-hop reasoning tasks (similar to 2WikiMultihopQA and MuSiQue), 8 steps is strongly recommended to achieve best performance.

\subsection{RQ3: Impact of Clue-Queries Amount}
\label{subsec:rq3_cluequeries}

To measure the impact of clue-queries amount, generated per search step, on the quality of generated answers, we conducted a series of experiments varying its maximum number from 1 to 8. Results are presented in  Table~\ref{tab:main_maxcluequeries_ablation}.

From Table~\ref{tab:main_maxcluequeries_ablation}, it can be observed that answer quality is improving with increasing number of clue-queries (used for managing graph traversal from associated starting vertices): an average 3 pp gain by LLM-as-a-Judge was achieved when changing the maximum number of clue-queries from 1 to 8. This observation may be attributed to characteristics of the memory graph construction algorithm. When a new document is added to memory, its extracted triples are validated for duplicates with triples in the existing memory graph. This validation and subsequent filtering (for duplicates) is performed by exact match metrics for triplet's text attributes. Because the same knowledge can be formulated in different ways, several subgraphs may appear in memory that contain and describe the same knowledge but using different entities. Such subgraphs may not share any common vertices and may be incomplete: information about a single object can be scattered across subgraphs. Therefore, using multiple vertices for a single entity from the search plan and subsequently generating clue-queries based on their linear combination allows us to traverse such subgraphs, find and aggregate requested information on a given object to generate an accurate and complete answer. This explains why higher clue-query counts particularly benefit benchmarks with complex entity relationships (2WikiMultihopQA and MuSiQue).

\subsection{RQ4: Plan Enhancing Mechanism Effectiveness}
\label{subsec:rq4_planenhancement}

We conducted a series of experiments to evaluate PAI-2 with disabled search plan enhancing mechanism. Table~\ref{tab:main_planenhancement_ablation} presents comprehensive results across different graph traversal/retrieval algorithm combinations and plan enhancing settings.

Table~\ref{tab:main_planenhancement_ablation} shows that, compared to PAI-2's configurations with enabled plan enhancing mechanism, accuracy of generated answers is degraded by an average of 18 pp by LLM-as-a-Judge when the mechanism is disabled. This observation can be attributed to the form/complexity of some questions that require a dynamic search strategy with intermediate grounding on available knowledge in a memory graph. This observation can be attributed to the form/complexity of some questions, that requires a dynamic search strategy with intermediate grounding on available knowledge in a memory graph. As an example, consider the question "Do both films Payment On Demand and My Cousin From Warsaw have the directors from the same country?". For it the following initial plan was generated: (1) "Who is the director of the film Payment On Demand?"; (2) "Who is the director of the film My Cousin From Warsaw?"; (3) "What country is the director of Payment On Demand from?"; (4) "What country is the director of My Cousin From Warsaw from?". It can be seen that the last two steps require clarification with use of information, obtained from the first, more specific and clear steps. With disabled plan enhancement, the following information will be available (by each search step) to generate final answer: (1) "Curtis Bernhardt"; (2) "Carl Boese is the director of the film My Cousin From Warsaw."; (3) "Germany"; (4) "<|NotEnoughtInfo|>" . It can be seen that it is not enough to generate a relevant answer. If the next search steps can be modified based on information, obtained by the previous ones, we get the following plan: (1) "Who is the director of the film Payment On Demand?"; (2) "Who is the director of the film My Cousin From Warsaw?"; (3) "What country is the director of Payment On Demand from?", (4) "What country is Carl Boese from?"; (5) "What country is Curtis Bernhardt from?". With that enhanced plan the following information will be available to generate the final answer: (1) "The director of the film Payment On Demand is Curtis Bernhardt."; (2) "The director of the film "My Cousin from Warsaw" is Carl Boese."; (3) "Curtis Bernhardt, the director of Payment On Demand, was born in New York, New York."; (4) "Carl Boese was a German film director, screenwriter, and producer."; (5) "Curtis Bernhardt, born as Kurt Bernhardt in Worms, Germany, was from Germany.". It can be seen that by improving the last steps, it was possible to extract crucial information from memory to generate an accurate answer.

Among retrieval algorithm combinations, BS+NR (BeamSearch + NaiveRetriever) with enabled plan enhancing mechanism achieves the best overall performance (0.67 mean LLM-as-a-Judge score), followed closely by BS+WC (BeamSearch + WaterCircles) at 0.67. The NR-only configuration shows the lowest performance (0.47 without enhancement, 0.66 with enhancement), indicating that combining multiple retrieval strategies provides more robust graph traversal capabilities.

\subsection{RQ5: Question Preprocessing Stage Evaluation}
\label{subsec:rq5_querypreprocessing}

To evaluate the impact of different Question Preprocessing stage configurations, we conducted experiments varying three preprocessing operations: Decomposition, Denoising, and Enhancing. Results are presented in Table~\ref{tab:main_queryprep_ablation}.

Several important observations emerge from Table~\ref{tab:main_queryprep_ablation}. Surprisingly, the configuration with all preprocessing operations disabled (-/-/-) achieves the highest mean LLM-as-a-Judge score (0.66). This suggests that for certain benchmark types, raw questions may preserve important semantic information that preprocessing might inadvertently remove. Secondly, different benchmarks benefit from different preprocessing configurations. For Natural Questions, Denoising alone (-/+/-) achieves the best score (0.67). For 2WikiMultihopQA, Decomposition alone (+/-/-) performs best (0.71). This indicates that preprocessing should be adapted based on the target dataset characteristics. Thirdly, the configuration with all three preprocessing operations enabled (+/+/+) achieves below-average performance (0.59), suggesting that excessive preprocessing may introduce noise or remove useful question information. Fourthly, Denoising operation shows consistent positive impact when used alone or in combination with one other operations, improving average performance by 2-3 pp across benchmarks.

Based on these results, we recommend: (1) for general-purpose deployments use no preprocessing or Denoising only; (2) for multi-hop reasoning tasks enable Decomposition operation; (3) for potentially noisy questions enable Denoising operation; (4) avoid enabling all three preprocessing operations simultaneously.

\subsection{Additional Evaluations}
\label{subsec:additional_evaluations}

Additionally, we evaluate our plain-text-to-knowledge-graph extraction algorithm (Memorize pipeline) on MINE benchmark to measure the factual completeness of constructed PAI's memory graphs. Our method demonstrates superior result with 89\% information-retention score: more details are provided in Appendix~\ref{app:mine_comparison}. Analysis of computational cost for PAI-2 is provided in Appendix~\ref{app:qapipe_execution_characteristics}.

\section{Conclusion}

In this paper we introduce PersonalAI 2.0 (PAI-2), a novel framework that integrates LLM capabilities with graph-based external memory for efficient knowledge retrieval and reasoning. Building upon GraphRAG approach, PAI-2 addresses critical limitations associated with traditional methods, such as inefficiencies in traversing complex knowledge graphs and deficiencies in retrieving precise, context-specific information.

Through a systematic question decomposition and dynamic planning of subgraph traversal/retrieval, PAI-2 demonstrates improvements over existing methods: LightRAG, RAPTOR, HippoRAG 2 and PAI-1. Evaluations conducted across five benchmarks (Natural Questions, TriviaQA, HotpotQA, 2WikiMultihopQA and MuSiQue) highlighted its effectiveness, achieving an average 9 pp improvement by LLM-as-a-Judge on 2 out of 5. According to ablation study, enabled search plan enhancing mechanism gives an 18 pp boost compared with its disabled version.

Additionally, experiments revealed that PAI-2’s memory construction algorithm exhibited greater stability than existing methods (KGGen and Wikontic) in the 7--14B LLM tier settings, yielding fewer parsing errors and achieving superior result on MINE-1 benchmark with an 89\% information-retention score.

Overall, PAI-2 represents a promising step forward in combining LLM capabilities with the structured data representation provided by knowledge graphs. It broadens approaches for developing LLM-based agentic systems capable of delivering nuanced responses and reliable factual outputs.


\section{Limitations}
\label{sec:limitations}

Despite its advantages, PAI-2 exhibits several limitations that represent opportunities for future research.

\textbf{Dependence on Underlying LLM.} PAI-2 performance depends on the capabilities of the backbone LLM used for planning, clue-query generation, entity extraction and answer synthesis. Our experiments used Qwen2.5 7B, selected based on few-shot evaluation (Section~\ref{subsec:research_questions}). However, smaller or less capable models may struggle with complex query decomposition and iterative reasoning tasks, which could lower retrieval quality. Larger models could improve performance but require more computational resources. This dependence introduces variability in system behavior across different LLM architectures and raises questions about generalizability to models with different training distributions or specialized domains. Future work should test robustness across diverse LLM families and explore techniques for reducing model dependency.


\textbf{Memory Construction Procedure.} The current memory graph construction relies on LLM-based triple extraction from raw documents, which introduces several constraints. First, the ontology structure is simple, offering limited capabilities for indexing and filtering information, which reduces the effectiveness of QA algorithms. Second, object vertices lack formal definitions, causing difficulties in disambiguating entities during pipeline execution. Searches involving polysemous terms require extensive graph traversals, leading to either incomplete responses or false positives. Third, temporal information relies on explicit conversion to plain text (which can be added to triplet's attributes as timestamp), creating inefficiencies and potential information loss due to the "Lost in the Middle" problem~\cite{liu-etal-2024-lost}. Finally, duplicate detection relies on exact string comparisons rather than semantic equivalence, resulting in unnecessary replication of synonymous triples that increases storage demands and complicates updates. Future work should explore more advanced ontology learning, semantic deduplication and temporal representation mechanisms.

\textbf{Retrieval Quality.} PAI-2 iterative retrieval mechanism, while effective for multi-hop reasoning, shows variable quality across benchmark types. On single-hop benchmarks (Natural Questions, TriviaQA), PAI-2 achieves comparable, but not superior performance to HippoRAG 2, suggesting the planning mechanism provides limited benefits for simpler queries. Retrieval quality also depends on the initial entity matching stage: incorrect entity links propagate errors through subsequent search steps, potentially leading to irrelevant context retrieval. Additionally, the beam search and water circles traversal algorithms, while effective, may miss relevant vertices in sparse graph regions. The recall-precision trade-off observed in our analysis (Section~\ref{subsec:rq1_comparison}) indicates PAI-2 prioritizes recall over precision, which may introduce noise in answer aggregation. Future work should investigate adaptive retrieval strategies that balance recall and precision based on query complexity.

\textbf{Computational Cost.} PAI-2 has higher computational costs compared to baseline methods, as detailed in Appendix~\ref{app:qapipe_execution_characteristics}. Mean latency is 5.7$\times$ higher than PAI-1 (136s vs. 24s), LLM calls average 42 per query (vs. 2 for PAI-1) and token consumption is 15.4$\times$ higher (39180 vs. 2535 prompt tokens). These costs may prevent deployment in latency-sensitive or resource-constrained environments. While hyperparameters ($S_m$, $C_m$) allow cost configuration, reducing them compromises the iterative reasoning benefits that distinguish PAI-2. The primary bottlenecks (LLM inference, vector search, and graph traversal) require optimization through caching, vector store partitioning and search space constraints. Future work should explore cost-efficient variants, such as early termination mechanisms, query difficulty prediction for adaptive resource allocation and distillation of the planning mechanism into lighter models.


Addressing these limitations represents key directions for future research aimed at improving the scalability, robustness and accessibility of personalized knowledge graph-based QA systems.


\section{Future work}

To address identified limitations, we propose the following enhancements.

\textbf{Thesis Vertex Labeling}. Each thesis vertex will receive dual categorisation: \textit{Episode} and \textit{Temporal}. The \textit{Episode} labeling classifies thesis formulations: FACT -- factual claims verifiable through independent evidence; OPINION -- subjective opinions, requiring contextual interpretation;  PREDICTION -- speculative predictions lacking immediate verification. Meanwhile, \textit{Temporal} labeling specifies the duration over which a statement remains relevant: STATIC -- statically enduring facts; DYNAMIC -- temporally limited assertions expiring upon subsequent developments; ATEMPORAL -- universally applicable truths unaffected by chronology.

\textbf{Time Interval Specification}. We introduce explicit timestamps for each thesis vertex, identifying its creation (t\_created), validity onset (t\_valid), expiration (t\_expired), invalidation (t\_invalid) and potential override by newer information (invalidated\_by). These parameters facilitate precise tracking of knowledge lifecycle stages~\cite{Rasmussen2025ZepAT}.

\textbf{Fixed Predicate Fields}. Text fields in simple triples predicates adopt predefined, time-independent values from verified and periodically updated collection (glossary). Object vertices store additional metadata such as entity types and brief descriptions, while predicates include textual representations specifying relationships between subjects and objects~\cite{zhang-soh-2024-extract}.

These modifications collectively aim to enhance reliability, scalability and usability of the proposed method, thereby mitigating current drawbacks effectively.

\section{Ethics statement}

During the preparation of this manuscript, the authors used GigaChat Max (01.09.26) to improve language, grammar, and overall clarity. After using this tool, the authors reviewed, edited, and verified all suggested changes for scientific accuracy, and take full responsibility for the final content.

\clearpage
\onecolumn
\appendices

\section{Pseudocode}
\label{app:medium_qa_pseudocode}

\begin{algorithm}[H]
\begin{algorithmic}[1]
    \State  \textsf{Input}: $Q$ - user question; $S_m$ - maximum number of search steps in a plan; $C_m$ - maximum number of clue-queries per search step; $V_m$ - maximum number of matched vertices to one entity; $F_m$ - maximum number of triples after filtering per clue-query. 
    \State  \textsf{Output}: $A$ - answer to user question.
    \State $SubQuestions$ $\leftarrow$ Preprocess($Q$) \Comment{$\{q_1, q_2, ..., q_N\}$} 
    \State $SubAnswers \leftarrow$  NewList()
    
    \ForAll{$q_i\in SubQuestions$}
        \State $SearchPlan$ $\leftarrow$ InitialPlanGen($q_i$) \Comment{$[s_1, s_2, ..., s_M]$}, where $0 \leq M \leq S_m$  
        \State $StepsAnswers \leftarrow$  NewList()
        \State $SubAnswerFound \leftarrow False$
        \State $StepNum \leftarrow 1$
        
        \While{$StepNum \leq S_m$}
            \State $StepEntities$ $\leftarrow$ NER($SearchPlan$[$StepNum$]) \Comment{$[e_1, e_2, ..., e_U]$}
        
            \State $MatchedVertices$ $\leftarrow$ Entities2VerticesMatching($StepEntities$, $V_m$) \Comment{$[[v_{11}, v_{12}, ..., v_{1V_m}],...,[v_{U1}, v_{U2}, ..., v_{UV_m}]]_{U \times V_m}$}
            
            \State $AcceptedVerticesLists$ $\leftarrow$ LinearCombination($MatchedVertices$, $C_m$) \Comment{$V_{C_m \times U}$}
            \State $ClueQueries$ $\leftarrow$ ClueQueriesGen($SearchPlan$[$StepNum$], $AcceptedVerticesLists$) \Comment{$[cq_1, cq_2, ..., cq_{C_m}]$}
            \State $ClueAnswers \leftarrow$ NewList()
            
            \ForAll{$cq_j \in ClueQueries$}
                \State $RetrievedTriples$ $\leftarrow$ KGraphTraverse($cq_j$, $AcceptedVerticesLists[j]$) \Comment{$\{t_1, t_2, ..., t_Y\}$} 
                
                \State $FilteredTriples$ $\leftarrow$ FilterByRelevance($RetrievedTriples$) \Comment{$\{t_1, t_2, ..., t_{F_m}\}$} 

                \State $ca_j$ $\leftarrow$ ClueAnswerGen($cq_j$, $FilteredTriples$)
                \State $ClueAnswers \leftarrow ClueAnswers + ca_j$    
            \EndFor
            
            \State $sa_{StepNum}$ $\leftarrow$ SummarizeClueAnswers($SearchPlan$[$StepNum$], $ClueQueries$, $ClueAnswers$)
            \State $StepsAnswers \leftarrow StepsAnswers + sa_{StepNum}$

            \If{Sufficient($q_i$, $SearchPlan$, $StepsAnswers$)}
                \State $SubAnswerFound \leftarrow True$
                \State break
            \Else
                \State $SearchPlan$ $\leftarrow$ SearchPlanEnhance($q_i$, $SearchPlan$, $StepsAnswers$) \Comment{$[s_1, s_2, ..., s_K]$}, where $0 \leq M \leq K \leq S_m$

                \State $StepNum \leftarrow StepNum + 1$ 
                \If{$StepNum$ > $K$}
                    \State break
                \EndIf
            \EndIf
        \EndWhile

        \If{$SubAnswerFound$}
            \State $a_i \leftarrow$ FinalizeSubAnswer($q_i$, $SearchPlan$, $StepsAnswers$)  
        \Else
            \State $a_i$ $\leftarrow$ NoAnswerStubGeneration($q_i$) 
        \EndIf
        \State $SubAnswers \leftarrow SubAnswers + a_i$
    \EndFor
    
    \State $A$ $\leftarrow$ AggregateSubAnswers($Q$, $SubQuestions$, $SubAnswers$)
\end{algorithmic}

\caption{PAI-2 QA pipeline}
\label{alg:medium_pipeline_pseudocode}
\end{algorithm}

\section{PAI-2 execution on example}
\label{app:pai2_example}

\begin{figure*}[th!]
	\centering
	\includegraphics[width=\textwidth]{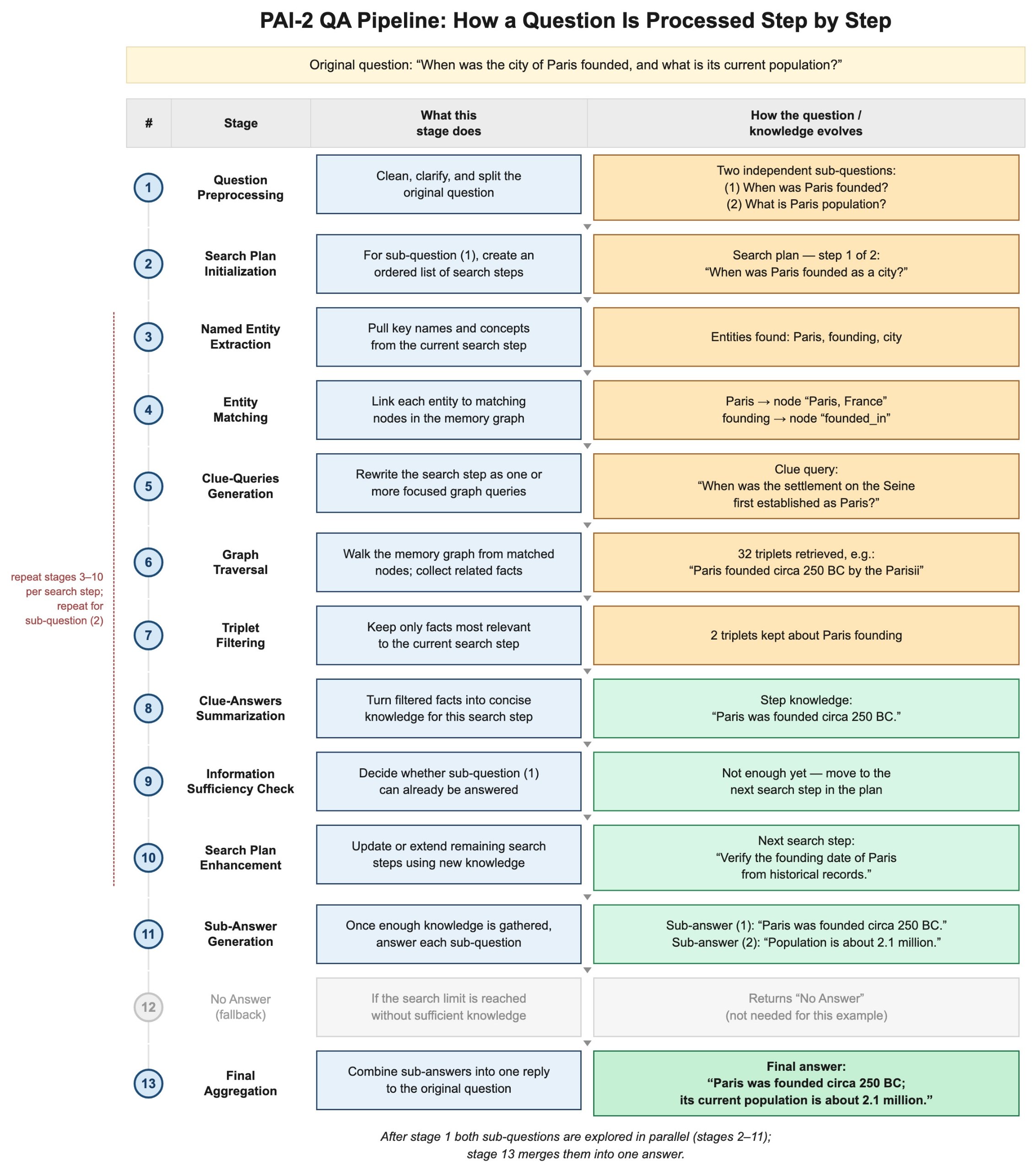}
	\caption{An example of PAI-2's QA pipeline execution for the specific question} 				
	\label{fig:pai2_example}
\end{figure*}

An example of PAI-2's QA pipeline execution for the question "When was the city of Paris founded, and what is its current population?" together with a description of the intermediate-stage results is presented in Figure~\ref{fig:pai2_example}.

\section{Benchmarks preprocessing operations for proposed QA pipeline evaluation}
\label{app:datasets}
For the original Natural Questions benchmark, the "train" subset was selected from HuggingFace repository \footnote{\url{https://huggingface.co/datasets/sentence-transformers/natural-questions}}, comprising $100231$ question-answer (QA) pairs. Important to note that benchmark does not contain documents, associated and relevant to QA pairs. To construct knowledge graphs we use corresponding texts in the "answer" column as relevant documents. Firstly, QA pairs were filtered to exclude those with associated answers falling outside a specified length range (in characters), retaining only between $64$ and $1024$ characters in length. This filtering process resulted in $67174$ remaining QA pairs. Secondly, we extracted the first $2000$ QA pairs. Finally, we expanded prepared documents subset with a $1970$ selected answers, yielding a final subset of $3970$ unique documents for graph construction. For more details you can view Jupyter Notebook\footnote{\url{https://github.com/Dzigen/PersonalAI/blob/master/notebooks/download\_datasets/natural\_questions.ipynb}} with full preprocessing pipeline implementation in the PersonalAI GitHub repository.

For the original TriviaQA benchmark, the "rc.wikipedia/validation" subset was selected from HuggingFace repository\footnote{\url{https://huggingface.co/datasets/mandarjoshi/trivia\_qa}}, comprising $7993$ question-answer (QA) pairs. Given the extensive length of contained documents, they were chunked using the "RecursiveCharacterTextSplitter" class from the LangChain library. The following hyperparameters were applied: (1) a chunk size of $1024$ characters; (2) separators set to double newline characters ("$\backslash$n$\backslash$n"); (3) a chunk overlap of $64$ characters; (4) the "len" function for length calculation, and is\_separator\_regex set to "False". This preprocessing yielded $278390$ unique chunks. Subsequently, QA pairs were discarded if their associated chunks fell outside the specified length bounds (minimum $64$ and maximum $1024$ characters), resulting in $238052$ retained chunks. Additionally, since the original documents were split without explicit tracking of which chunk contains the necessary information to answer associated question, the following filter was applied: if any chunk from a document was discarded, all remaining chunks were also removed to ensure coherence. This step further reduced the corpus to $9975$ unique chunks. Finally, the first $500$ QA pairs were selected, yielding a final subset of 4925 unique chunks for graph construction. For more details you can view Jupyter Notebook\footnote{\url{https://github.com/Dzigen/PersonalAI/blob/master/notebooks/download\_datasets/triviaqa.ipynb}} with full preprocessing pipeline implementation in the GitHub repository of PersonalAI library.

For the original HotpotQA benchmark, the "distractor/validation" subset was selected from HuggingFace repository\footnote{\url{https://huggingface.co/datasets/hotpotqa/hotpot\_qa}}, comprising $7405$ question-answer (QA) pairs and $13781$ unique documents. QA pairs were then filtered to exclude those with associated documents falling outside a specified length range, retaining only documents between $64$ and $1024$ characters. This filtering process resulted in $13291$ remaining documents. Finally, the first $2000$ QA pairs were selected, yielding a final subset of $3933$ unique documents for graph construction. For more details you can view Jupyter Notebook\footnote{\url{https://github.com/Dzigen/PersonalAI/blob/master/notebooks/download\_datasets/hotpotqa.ipynb}} with full preprocessing pipeline implementation in the GitHub repository of PersonalAI library.

For the original 2WikiMultihopQA benchmark, the "dev" subset was selected from GitHub repository\footnote{\url{https://github.com/Alab-NII/2wikimultihop}}, comprising $12576$ question-answer (QA) pairs and $56687$ unique documents. QA pairs were then filtered to exclude those with associated documents falling outside a specified length range, retaining only documents between $64$ and $1024$ characters. This filtering process resulted in $49299$ remaining documents. Finally, the first $2000$ QA pairs were selected, yielding a final subset of $4596$ unique documents for graph construction. For more details you can view Jupyter Notebook\footnote{\url{https://github.com/Dzigen/PersonalAI/blob/master/notebooks/download\_datasets/2wikimultihopqa.ipynb}} with full preprocessing pipeline implementation in the GitHub repository of PersonalAI library.

For the original MuSiQue benchmark, the "validation" subset was selected from HuggingFace repository\footnote{\url{https://huggingface.co/datasets/dgslibisey/MuSiQue}}, comprising $2417$ question-answer (QA) pairs and $21100$ unique documents. QA pairs were then filtered to exclude those with associated documents falling outside a specified length range, retaining only contexts between $64$ and $1024$ characters. This filtering process resulted in $19867$ remaining documents. Second, the first $1931$ QA pairs were selected. Finally, we expanded prepared subset with $2000$ selected documents, yielding a final subset of 4185 unique documents for graph construction. For more details you can view Jupyter Notebook\footnote{\url{https://github.com/Dzigen/PersonalAI/blob/master/notebooks/download\_datasets/musique.ipynb}} with full preprocessing pipeline implementation in the GitHub repository of PersonalAI library.


Thus, evaluation sets were obtained: see preprocesssing summary in Table~\ref{tab:bench_preprocessing_summary}.

\begin{table}[H]
	\centering
	\renewcommand{\arraystretch}{1.3}
    \caption{Summary of benchmarks preprocessing operations}
    \resizebox{\textwidth}{!}{%
    \begin{tabular}{|c|c|c|c|c|c|c|}
    \hline
    \textbf{Benchmark} & \textbf{Split} & \begin{tabular}[c]{@{}c@{}}\textbf{Documents Chunking} \\ \textbf{Rules}\end{tabular} & \begin{tabular}[c]{@{}c@{}}\textbf{Documents Filtering} \\ \textbf{Rules}\end{tabular} & \begin{tabular}[c]{@{}c@{}}\textbf{Question-Answer Pairs}\\ \textbf{Selection Rules}\end{tabular} & \begin{tabular}[c]{@{}c@{}}\textbf{Number of} \\ \textbf{Supporting / Non-Supporting} \\ \textbf{Documents}\end{tabular} & \begin{tabular}[c]{@{}c@{}}\textbf{Evidence Type} \\ \textbf{(to store in memory)}\end{tabular} \\ 
    \hhline{=======}
    Natural Questions & train & none & \multirow{10}{*}{\begin{tabular}[c]{@{}c@{}}(1) unique documents,\\ (2) length range (in characters) \\ between 64 and 1024\end{tabular}} & \begin{tabular}[c]{@{}c@{}}(1) relevant documents\\ remained after filtering,\\ (2) first 2000 pairs\end{tabular} & 2000 / 1970 & \begin{tabular}[c]{@{}c@{}}golden\\ answer\end{tabular} \\ 
    \cline{1-3} \cline{5-7} 
    TriviaQA & rc.wikipedia/validation" & \begin{tabular}[c]{@{}c@{}}(1) chunk size = 1024 characters, \\ (2) separators = "\textbackslash{}n\textbackslash{}n",\\ (3) chunk\_overlap = 64 characters\end{tabular} &  & \begin{tabular}[c]{@{}c@{}}(1) relevant documents\\ remained after filtering,\\ (2) first 500 pairs\end{tabular} & 4925 / 0 & \begin{tabular}[c]{@{}c@{}}relevant\\ documents\end{tabular} \\ 
    \cline{1-3} \cline{5-7} 
    HotpotQA & distractor/validation & none &  & \begin{tabular}[c]{@{}c@{}}(1) relevant documents\\ remained after filtering,\\ (2) first 2000 pairs\end{tabular} & 3933 / 0 & \begin{tabular}[c]{@{}c@{}}relevant\\ documents\end{tabular} \\ 
    \cline{1-3} \cline{5-7} 
    2WikiMultihopQA & dev & none &  & \begin{tabular}[c]{@{}c@{}}(1) relevant documents\\ remained after filtering,\\ (2) first 2000 pairs\end{tabular} & 4596 / 0 & \begin{tabular}[c]{@{}c@{}}relevent\\ documents\end{tabular} \\ 
    \cline{1-3} \cline{5-7} 
    MuSiQue & validation & none &  & \begin{tabular}[c]{@{}c@{}}(1) relevant documents\\ remained after filtering,\\ (2) first 1931 pairs\end{tabular} & 2185 / 2000 & \begin{tabular}[c]{@{}c@{}}relevant\\ documents\end{tabular} \\ 
    \hline
    \end{tabular}
    }
	\label{tab:bench_preprocessing_summary}
\end{table}

The characteristics of Natural Questions, TriviaQA, HotpotQA, 2WikiMultihopQA and MuSiQue subsets can be found in Table~\ref{tab:datasets_stats}. Resulted subsets of original benchmarks are published and available in our HuggingFace repository\footnote{\url{https://huggingface.co/datasets/dzigen/PAI-2_PreprocessedBenchmarks}}.

\begin{table}[H]
	\centering
	\renewcommand{\arraystretch}{1.5}
    \caption{Benchmarks characteristics used for PAI-2 and baselines evaluation}
		\begin{tabular}{|c|ccccccc|cccc|}
        \hline
        \multirow{4}{*}{\textbf{Benchmark}} & \multicolumn{7}{c|}{\textbf{QA-Pairs}} & \multicolumn{4}{c|}{\textbf{Relevant Documents}} \\ \cline{2-12} 
         & \multicolumn{1}{c|}{\multirow{2}{*}{\textbf{Amount}}} & \multicolumn{3}{c|}{\begin{tabular}[c]{@{}c@{}}\textbf{Questions Length}\\ \textbf{(in characters)}\end{tabular}} & \multicolumn{3}{c|}{\begin{tabular}[c]{@{}c@{}}\textbf{Answers Length}\\ \textbf{(in characters)}\end{tabular}} & \multicolumn{1}{c|}{\multirow{2}{*}{\textbf{Amount}}} & \multicolumn{3}{c|}{\begin{tabular}[c]{@{}c@{}}\textbf{Length}\\ \textbf{(in characters)}\end{tabular}} \\ \cline{3-8} \cline{10-12} 
         & \multicolumn{1}{c|}{} & \multicolumn{1}{c|}{\textbf{median}} & \multicolumn{1}{c|}{\textbf{mean}} & \multicolumn{1}{c|}{\textbf{std}} & \multicolumn{1}{c|}{\textbf{median}} & \multicolumn{1}{c|}{\textbf{mean}} & \textbf{std} & \multicolumn{1}{c|}{} & \multicolumn{1}{c|}{\textbf{median}} & \multicolumn{1}{c|}{\textbf{mean}} & \textbf{std} \\ \hline \hline
        Natural Questions & \multicolumn{1}{c|}{2000} & \multicolumn{1}{c|}{44} & \multicolumn{1}{c|}{47} & \multicolumn{1}{c|}{11} & \multicolumn{1}{c|}{515} & \multicolumn{1}{c|}{534} & 218 & \multicolumn{1}{c|}{3970} & \multicolumn{1}{c|}{522} & \multicolumn{1}{c|}{536} & 220 \\ \hline
        TriviaQA & \multicolumn{1}{c|}{500} & \multicolumn{1}{c|}{66} & \multicolumn{1}{c|}{76} & \multicolumn{1}{c|}{39} & \multicolumn{1}{c|}{9} & \multicolumn{1}{c|}{10} & 6 & \multicolumn{1}{c|}{4925} & \multicolumn{1}{c|}{807} & \multicolumn{1}{c|}{765} & 196 \\ \hline
        HotpotQA & \multicolumn{1}{c|}{2000} & \multicolumn{1}{c|}{87} & \multicolumn{1}{c|}{93} & \multicolumn{1}{c|}{33} & \multicolumn{1}{c|}{13} & \multicolumn{1}{c|}{15} & 12 & \multicolumn{1}{c|}{3933} & \multicolumn{1}{c|}{384} & \multicolumn{1}{c|}{414} & 201 \\ \hline
        2WikiMultihopQA & \multicolumn{1}{c|}{2000} & \multicolumn{1}{c|}{69} & \multicolumn{1}{c|}{70} & \multicolumn{1}{c|}{17} & \multicolumn{1}{c|}{13} & \multicolumn{1}{c|}{14} & 9 & \multicolumn{1}{c|}{4596} & \multicolumn{1}{c|}{300} & \multicolumn{1}{c|}{362} & 227 \\ \hline
        MuSiQue & \multicolumn{1}{c|}{1931} & \multicolumn{1}{c|}{89} & \multicolumn{1}{c|}{96} & \multicolumn{1}{c|}{37} & \multicolumn{1}{c|}{14} & \multicolumn{1}{c|}{17} & 13 & \multicolumn{1}{c|}{4185} & \multicolumn{1}{c|}{384} & \multicolumn{1}{c|}{426} & 216 \\ \hline \hline
        Mean & \multicolumn{1}{c|}{1686} & \multicolumn{1}{c|}{71} & \multicolumn{1}{c|}{76} & \multicolumn{1}{c|}{27} & \multicolumn{1}{c|}{113} & \multicolumn{1}{c|}{118} & 52 & \multicolumn{1}{c|}{4322} & \multicolumn{1}{c|}{479} & \multicolumn{1}{c|}{501} & 212 \\ \hline
		\end{tabular}
	\renewcommand{\arraystretch}{1}
	\label{tab:datasets_stats}
\end{table}

Table~\ref{tab:datasets_stats} shows that longest and shortest questions (on average) belong to MuSiQue and NaturalQuestions subsets, respectively: $96$ and $47$ characters. At the same time, TriviaQA and NaturalQuestions contain shortest and longest (on average) answers, respectively: $10$ and $534$ characters. Also, longest and shortest relevant documents (on average) belong to TriviaQA and 2WikiMultihopQA subsets, respectively: $765$ and $362$ characters. In addition, most and fewest amount of relevant documents belong to TriviaQA and HotpotQA subsets, respectively: $4925$ and $3933$ documents.

\section{Specified hyperparameters for graph traversal / triples retrieval methods}
\label{app:retriev_hyper}

\begin{itemize}
    \item \textbf{BeamSearch}: 
    \begin{itemize}
        \item main\_hyperparams: max\_depth = $3$, max\_paths = $18$, same\_path\_intersection\_by\_node = $False$, \\diff\_paths\_intersection\_by\_node = $False$, diff\_paths\_intersection\_by\_rel = $False$, \\mean\_alpha = $0.75$, final\_sorting\_mode = "mixed";
        \item reranker\_method = "single\_step";
        \item reranker\_config: vdb\_name = "dense\_triplets", threshold = $0.5$, fetch\_n = $50$.
    \end{itemize}
    
    \item \textbf{WaterCircles}: 
    \begin{itemize}
        \item main\_hyperparams: strict\_filter = $True$, hyper\_num = $15$, episodic\_num = $15$, chain\_triplets\_num = $25$, other\_triplets\_num = $6$, do\_text\_pruning = $False$.
    \end{itemize}
    
        
    \item \textbf{NaiveRetriever}: 
    \begin{itemize}
        \item main\_hyperparams: max\_k = $50$; 
        \item reranker\_method = "single\_step";
        \item reranker\_config: vdb\_name = "dense\_triplets", threshold = $0.5$, fetch\_n = $50$.
    \end{itemize}
\end{itemize}

\section{Characteristics of constructed memory graphs}
\label{app:constr_graphs}

To evaluate PAI-2's QA pipeline we constructed five memory graphs based on selected benchmarks and using Qwen2.5 7B. The structural characteristics of constructed graphs are detailed in Table~\ref{tab:kg_stats}. 
    
\begin{table}[H]
	\centering
	\renewcommand{\arraystretch}{1.5}
    \caption{Characteristics of constructed (with Qwen2.5 7B) memory graphs on given benchmarks for PAI-2 evaluation}
	\resizebox{\textwidth}{!}{%
    \begin{tabular}{|c|c|ccc|ccc|cccc|}
    \hline
    \multirow{3}{*}{\textbf{Benchmark}} & \multirow{3}{*}{\begin{tabular}[c]{@{}c@{}}\textbf{Number of documents} \\ \textbf{to store in graph}\end{tabular}} & \multicolumn{3}{c|}{\textbf{Number of vertices}} & \multicolumn{3}{c|}{\textbf{Number of edges}} & \multicolumn{4}{c|}{\textbf{Mean / Std of neighbor vertices (by type)}} \\ \cline{3-12} 
     &  & \multicolumn{1}{c|}{\textbf{episodic}} & \multicolumn{1}{c|}{\textbf{thesis}} & \textbf{object} & \multicolumn{1}{c|}{\begin{tabular}[c]{@{}c@{}}\textbf{hyper} \\ \textbf{(to episodic)}\end{tabular}} & \multicolumn{1}{c|}{\begin{tabular}[c]{@{}c@{}}\textbf{hyper} \\ \textbf{(to thesis)}\end{tabular}} & \begin{tabular}[c]{@{}c@{}}\textbf{simple} \\ \textbf{(between objects)}\end{tabular} & \multicolumn{1}{c|}{\begin{tabular}[c]{@{}c@{}}\textbf{object neighbours}\\ \textbf{(to episodic vertices)}\end{tabular}} & \multicolumn{1}{c|}{\begin{tabular}[c]{@{}c@{}}\textbf{object neighbours}\\ \textbf{(to thesis vertices)}\end{tabular}} & \multicolumn{1}{c|}{\begin{tabular}[c]{@{}c@{}}\textbf{object neighbours}\\ \textbf{(to object vertices)}\end{tabular}} & \begin{tabular}[c]{@{}c@{}}\textbf{thesis neighbours}\\ \textbf{(to episodic vertices)}\end{tabular} \\ \hline \hline
    Natural Questions & 3970 & \multicolumn{1}{c|}{3970} & \multicolumn{1}{c|}{32652} & 67104 & \multicolumn{1}{c|}{131935} & \multicolumn{1}{c|}{114083} & 37377 & \multicolumn{1}{c|}{24.97 / 10.48} & \multicolumn{1}{c|}{3.49 / 1.30} & \multicolumn{1}{c|}{1.26 / 1.80} & 8.31 / 3.34 \\ \hline
    TriviaQA & 4925 & \multicolumn{1}{c|}{4921} & \multicolumn{1}{c|}{53079} & 106727 & \multicolumn{1}{c|}{221133} & \multicolumn{1}{c|}{187901} & 61848 & \multicolumn{1}{c|}{34.15 / 11.35} & \multicolumn{1}{c|}{3.54 / 1.37} & \multicolumn{1}{c|}{1.30 / 1.92} & 10.9 / 4.47 \\ \hline
    HotpotQA & 3933 & \multicolumn{1}{c|}{3933} & \multicolumn{1}{c|}{31653} & 56178 & \multicolumn{1}{c|}{119913} & \multicolumn{1}{c|}{105978} & 38644 & \multicolumn{1}{c|}{22.44 / 10.19} & \multicolumn{1}{c|}{3.35 / 1.17} & \multicolumn{1}{c|}{1.37 / 4.46} & 8.12 / 3.54 \\ \hline
    2WikiMultihopQA & 4596 & \multicolumn{1}{c|}{4596} & \multicolumn{1}{c|}{34868} & 54961 & \multicolumn{1}{c|}{135657} & \multicolumn{1}{c|}{120111} & 45715 & \multicolumn{1}{c|}{21.86 / 11.27} & \multicolumn{1}{c|}{3.44 / 1.18} & \multicolumn{1}{c|}{1.51 / 6.47} & 7.70 / 3.59 \\ \hline
    MuSiQue & 4185 & \multicolumn{1}{c|}{4184} & \multicolumn{1}{c|}{32062} & 61024 & \multicolumn{1}{c|}{125308} & \multicolumn{1}{c|}{108710} & 37663 & \multicolumn{1}{c|}{22.24 / 10.49} & \multicolumn{1}{c|}{3.39 / 1.16} & \multicolumn{1}{c|}{1.32 / 2.10} & 7.79 / 3.63 \\ \hline \hline
    Mean & 4322 & \multicolumn{1}{c|}{4321} & \multicolumn{1}{c|}{36863} & 69199 & \multicolumn{1}{c|}{146789} & \multicolumn{1}{c|}{127357} & 44249 & \multicolumn{1}{c|}{25.13 / 10.76} & \multicolumn{1}{c|}{3.44 / 1.24} & \multicolumn{1}{c|}{1.35 / 3.35} & 8.56 / 3.71 \\ \hline
    \end{tabular}}
	\renewcommand{\arraystretch}{1}
	\label{tab:kg_stats}
\end{table}

From Table~\ref{tab:kg_stats} it can be observed that during construction of several memory graphs LLM parsing errors occurred, resulting in loss of some documents and minor incompleteness of generated knowledge graph. Across selected benchmarks, average parsing error rates were the following: TriviaQA -- $0.08\%$; MuSiQue -- $0.02\%$; NaturalQuestions/HotpotQA/2WikiMultihopQA -- $0.0\%$. Nevertheless, as expected, the memory graph that was built on TriviaQA contains the largest number of vertices and edges due to the largest average document length and its amount. It is also worth to note that the TriviaQA-based memory graph contains 20 k more thesis vertices than the other graphs, which contain approximately 30 k vertices each. Despite the differing numbers and lengths of documents across the selected benchmarks, the constructed graphs have roughly the same average number of object vertices adjacent to thesis vertices ($\approx$3.44).  In general, as document length increases, the amount of extracted information stored in the memory graph also grows: for both thesis memories and named entities.


In addition to characteristics of constructed memory graphs, we collected information about time and speed of Memorize pipeline, responsible for parsing incoming documents and storing them in memory graph, and required number of input (prompt) and output (completion) LLM tokens: see Table~\ref{tab:mempipe_performance} and Table~\ref{tab:llmtokens_stat}.

\begin{table}[H]
	\centering
	\renewcommand{\arraystretch}{1.3}
    \caption{Time (hours), speed (documents per minute) of PAI's memory graph construction algorithm (with Qwen2.5 7B) and required disk space for constructed memory graphs (GB) on five benchmarks}
    \begin{tabular}{|c|c|c|c|c|c||c|}
    \hhline{------||-}
    \multirow{2}{*}{\begin{tabular}[c]{@{}c@{}}\textbf{Memory graph} \\ \textbf{characteristic}\end{tabular}} & \multicolumn{5}{c||}{\textbf{Benchmark}} & \multirow{2}{*}{Mean} \\ 
    \hhline{|~|-----||~|}
     & \textbf{Natural Questions} & \textbf{TriviaQA} & \textbf{HotpotQA} & \textbf{2WikiMultihopQA} & \textbf{MuSiQue} &  \\ 
     \hhline{======::=}
    \begin{tabular}[c]{@{}c@{}}Construction time \\ (hours)\end{tabular} & 36 & 86 & 35 & 39 & 39 & 47 \\ 
    \hhline{------||-}
    \begin{tabular}[c]{@{}c@{}}Construction speed \\ (doc. per min)\end{tabular} & 1.81 & 0.96 & 1.87 & 1.94 & 1.79 & 1.67 \\ 
    \hhline{------||-}
    \begin{tabular}[c]{@{}c@{}}Required disk space \\ (GB)\end{tabular} & 2.0 & 2.7 & 1.7 & 2.0 & 1.7 & 2.0 \\ \hhline{------||-}
    \end{tabular}
	\renewcommand{\arraystretch}{1}
	\label{tab:mempipe_performance}
\end{table}

\begin{table}[H]
	\centering
	\renewcommand{\arraystretch}{1.3}
    \caption{LLM tokens amount (in millions) that were spend during memory graph construction (with Qwen2.5 7B) for given benchmarks}
    \begin{tabular}{|cc|c|c|c|c|cc}
    \hhline{|-|-|-----||-}
    \multicolumn{1}{|c|}{\multirow{2}{*}{\textbf{LLM Task}}} & \multirow{2}{*}{\begin{tabular}[c]{@{}c@{}}\textbf{Tokens} \\ \textbf{Category}\end{tabular}} & \multicolumn{5}{c||}{\textbf{Benchmark}} & \multicolumn{1}{c|}{\multirow{2}{*}{Mean}} \\ 
    \hhline{|~|~|-----||~|}
    \multicolumn{1}{|c|}{} &  & \textbf{Natural Questions} & \textbf{TriviaQA} & \textbf{HotpotQA} & \textbf{2WikiMultihopQA} & \multicolumn{1}{c||}{\textbf{MuSiQue}} & \multicolumn{1}{c|}{} \\ 
    \hhline{=======::=}
    \multicolumn{1}{|c|}{\multirow{2}{*}{\begin{tabular}[c]{@{}c@{}}Thesis triples \\ generaion\end{tabular}}} & prompt & 2.6 & 3.6 & 2.6 & 3.0 & \multicolumn{1}{c||}{2.8} & \multicolumn{1}{c|}{2.9} \\ 
    \hhline{|~------||-}
    \multicolumn{1}{|c|}{} & completion & 1.1 & 1.8 & 1.0 & 1.2 & \multicolumn{1}{c||}{1.1} & \multicolumn{1}{c|}{1.2} \\ 
    \hhline{|-------||-}
    \multicolumn{1}{|c|}{\multirow{2}{*}{\begin{tabular}[c]{@{}c@{}}Simple triples \\ generaion\end{tabular}}} & prompt & 2.7 & 3.7 & 2.6 & 3.0 & \multicolumn{1}{c||}{2.8} & \multicolumn{1}{c|}{3.0} \\ 
    \hhline{|~------||-}
    \multicolumn{1}{|c|}{} & completion & 0.5 & 0.9 & 0.5 & 0.6 & \multicolumn{1}{c||}{0.5} & \multicolumn{1}{c|}{0.6} \\ 
    \hhline{=======::=}
    \multicolumn{2}{|c|}{Sum} & 6.9 & 10 & 6.7 & 7.8 & \multicolumn{1}{c||}{7.2} & \multicolumn{1}{c|}{7.7} \\ \hhline{|--|-|-|-|-|-||-|}
    \end{tabular}
	\renewcommand{\arraystretch}{1}
	\label{tab:llmtokens_stat}
\end{table}

From Table~\ref{tab:mempipe_performance} and Table~\ref{tab:llmtokens_stat} we observe that storing 4322 documents with an average length of 501 characters in a memory graph requires $\approx$ 7.7 M tokens, $\approx$ 47 hours, and $\approx$ 2 GB of disk space.

\section{LLM--as--a--Judge instructions}
\label{app:judgescore_hyperp}

To ensure the reproducibility of the obtained results, LLM inference was conducted using a deterministic generation strategy. The following hyperparameters were applied: num\_predict = $2048$, seed = $42$, temperature = $0.0$, and top\_k = $1$. The Qwen2.5 7B, sourced from the Ollama repository\footnote{\url{https://ollama.com/library/qwen2.5:7b}}, was prompted to evaluate whether the response of the method correctly answered given question. LLM prompts that were used for this assessment are provided in Table~\ref{tab:prompts_judge}.

\begin{table}[H]
	\renewcommand{\arraystretch}{1.5}
	\centering
    \caption{LLM prompts for LLM-as-a-Judge framework}
	\resizebox{\textwidth}{!}{%
		\begin{tabular}{|c|l|}
			\hline
			\textbf{Type} & \multicolumn{1}{c|}{\textbf{Prompt}} \\ \hline \hline
			\multirow{1}{*}{System} & 
			\begin{minipage}{\textwidth}
				\includegraphics[clip,trim={.02\textwidth} {0.73\textheight} {.03\textwidth} 0mm, width=\textwidth,valign=b]{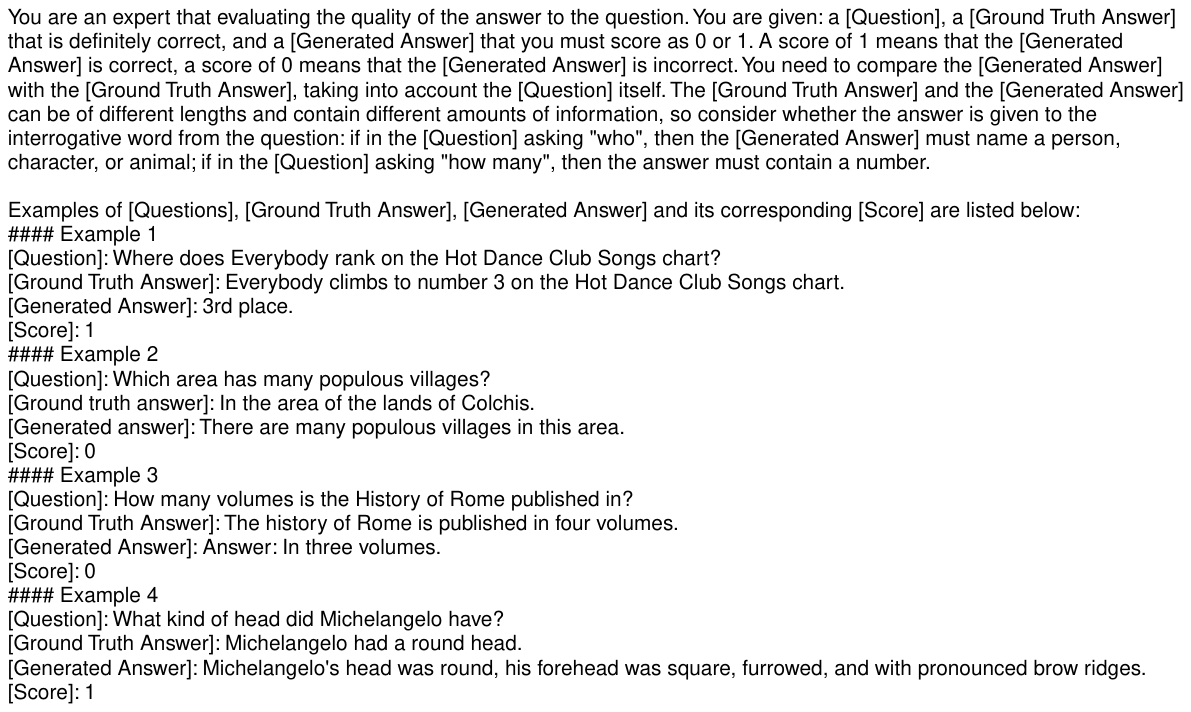}
			\end{minipage}
			\\ \hline
			
			\multirow{1}{*}{User} & 
			\begin{minipage}{\textwidth}
				\includegraphics[clip,trim={.02\textwidth} {1.18\textheight} {.03\textwidth} 0mm, width=\textwidth,valign=b]{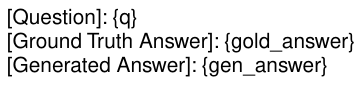}
			\end{minipage}
			\\ \hline
			
			\multirow{1}{*}{Assistant} & 
			\begin{minipage}{\textwidth}
				\includegraphics[clip,trim={.02\textwidth} {1.2\textheight} {.03\textwidth} 0mm, width=\textwidth,valign=b]{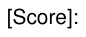}
			\end{minipage}
			\\ \hline
		\end{tabular}%
	}
	\label{tab:prompts_judge}
\end{table}

\section{Human Evaluation}
\label{app:human_evaluation}

Since LLM-as-a-Judge was used as the primary metric to assess accuracy, we perform a verification stage to be confident that there is a strong correlation between LLM and human decisions and that the resulting scores are statistically significant. For each of the two systems (HippoRAG 2 and PAI-2 with and without question preprocessing) and each of the five benchmarks, we sampled 100 questions uniformly at random seed 42 from the RQ1 evaluation set; the same questions were used for all systems, giving 1500 annotated answers in total. Each answer was independently labeled by all three assessors as correct (1) or incorrect (0), given the question, the reference answer and the generated answer, following the instruction from Table~\ref{tab:prompts_judge}. Assessors were blinded to the system identity, and answers were presented in random order. Labels were aggregated by majority vote. All assessors held master's degrees and had prior experience in evaluation of LLM responses. Inter-annotator agreement was measured with Krippendorff's alpha (nominal level, computed per system-benchmark cell over 100 items) and can be seen in Tables \ref{tab:humaneval_kripalpha}.

\begin{table}[H]
    \centering
	\renewcommand{\arraystretch}{1.3}
    \caption{Krippendorff's alpha coefficients of HumanEval scores, calculated for best HippoRAG 2 and PAI-2 configurations across five benchmarks}
    \begin{tabular}{|c|c|c|c|c|c||c|}
    \hhline{|------||-|}
    \multirow{2}{*}{\textbf{Method}} & \multicolumn{5}{c||}{\textbf{Benchmark}} & \multirow{2}{*}{Mean} \\ \hhline{|~|-----||~|}
     & \textbf{Natural Questions} & \textbf{TriviaQA} & \textbf{HotpotQA} & \textbf{2WikiMultihopQA} & \textbf{MuSiQue} &  \\ 
    \hhline{|======::=}
    HippoRAG 2 & 0.93 & 0.92 & 0.93 & 0.92 & 0.97 & 0.93 \\ 
    \hhline{|------||-|}
   \multicolumn{1}{|c|}{\begin{tabular}[c]{@{}c@{}}PAI-2 (w/o question \\ preprocessing)\end{tabular}} & 0.86 & 0.95 & 0.95 & 0.95 & 0.91 & 0.92 \\
    \hhline{|------||-|}
    \multicolumn{1}{|c|}{\begin{tabular}[c]{@{}c@{}}PAI-2 (w question \\ preprocessing)\end{tabular}} & 0.91 & 0.93 & 0.96 & 0.97 & 0.94 & 0.94\\
    \hhline{|------||-|}
    \end{tabular}
    \renewcommand{\arraystretch}{1}
	\label{tab:humaneval_kripalpha}
\end{table}

All Krippendorff's alpha values from Table~\ref{tab:humaneval_kripalpha} exceed 0.86, with a mean of 0.93 across all configurations. According to standard interpretation guidelines, values above 0.80 indicate strong agreement, while values above 0.90 suggest excellent reliability. The highest agreement was observed for HippoRAG 2 on MuSiQue (0.97) and PAI-2 with question preprocessing on 2WikiMultihopQA (0.97). The lowest value (0.86 for PAI-2 without question preprocessing on Natural Questions) still indicates strong inter-annotator consistency. These results confirm that our human evaluators applied consistent criteria across all benchmarks and methods, establishing a reliable ground truth for subsequent comparison with the LLM judge.

Agreement between LLM judge and human majority vote is reported as raw agreement, Pearson correlation and Cohen's kappa: see Table~\ref{tab:humaneval_vs_judge_correlations}.

\begin{table*}[ht!]
    \centering
	\renewcommand{\arraystretch}{1.3}
    \caption{Raw agreement, Pearson correlation and Cohen's kappa between LLM-as-a-Judge and HumanEval scores, calculated for HippoRAG 2 and PAI-2 best configurations across five benchmarks}
    \begin{tabular}{|cccccc||c|}
    \hhline{------||-}
    \multicolumn{1}{|c|}{\multirow{2}{*}{\textbf{Method}}} & \multicolumn{5}{c||}{\textbf{Benchmark}} & \multirow{2}{*}{Mean} \\ 
    \hhline{~-----||~}
    \multicolumn{1}{|c|}{} & \multicolumn{1}{c|}{\textbf{Natural Questions}}  & \multicolumn{1}{c|}{\textbf{TriviaQA}} & \multicolumn{1}{c|}{\textbf{HotpotQA}} & \multicolumn{1}{c|}{\textbf{2WikiMultihopQA}} & \multicolumn{1}{c||}{\textbf{MuSiQue}} &  \\
    \hhline{======::=}
    \multicolumn{7}{|c|}{\textbf{Raw agreement}} \\ 
    \hline
    \multicolumn{1}{|c|}{HippoRAG 2} & \multicolumn{1}{c|}{0.92} & \multicolumn{1}{c|}{0.98} & \multicolumn{1}{c|}{0.94} & \multicolumn{1}{c|}{0.91} & \multicolumn{1}{c||}{0.92} & 0.93 \\ 
    \hhline{------||-}
    \multicolumn{1}{|c|}{\begin{tabular}[c]{@{}c@{}}PAI-2 (w/o question \\ preprocessing)\end{tabular}} & \multicolumn{1}{c|}{0.96} & \multicolumn{1}{c|}{0.95} & \multicolumn{1}{c|}{0.94} & \multicolumn{1}{c|}{0.97} & \multicolumn{1}{c||}{0.97} & 0.96 \\ 
    \hhline{------||-}
    \multicolumn{1}{|c|}{\begin{tabular}[c]{@{}c@{}}PAI-2 (w question \\ preprocessing)\end{tabular}} & \multicolumn{1}{c|}{0.98} & \multicolumn{1}{c|}{0.96} & \multicolumn{1}{c|}{0.97} & \multicolumn{1}{c|}{0.98} & \multicolumn{1}{c||}{0.96} & 0.97 \\
    \hhline{======::=}
    \multicolumn{7}{|c|}{\textbf{Pearson correlation coefficient}} \\ 
    \hline
    \multicolumn{1}{|c|}{HippoRAG 2} & \multicolumn{1}{c|}{0.74} & \multicolumn{1}{c|}{0.94} & \multicolumn{1}{c|}{0.84} & \multicolumn{1}{c|}{0.82} & \multicolumn{1}{c||}{0.82} & 0.83 \\ 
    \hhline{------||-}
    \multicolumn{1}{|c|}{\begin{tabular}[c]{@{}c@{}}PAI-2 (w/o question \\ preprocessing)\end{tabular}} & \multicolumn{1}{c|}{0.90} & \multicolumn{1}{c|}{0.83} & \multicolumn{1}{c|}{0.84} & \multicolumn{1}{c|}{0.94} & \multicolumn{1}{c||}{0.94} & 0.89 \\ 
    \hhline{------||-}
    \multicolumn{1}{|c|}{\begin{tabular}[c]{@{}c@{}}PAI-2 (w question \\ preprocessing)\end{tabular}} & \multicolumn{1}{c|}{0.95} & \multicolumn{1}{c|}{0.88} & \multicolumn{1}{c|}{0.94} & \multicolumn{1}{c|}{0.96} & \multicolumn{1}{c||}{0.92} & 0.93 \\ 
    \hhline{======::=}
    \multicolumn{7}{|c|}{\textbf{Cohen's kappa}} \\ 
    \hline
    \multicolumn{1}{|c|}{HippoRAG 2} & \multicolumn{1}{c|}{0.73} & \multicolumn{1}{c|}{0.94} & \multicolumn{1}{c|}{0.84} & \multicolumn{1}{c|}{0.82} & \multicolumn{1}{c||}{0.81} & 0.83 \\ 
    \hhline{------||-}
    \multicolumn{1}{|c|}{\begin{tabular}[c]{@{}c@{}}PAI-2 (w/o question \\ preprocessing)\end{tabular}} & \multicolumn{1}{c|}{0.89} & \multicolumn{1}{c|}{0.83} & \multicolumn{1}{c|}{0.83} & \multicolumn{1}{c|}{0.94} & \multicolumn{1}{c||}{0.94} & 0.89 \\ 
    \hhline{------||-}
    \multicolumn{1}{|c|}{\begin{tabular}[c]{@{}c@{}}PAI-2 (w question \\ preprocessing)\end{tabular}} & \multicolumn{1}{c|}{0.95} & \multicolumn{1}{c|}{0.88} & \multicolumn{1}{c|}{0.94} & \multicolumn{1}{c|}{0.96} & \multicolumn{1}{c||}{0.92} & 0.93 \\ 
    \hhline{------||-}
    \end{tabular}
    \renewcommand{\arraystretch}{1}
	\label{tab:humaneval_vs_judge_correlations}
\end{table*}


The LLM judge achieves 91-98\% raw agreement with human majority votes across all configurations and benchmarks. Correlation coefficients range from 0.74 to 0.96, indicating strong to very strong linear relationships between automated and human scores. The lowest correlation (0.74 for HippoRAG 2 on Natural Questions) still represents substantial agreement, while the highest values (0.95-0.96 for PAI-2 with question preprocessing on Natural Questions and 2WikiMultihopQA) indicate near-perfect alignment. Mean correlations improve from 0.83 (HippoRAG 2) to 0.89-0.93 (PAI-2 variants). Cohen's kappa agreement metric shows values from 0.73 to 0.96. According to Landis and Koch's benchmark, all values exceed 0.60 (substantial agreement), with most exceeding 0.80 (almost perfect agreement). The pattern mirrors Pearson correlation, with PAI-2 configurations demonstrating stronger alignment with human judgments than HippoRAG 2.

Table~\ref{tab:humaneval_vs_judgescores} presents the direct comparison of accuracy scores between the two evaluation methods. The LLM-as-a-Judge consistently produces slightly lower scores than HumanEval (mean difference: 0.02-0.03), indicating a conservative bias. This pattern holds across all methods and benchmarks, suggesting the automated judge applies stricter evaluation criteria than human annotators. Importantly, the relative ranking of systems remains consistent between both evaluation methods, preserving the validity of comparative conclusions.

\begin{table}[H]
    \centering
	\renewcommand{\arraystretch}{1.3}
    \caption{HumanEval and LLM-as-a-Judge scores for HippoRAG 2 and PAI-2 best configurations across five benchmarks}
    \begin{tabular}{|c|c|c|c|c|c|c||c|}
    \hhline{|-|-|-----||-|}
    \multirow{2}{*}{\textbf{Method}} & \multirow{2}{*}{\textbf{Metric}} & \multicolumn{5}{c||}{\textbf{Benchmark}} & \multirow{2}{*}{Mean} \\ 
    \hhline{|~|~|-----||~|}
     &  & \textbf{Natural Questions} & \textbf{TriviaQA} & \textbf{HotpotQA} & \textbf{2WikiMultihopQA} & \textbf{MuSiQue} &  \\ 
    \hhline{=======::=}
    \multirow{2}{*}{HippoRAG 2} & HumanEval & 0.84 & 0.79 & 0.75 & 0.53 & 0.33 & 0.65 \\
    \hhline{|~|-|-|-|-|-|-||-|}
     & LLM-as-a-Judge & 0.80 & 0.77 & 0.73 & 0.56 & 0.29 & 0.63 \\
     \hhline{=======:|=}
    \multirow{2}{*}{\begin{tabular}[c]{@{}c@{}}PAI-2 (w/o question\\preprocessing)\end{tabular}} & HumanEval & 0.75 & 0.82 & 0.81 & 0.64 & 0.50 & 0.70 \\ \hhline{|~|-|-|-|-|-|-||-|}
     & LLM-as-a-Judge & 0.71 & 0.81 & 0.75 & 0.61 & 0.47 & 0.67 \\ 
     \hhline{=======:|=}
    \multirow{2}{*}{\begin{tabular}[c]{@{}c@{}}PAI-2 (w question\\preprocessing) \end{tabular}} & HumanEval & 0.70 & 0.81 & 0.65 & 0.65 & 0.48 & 0.66 \\ \hhline{|~|-|-|-|-|-|-||-|}
     & LLM-as-a-Judge & 0.68 & 0.79 & 0.62 & 0.63 & 0.44 & 0.63 \\ 
     \hhline{|-|-|-|-|-|-|-||-|}
    \end{tabular}
    \renewcommand{\arraystretch}{1}
	\label{tab:humaneval_vs_judgescores}
\end{table}


To formally test whether differences between LLM-as-a-Judge and HumanEval are statistically significant, we conducted McNemar's paired test for each system-benchmark combination. Table~\ref{tab:mcnemar_test_humaneval_vs_judge} presents the complete results, including discordant pair counts, accuracy differences with 95\% confidence intervals, odds ratios and p-values.

\begin{table*}[ht!]
    \centering
	\renewcommand{\arraystretch}{1.2}
    \caption{McNemar's paired test results comparing HumanEval against LLM-as-a-Judge evaluations for HippoRAG 2 and PAI-2 (with and without question preprocessing) across five benchmarks. The table displays discordant pair counts (b: LLM-as-a-Judge correct \& HumanEval incorrect; c: LLM-as-a-Judge incorrect \& HumanEval correct), accuracy differences with 95\% confidence intervals, odds ratios of discordant pairs and p-values (exact binomial and chi-square with continuity correction). Green cells indicate non-significant differences between evaluation methods (p-value > 0.05), suggesting good alignment between automated and human judgments, while pink cells indicate statistically significant differences (p-value < 0.05). Each benchmark evaluated on 100 question-answer pairs.}
	\resizebox{\textwidth}{!}{%
    \begin{tabular}{|ccccccc|}
    \hline
    \multicolumn{2}{|c|}{\multirow{3}{*}{\textbf{Metric}}} & \multicolumn{5}{c|}{\textbf{Benchmark}} \\ 
    \cline{3-7} 
    \multicolumn{2}{|c|}{} & \multicolumn{1}{c|}{\begin{tabular}[c]{@{}c@{}}\textbf{Natural} \\ \textbf{Questions}\end{tabular}} & \multicolumn{1}{c|}{\textbf{TriviaQA}} & \multicolumn{1}{c|}{\textbf{HotpotQA}} & \multicolumn{1}{c|}{\begin{tabular}[c]{@{}c@{}}\textbf{2Wiki}\\ \textbf{MultihopQA}\end{tabular}} & \textbf{MuSiQue} \\ 
    \hhline{=======}
    \multicolumn{2}{|c|}{N (number of samples)} & \multicolumn{1}{c|}{100} & \multicolumn{1}{c|}{100} & \multicolumn{1}{c|}{100} & \multicolumn{1}{c|}{100} & 100 \\ 
    \hhline{=======}
    \multicolumn{7}{|c|}{\textbf{HippoRAG 2 (qwen2.5:7b + contriever): HumanEval vs. LLM-as-a-Judge}} \\ 
    \hline
    \multicolumn{2}{|c|}{\begin{tabular}[c]{@{}c@{}}b (LLM-as-a-Judge = 1 \& \\ HumanEval = 0)\end{tabular}} & \multicolumn{1}{c|}{2} & \multicolumn{1}{c|}{0} & \multicolumn{1}{c|}{2} & \multicolumn{1}{c|}{6} & 2 \\ 
    \hline
    \multicolumn{2}{|c|}{\begin{tabular}[c]{@{}c@{}}c (LLM-as-a-Judge = 0 \& \\ HumanEval = 1)\end{tabular}} & \multicolumn{1}{c|}{6} & \multicolumn{1}{c|}{2} & \multicolumn{1}{c|}{4} & \multicolumn{1}{c|}{3} & 6 \\ 
    \hline
    \multicolumn{2}{|c|}{\begin{tabular}[c]{@{}c@{}}accuracy difference / \\ 95\% confidence interval\end{tabular}} & \multicolumn{1}{c|}{0.04 / {[}-0.014; 0.094{]}} & \multicolumn{1}{c|}{0.02 / {[}-0.007; 0.047{]}} & \multicolumn{1}{c|}{0.02 / {[}-0.027; 0.067{]}} & \multicolumn{1}{c|}{0.03 / {[}-0.028; 0.088{]}} & 0.04 / {[}-0.014; 0.094{]} \\ 
    \hline
    \multicolumn{2}{|c|}{\begin{tabular}[c]{@{}c@{}}odds ratio of discordant pairs /\\ 95\% confidence interval\end{tabular}} & \multicolumn{1}{c|}{0.333 / {[}0.067; 1.651{]}} & \multicolumn{1}{c|}{0.0 / -} & \multicolumn{1}{c|}{0.5 / {[}0.091; 2.729{]}} & \multicolumn{1}{c|}{2.0 / {[}0.500; 7.997{]}} & 0.333 / {[}0.067; 1.651{]} \\ 
    \hline
    \multicolumn{1}{|c|}{\multirow{3}{*}{\begin{tabular}[c]{@{}c@{}}McNemar's \\ paired test\end{tabular}}} & \multicolumn{1}{c|}{\begin{tabular}[c]{@{}c@{}}exact p-value \\ (binomial distribution)\end{tabular}} & \multicolumn{1}{c|}{ \cellcolor{grannysmithapple} 0.289} & \multicolumn{1}{c|}{\cellcolor{grannysmithapple} 0.5} & \multicolumn{1}{c|}{ \cellcolor{grannysmithapple} 0.687} & \multicolumn{1}{c|}{\cellcolor{grannysmithapple} 0.507} & \cellcolor{grannysmithapple} 0.289 \\ 
    \cline{2-7} 
    \multicolumn{1}{|c|}{} & \multicolumn{1}{c|}{\begin{tabular}[c]{@{}c@{}}p-value with \\ continuity correction \\ (chisquare distribution)\end{tabular}} & \multicolumn{1}{c|}{0.288} & \multicolumn{1}{c|}{0.479} & \multicolumn{1}{c|}{0.683} & \multicolumn{1}{c|}{0.504} & 0.288 \\ 
    \hhline{=======}
    \multicolumn{7}{|c|}{\textbf{PAI-2 (w/o question preprocessing): HumanEval vs. LLM-as-a-Judge}} \\ 
    \hline
    \multicolumn{2}{|c|}{\begin{tabular}[c]{@{}c@{}}b (LLM-as-a-Judge = 1 \& \\ HumanEval = 0)\end{tabular}} & \multicolumn{1}{c|}{0} & \multicolumn{1}{c|}{2} & \multicolumn{1}{c|}{0} & \multicolumn{1}{c|}{0} & 0 \\ 
    \hline
    \multicolumn{2}{|c|}{\begin{tabular}[c]{@{}c@{}}c (LLM-as-a-Judge = 0 \& \\ HumanEval = 1)\end{tabular}} & \multicolumn{1}{c|}{4} & \multicolumn{1}{c|}{3} & \multicolumn{1}{c|}{6} & \multicolumn{1}{c|}{3} & 3 \\ 
    \hline
    \multicolumn{2}{|c|}{\begin{tabular}[c]{@{}c@{}}accuracy difference / \\ 95\% confidence interval\end{tabular}} & \multicolumn{1}{c|}{0.04 / {[}0.001; 0.078{]}} & \multicolumn{1}{c|}{0.01 / {[}-0.033; 0.053{]}} & \multicolumn{1}{c|}{0.06 / {[}0.013; 0.106{]}} & \multicolumn{1}{c|}{0.03 / {[}-0.003; 0.063{]}} & 0.03 / {[}-0.003; 0.063{]} \\ 
    \hline
    \multicolumn{2}{|c|}{\begin{tabular}[c]{@{}c@{}}odds ratio of discordant pairs /\\ 95\% confidence interval\end{tabular}} & \multicolumn{1}{c|}{0.0 / -} & \multicolumn{1}{c|}{0.666 / {[}0.111; 3.989{]}} & \multicolumn{1}{c|}{0.0 / -} & \multicolumn{1}{c|}{0.0 / -} & 0.0 / - \\ 
    \hline
    \multicolumn{1}{|c|}{\multirow{3}{*}{\begin{tabular}[c]{@{}c@{}}McNemar's\\ paired test\end{tabular}}} & \multicolumn{1}{c|}{\begin{tabular}[c]{@{}c@{}}exact p-value \\ (binomial distribution)\end{tabular}} & \multicolumn{1}{c|}{\cellcolor{grannysmithapple} 0.125} & \multicolumn{1}{c|}{\cellcolor{grannysmithapple} 1.0} & \multicolumn{1}{c|}{\cellcolor{lightsalmonpink} 0.031} & \multicolumn{1}{c|}{\cellcolor{grannysmithapple} 0.25} & \cellcolor{grannysmithapple} 0.25 \\ 
    \cline{2-7} 
    \multicolumn{1}{|c|}{} & \multicolumn{1}{c|}{\begin{tabular}[c]{@{}c@{}}p-value with \\ continuity correction \\ (chisquare distribution)\end{tabular}} & \multicolumn{1}{c|}{0.133} & \multicolumn{1}{c|}{1.0} & \multicolumn{1}{c|}{0.041} & \multicolumn{1}{c|}{0.248} & 0.248 \\ 
    \hhline{=======}
    \multicolumn{7}{|c|}{\textbf{PAI-2 (w question preprocessing): HumanEval vs. LLM-as-a-Judge}} \\ \hline
    \multicolumn{2}{|c|}{\begin{tabular}[c]{@{}c@{}}b (LLM-as-a-Judge = 1 \& \\ HumanEval = 0)\end{tabular}} & \multicolumn{1}{c|}{0} & \multicolumn{1}{c|}{1} & \multicolumn{1}{c|}{0} & \multicolumn{1}{c|}{0} & 0 \\ 
    \hline
    \multicolumn{2}{|c|}{\begin{tabular}[c]{@{}c@{}}c (LLM-as-a-Judge = 0 \& \\ HumanEval = 1)\end{tabular}} & \multicolumn{1}{c|}{2} & \multicolumn{1}{c|}{3} & \multicolumn{1}{c|}{3} & \multicolumn{1}{c|}{2} & 4 \\ 
    \hline
    \multicolumn{2}{|c|}{\begin{tabular}[c]{@{}c@{}}accuracy difference / \\ 95\% confidence interval\end{tabular}} & \multicolumn{1}{c|}{0.02 / {[}-0.007; {]}0.047} & \multicolumn{1}{c|}{0.02 / {[}-0.019; 0.059{]}} & \multicolumn{1}{c|}{0.03 / {[}-0.003; 0.063{]}} & \multicolumn{1}{c|}{0.02 / {[}-0.007; 0.047{]}} & 0.04 / {[}0.001; 0.078{]} \\ 
    \hline
    \multicolumn{2}{|c|}{\begin{tabular}[c]{@{}c@{}}odds ratio of discordant pairs /\\ 95\% confidence interval\end{tabular}} & \multicolumn{1}{c|}{0.0 / -} & \multicolumn{1}{c|}{0.333 / {[}0.034; 3.204{]}} & \multicolumn{1}{c|}{0.0 / -} & \multicolumn{1}{c|}{0.0 / -} & 0.0 / - \\ 
    \hline
    \multicolumn{1}{|c|}{\multirow{3}{*}{\begin{tabular}[c]{@{}c@{}}McNemar's\\ paired test\end{tabular}}} & \multicolumn{1}{c|}{\begin{tabular}[c]{@{}c@{}}exact p-value \\ (binomial distribution)\end{tabular}} & \multicolumn{1}{c|}{\cellcolor{grannysmithapple} 0.5} & \multicolumn{1}{c|}{\cellcolor{grannysmithapple} 0.625} & \multicolumn{1}{c|}{\cellcolor{grannysmithapple} 0.25} & \multicolumn{1}{c|}{\cellcolor{grannysmithapple} 0.5} & \cellcolor{grannysmithapple} 0.125 \\ 
    \cline{2-7} 
    \multicolumn{1}{|c|}{} & \multicolumn{1}{c|}{\begin{tabular}[c]{@{}c@{}}p-value with \\ continuity correction \\ (chisquare distribution)\end{tabular}} & \multicolumn{1}{c|}{0.479} & \multicolumn{1}{c|}{0.617} & \multicolumn{1}{c|}{0.248} & \multicolumn{1}{c|}{0.479} & 0.133 \\ 
    \hline
    \end{tabular}}
    \renewcommand{\arraystretch}{1}
	\label{tab:mcnemar_test_humaneval_vs_judge}
\end{table*}

Out of 15 system-benchmark comparisons, only one shows a statistically significant difference between evaluation methods: PAI-2 without question preprocessing on HotpotQA (exact $p = 0.031$). All other comparisons yield $p > 0.05$, indicating no significant systematic bias between LLM-as-a-Judge and HumanEval. For HippoRAG 2, all five benchmarks show non-significant differences ($p = 0.289$ to $p = 0.687$), with accuracy differences ranging from 0.02 to 0.04 and 95\% confidence intervals that include zero. This confirms that the LLM judge provides statistically equivalent evaluations to human annotators for this baseline. For PAI-2 configurations, 9 out of 10 comparisons show non-significant differences. The single exception (HotpotQA, PAI-2 without question preprocessing) shows the LLM judge scoring 6 pp lower than human evaluators (0.75 vs. 0.81), with the confidence interval [0.013; 0.106] not including zero. 

The human evaluation study yields three principal conclusions:
\begin{itemize}
    \item Krippendorff's alpha values (mean: 0.93) confirm that human evaluators applied consistent criteria, establishing trustworthy ground truth labels for validation purposes.
    \item Raw agreement (mean: 0.95), Pearson correlation (mean: 0.88) and Cohen's kappa (mean: 0.88) indicate substantial to near-perfect agreement between automated and human judgments across all systems and benchmarks.
    \item McNemar's test shows no significant differences between evaluation methods in 14 out of 15 comparisons ($p > 0.05$), validating LLM-as-a-Judge as a reliable proxy for human evaluation in our experiments.
\end{itemize}

These findings justify our use of LLM-as-a-Judge as the primary evaluation metric throughout the paper. The automated judge provides scalable, consistent and statistically valid assessments that closely mirror human judgments, enabling comprehensive evaluation across all experimental configurations. The slight conservative bias observed (LLM scores 2-3 pp lower than HumanEval on average) suggests our reported performance metrics may underestimate the true system capabilities, strengthening rather than weakening our comparative conclusions.

\section{Statistical significance evaluation of RQ1 results}
\label{app:main_statsign_analysis}

To validate the performance differences observed in RQ1 experiments, we conducted statistical significance testing using McNemar's paired test across all method comparisons and benchmarks. This analysis determines whether observed accuracy differences between PAI-2 and baseline methods reflect true performance improvements rather than random variation.

McNemar's paired test is appropriate for comparing two classifiers on the same samples, accounting for the paired nature of the data. For each benchmark, all methods were evaluated on identical question-answer pairs, enabling direct statistical comparison. We report exact p-values based on the binomial distribution and apply Holm-Bonferroni correction to control the family-wise error rate across multiple comparisons. Table~\ref{tab:exactpvalues_mcnemar_test} presents exact p-values for all PAI-2 versus baseline comparisons. Table~\ref{tab:detailed_mcnemar_test_hr2_vs_pai2} provides detailed statistics for the PAI-2 versus HippoRAG 2 comparison.

\begin{table*}[ht!]
    \centering
	\renewcommand{\arraystretch}{1.3}
    \caption{Exact p-values of McNemar's paired test (binomial distribution) across five benchmarks for RQ1 experiments: comparison of PAI-2's settings with baseline methods. Green cells indicate not statistically significant difference in resulted LLM-as-a-Judge scores for corresponding pair by McNemar's test with p-value < 0.05. Red cells indicate statistically significant difference in resulted LLM-as-a-Judge scores for corresponding pair by McNemar's test with p-value > 0.05.}
    \begin{tabular}{|cccccccc|}
    \hline
    \multicolumn{1}{|c|}{\multirow{3}{*}{\textbf{Method}}} & \multicolumn{1}{c|}{\multirow{3}{*}{\textbf{LLM}}} & \multicolumn{1}{c|}{\multirow{3}{*}{\textbf{Embedder}}} & \multicolumn{5}{c|}{\textbf{Benchmark}} \\ 
    \cline{4-8} 
    \multicolumn{1}{|c|}{} & \multicolumn{1}{c|}{} & \multicolumn{1}{c|}{} & \multicolumn{1}{c|}{\begin{tabular}[c]{@{}c@{}}\textbf{Natural} \\ \textbf{Questions}\end{tabular}} & \multicolumn{1}{c|}{\textbf{TriviaQA}} & \multicolumn{1}{c|}{\textbf{HotpotQA}} & \multicolumn{1}{c|}{\begin{tabular}[c]{@{}c@{}}\textbf{2Wiki}\\ \textbf{MultihopQA}\end{tabular}} & \textbf{MuSiQue} \\ 
    \hhline{========}
    \multicolumn{8}{|c|}{\textbf{PAI-2 w/o query preprocessing}} \\ 
    \hline
    \multicolumn{1}{|c|}{\multirow{2}{*}{LightRAG}} & \multicolumn{1}{c|}{\multirow{8}{*}{Qwen2.5 7B}} & \multicolumn{1}{c|}{bge-m3} & \multicolumn{1}{c|}{\cellcolor{grannysmithapple} 3.365e-4} & \multicolumn{1}{c|}{\cellcolor{grannysmithapple} 4.266e-05} & \multicolumn{1}{c|}{\cellcolor{grannysmithapple} 2.327e-04} & \multicolumn{1}{c|}{\cellcolor{grannysmithapple} 1.149e-35} & \cellcolor{grannysmithapple} 3.910e-07 \\ 
    \cline{3-8}
    \multicolumn{1}{|c|}{} & \multicolumn{1}{c|}{} & \multicolumn{1}{c|}{m-e5-large} & \multicolumn{1}{c|}{\cellcolor{grannysmithapple} 6.719e-12} & \multicolumn{1}{c|}{\cellcolor{lightsalmonpink} 3.291e-01} & \multicolumn{1}{c|}{\cellcolor{lightsalmonpink} 6.768e-01} & \multicolumn{1}{c|}{\cellcolor{grannysmithapple} 1.644e-43} & \cellcolor{grannysmithapple} 3.082e-06 \\ 
    \cline{1-1} \cline{3-8} 
    \multicolumn{1}{|c|}{\multirow{2}{*}{RAPTOR}} & \multicolumn{1}{c|}{} & \multicolumn{1}{c|}{\begin{tabular}[c]{@{}c@{}}multi-qa-mpnet-\\ base-cos-v1\end{tabular}} & \multicolumn{1}{c|}{\cellcolor{lightsalmonpink} 3.548e-01} & \multicolumn{1}{c|}{\cellcolor{grannysmithapple} 1.987e-02} & \multicolumn{1}{c|}{\cellcolor{grannysmithapple} 6.161e-29} & \multicolumn{1}{c|}{\cellcolor{grannysmithapple} 4.204e-88} & \cellcolor{grannysmithapple} 1.102e-29 \\ 
    \cline{3-8} 
    \multicolumn{1}{|c|}{} & \multicolumn{1}{c|}{} & \multicolumn{1}{c|}{m-e5-large} & \multicolumn{1}{c|}{\cellcolor{lightsalmonpink} 2.642e-01} & \multicolumn{1}{c|}{\cellcolor{lightsalmonpink} 8.422e-02} & \multicolumn{1}{c|}{\cellcolor{grannysmithapple} 4.813e-16} & \multicolumn{1}{c|}{\cellcolor{grannysmithapple} 3.302e-73} & \cellcolor{grannysmithapple} 8.333e-24 \\ 
    \cline{1-1} \cline{3-8}
    \multicolumn{1}{|c|}{\multirow{2}{*}{HippoRAG 2}} & \multicolumn{1}{c|}{} & \multicolumn{1}{c|}{contriever} & \multicolumn{1}{c|}{\cellcolor{grannysmithapple} 4.768e-06} & \multicolumn{1}{c|}{\cellcolor{lightsalmonpink} 8.205e-02} & \multicolumn{1}{c|}{\cellcolor{lightsalmonpink} 1.0} & \multicolumn{1}{c|}{\cellcolor{grannysmithapple} 1.470e-02} & \cellcolor{grannysmithapple} 1.182e-05 \\ 
    \cline{3-8} 
    \multicolumn{1}{|c|}{} & \multicolumn{1}{c|}{} & \multicolumn{1}{c|}{m-e5-large} & \multicolumn{1}{c|}{\cellcolor{grannysmithapple} 1.482e-06} & \multicolumn{1}{c|}{\cellcolor{lightsalmonpink} 2.836e-01} & \multicolumn{1}{c|}{\cellcolor{lightsalmonpink} 1.813e-01} & \multicolumn{1}{c|}{\cellcolor{grannysmithapple} 8.226e-09} & \cellcolor{grannysmithapple} 1.643e-08 \\ 
    \cline{1-1} \cline{3-8} 
    \multicolumn{1}{|c|}{PAI-1} & \multicolumn{1}{c|}{} & \multicolumn{1}{c|}{m-e5-large} & \multicolumn{1}{c|}{\cellcolor{grannysmithapple} 6.744e-20} & \multicolumn{1}{c|}{\cellcolor{grannysmithapple} 1.307e-03} & \multicolumn{1}{c|}{\cellcolor{grannysmithapple} 1.121e-19} & \multicolumn{1}{c|}{\cellcolor{grannysmithapple} 2.116e-48} & \cellcolor{grannysmithapple} 1.546e-28 \\ 
    \hhline{========}
    \multicolumn{8}{|c|}{\textbf{PAI-2 w query preprocessing}} \\ 
    \hline
    \multicolumn{1}{|c|}{\multirow{2}{*}{LightRAG}} & \multicolumn{1}{c|}{\multirow{8}{*}{Qwen2.5 7B}} & \multicolumn{1}{c|}{bge-m3} & \multicolumn{1}{c|}{\cellcolor{grannysmithapple} 2.069e-05} & \multicolumn{1}{c|}{\cellcolor{grannysmithapple} 1.631e-03} & \multicolumn{1}{c|}{\cellcolor{lightsalmonpink} 4.549e-01} & \multicolumn{1}{c|}{\cellcolor{grannysmithapple} 3.886e-41} & \cellcolor{grannysmithapple} 7.087e-04 \\ 
    \cline{3-8} 
    \multicolumn{1}{|c|}{} & \multicolumn{1}{c|}{} & \multicolumn{1}{c|}{m-e5-large} & \multicolumn{1}{c|}{\cellcolor{grannysmithapple} 6.651e-14} & \multicolumn{1}{c|}{\cellcolor{lightsalmonpink} 1.0} & \multicolumn{1}{c|}{\cellcolor{grannysmithapple} 5.174e-05} & \multicolumn{1}{c|}{\cellcolor{grannysmithapple} 1.175e-48} & \cellcolor{grannysmithapple} 3.465e-03 \\ 
    \cline{1-1} \cline{3-8} 
    \multicolumn{1}{|c|}{\multirow{2}{*}{RAPTOR}} & \multicolumn{1}{c|}{} & \multicolumn{1}{c|}{\begin{tabular}[c]{@{}c@{}}multi-qa-mpnet-\\ base-cos-v1\end{tabular}} & \multicolumn{1}{c|}{\cellcolor{lightsalmonpink} 8.206e-01} & \multicolumn{1}{c|}{\cellcolor{lightsalmonpink} 1.991e-01} & \multicolumn{1}{c|}{\cellcolor{grannysmithapple} 5.777e-11} & \multicolumn{1}{c|}{\cellcolor{grannysmithapple} 2.519e-96} & \cellcolor{grannysmithapple} 1.397e-21 \\ 
    \cline{3-8} 
    \multicolumn{1}{|c|}{} & \multicolumn{1}{c|}{} & \multicolumn{1}{c|}{m-e5-large} & \multicolumn{1}{c|}{\cellcolor{lightsalmonpink} 6.532e-01} & \multicolumn{1}{c|}{\cellcolor{lightsalmonpink} 5.122e-01} & \multicolumn{1}{c|}{\cellcolor{grannysmithapple} 4.460e-04} & \multicolumn{1}{c|}{\cellcolor{grannysmithapple} 1.913e-75} & \cellcolor{grannysmithapple} 1.310e-16 \\ 
    \cline{1-1} \cline{3-8}
    \multicolumn{1}{|c|}{\multirow{2}{*}{HippoRAG 2}} & \multicolumn{1}{c|}{} & \multicolumn{1}{c|}{contriever} & \multicolumn{1}{c|}{\cellcolor{grannysmithapple} 2.856e-07} & \multicolumn{1}{c|}{\cellcolor{grannysmithapple} 6.092e-03} & \multicolumn{1}{c|}{\cellcolor{grannysmithapple} 8.220e-06} & \multicolumn{1}{c|}{\cellcolor{grannysmithapple} 7.316e-06} & \cellcolor{grannysmithapple} 5.923e-03 \\ 
    \cline{3-8} 
    \multicolumn{1}{|c|}{} & \multicolumn{1}{c|}{} & \multicolumn{1}{c|}{m-e5-large} & \multicolumn{1}{c|}{\cellcolor{grannysmithapple} 9.717e-08} & \multicolumn{1}{c|}{\cellcolor{grannysmithapple} 3.895e-02} & \multicolumn{1}{c|}{\cellcolor{grannysmithapple} 3.538e-09} & \multicolumn{1}{c|}{\cellcolor{grannysmithapple} 5.754e-15} & \cellcolor{grannysmithapple} 5.195e-05 \\ 
    \cline{1-1} \cline{3-8}  
    \multicolumn{1}{|c|}{PAI-1} & \multicolumn{1}{c|}{} & \multicolumn{1}{c|}{m-e5-large} & \multicolumn{1}{c|}{\cellcolor{grannysmithapple} 7.399e-18} & \multicolumn{1}{c|}{\cellcolor{grannysmithapple} 9.210e-03} & \multicolumn{1}{c|}{\cellcolor{grannysmithapple} 4.096e-06} & \multicolumn{1}{c|}{\cellcolor{grannysmithapple} 9.805e-59} & \cellcolor{grannysmithapple} 4.214e-20 \\ 
    \hline
    \end{tabular}
    \renewcommand{\arraystretch}{1}
	\label{tab:exactpvalues_mcnemar_test}
\end{table*}

Analysis of Table~\ref{tab:exactpvalues_mcnemar_test} reveals distinct patterns across baseline comparisons. PAI-2 without question preprocessing achieves statistically significant improvements over LightRAG on 2WikiMultihopQA ($p < 10^{-35}$) and MuSiQue ($p < 10^{-6}$). Against RAPTOR, significance is achieved on HotpotQA, 2WikiMultihopQA, and MuSiQue ($p < 10^{-16}$), but not on Natural Questions and TriviaQA ($p > 0.05$). Compared to HippoRAG 2, significant improvements are observed on 2WikiMultihopQA ($p < 0.015$), and MuSiQue ($p < 10^{-5}$), while TriviaQA and HotpotQA show non-significant differences ($p > 0.05$). Against PAI-1, PAI-2 demonstrates highly significant improvements across all benchmarks ($p < 10^{-3}$ to $p < 10^{-48}$). PAI-2 with question preprocessing shows a different significance pattern. Against HippoRAG 2, all five benchmarks show statistically significant differences ($p < 0.006$), including HotpotQA and TriviaQA which were non-significant with disabled Question Preprocessing stage. Against LightRAG and RAPTOR, the significance pattern similar to w/o-preprocessing variant, with strong effects on multi-hop benchmarks. Against PAI-1, all comparisons remain significant ($p <10^{-3}$ to $p < 10^{-59}$).

Table~\ref{tab:detailed_mcnemar_test_hr2_vs_pai2} provides comprehensive statistics for the strongest baseline comparison.

\begin{table*}[ht!]
    \centering
	\renewcommand{\arraystretch}{1.2}
    \caption{McNemar's paired test results comparing PAI-2 (with and without question preprocessing) against HippoRAG 2 (Qwen2.5 7B + Contriever) from RQ1 across five benchmarks. The table displays discordant pair counts (b: PAI-2 correct \& HippoRAG 2 incorrect; c: PAI-2 incorrect \& HippoRAG 2 correct), accuracy differences with 95\% confidence intervals, odds ratios of discordant pairs and p-values (exact binomial, chi-square with continuity correction and Holm-adjusted for multiple comparisons). Green cells indicate statistically significant differences (adjusted p-value < 0.05), while pink cells indicate non-significant differences.}
	\resizebox{\textwidth}{!}{%
    \begin{tabular}{|ccccccc|}
    \hline
    \multicolumn{2}{|c|}{\multirow{3}{*}{\textbf{Metric}}} & \multicolumn{5}{c|}{\textbf{Benchmark}} \\ 
    \cline{3-7} 
    \multicolumn{2}{|c|}{} & \multicolumn{1}{c|}{\begin{tabular}[c]{@{}c@{}}\textbf{Natural} \\ \textbf{Questions}\end{tabular}} & \multicolumn{1}{c|}{\textbf{TriviaQA}} & \multicolumn{1}{c|}{\textbf{HotpotQA}} & \multicolumn{1}{c|}{\begin{tabular}[c]{@{}c@{}}\textbf{2Wiki}\\ \textbf{MultihopQA}\end{tabular}} & \textbf{MuSiQue} \\ 
    \hhline{=======}
    \multicolumn{2}{|c|}{N (number of samples)} & \multicolumn{1}{c|}{1000} & \multicolumn{1}{c|}{500} & \multicolumn{1}{c|}{1000} & \multicolumn{1}{c|}{1000} & 1000 \\ 
    \hhline{=======}
    \multicolumn{7}{|c|}{\textbf{PAI-2 (w/o question preprocessing) vs. HippoRAG 2 (qwen2.5:7b + contriever)}} \\ 
    \hline
    \multicolumn{2}{|c|}{b (PAI2 = 1 \& HippoRAG2 = 0)} & \multicolumn{1}{c|}{95} & \multicolumn{1}{c|}{34} & \multicolumn{1}{c|}{147} & \multicolumn{1}{c|}{210} & 223 \\ 
    \hline
    \multicolumn{2}{|c|}{c (PAI2 = 0 \& HippoRAG2 = 1)} & \multicolumn{1}{c|}{170} & \multicolumn{1}{c|}{51} & \multicolumn{1}{c|}{147} & \multicolumn{1}{c|}{162} & 139 \\ 
    \hline
    \multicolumn{2}{|c|}{\begin{tabular}[c]{@{}c@{}}accuracy difference / \\ 95\% confidence interval\end{tabular}} & \multicolumn{1}{c|}{0.075 / {[}0.043; 0.106{]}} & \multicolumn{1}{c|}{0.034 / {[}-0.002; 0.070{]}} & \multicolumn{1}{c|}{0.0 / {[}-0.033; 0.033{]}} & \multicolumn{1}{c|}{0.048 / {[}0.010; 0.085{]}} & 0.084 / {[}0.047; 0.120{]} \\ 
    \hline
    \multicolumn{2}{|c|}{\begin{tabular}[c]{@{}c@{}}odds ratio of discordant pairs /\\ 95\% confidence interval\end{tabular}} & \multicolumn{1}{c|}{0.558 / {[}0.434; 0.718{]}} & \multicolumn{1}{c|}{0.666 / {[}0.431; 1.028{]}} & \multicolumn{1}{c|}{1.0 / {[}0.795; 1.256{]}} & \multicolumn{1}{c|}{1.296 / {[}1.056; 1.591{]}} & 1.604 / {[}1.298; 1.982{]} \\ 
    \hline
    \multicolumn{1}{|c|}{\multirow{4}{*}{\begin{tabular}[c]{@{}c@{}}McNemar's \\ paired test\end{tabular}}} & \multicolumn{1}{c|}{\begin{tabular}[c]{@{}c@{}}exact p-value \\ (binomial distribution)\end{tabular}} & \multicolumn{1}{c|}{4.768e-06} & \multicolumn{1}{c|}{8.205e-02} & \multicolumn{1}{c|}{1.0} & \multicolumn{1}{c|}{1.470e-02} & 1.182e-05 \\ 
    \cline{2-7} 
    \multicolumn{1}{|c|}{} & \multicolumn{1}{c|}{\begin{tabular}[c]{@{}c@{}}p-value with \\ continuity correction \\ (chisquare distribution)\end{tabular}} & \multicolumn{1}{c|}{5.473e-06} & \multicolumn{1}{c|}{8.266e-02} & \multicolumn{1}{c|}{9.534e-01} & \multicolumn{1}{c|}{1.481e-02} & 1.286e-05 \\ 
    \cline{2-7} 
    \multicolumn{1}{|c|}{} & \multicolumn{1}{c|}{Holm-adjusted p-value} & \multicolumn{1}{c|}{\cellcolor{grannysmithapple} 8.105e-05} & \multicolumn{1}{c|}{\cellcolor{lightsalmonpink} 7.384e-01} & \multicolumn{1}{c|}{\cellcolor{lightsalmonpink} 1.000} & \multicolumn{1}{c|}{\cellcolor{lightsalmonpink} 1.617e-01} & \cellcolor{grannysmithapple} 1.891e-04 \\ 
    \hhline{=======}
    \multicolumn{7}{|c|}{\textbf{PAI-2 (w question preprocessing) vs. HippoRAG 2 (qwen2.5:7b + contriever)}} \\ 
    \hline
    \multicolumn{2}{|c|}{b (PAI2 = 1 \& HippoRAG2 = 0)} & \multicolumn{1}{c|}{99} & \multicolumn{1}{c|}{35} & \multicolumn{1}{c|}{122} & \multicolumn{1}{c|}{229} & 213 \\ 
    \hline
    \multicolumn{2}{|c|}{c (PAI2 = 0 \& HippoRAG2 = 1)} & \multicolumn{1}{c|}{186} & \multicolumn{1}{c|}{63} & \multicolumn{1}{c|}{203} & \multicolumn{1}{c|}{142} & 159 \\ 
    \hline
    \multicolumn{2}{|c|}{\begin{tabular}[c]{@{}c@{}}accuracy difference / \\ 95\% confidence interval\end{tabular}} & \multicolumn{1}{c|}{0.087 / {[}0.054; 0.119{]}} & \multicolumn{1}{c|}{0.056 / {[}0.017; 0.094{]}} & \multicolumn{1}{c|}{0.081 / {[}0.046; 0.115{]}} & \multicolumn{1}{c|}{0.087 / {[}0.049; 0.124{]}} & 0.054 / {[}0.016; 0.091{]} \\ 
    \hline
    \multicolumn{2}{|c|}{\begin{tabular}[c]{@{}c@{}}odds ratio of discordant pairs /\\ 95\% confidence interval\end{tabular}} & \multicolumn{1}{c|}{0.532 / {[}0.417; 0.679{]}} & \multicolumn{1}{c|}{0.555 / {[}0.367; 0.839{]}} & \multicolumn{1}{c|}{0.600 / {[}0.480; 0.752{]}} & \multicolumn{1}{c|}{1.612 / {[}1.308; 1.988{]}} & 1.339 / {[}1.090; 1.645{]} \\ 
    \hline
    \multicolumn{1}{|c|}{\multirow{4}{*}{\begin{tabular}[c]{@{}c@{}}McNemar's \\ paired test\end{tabular}}} & \multicolumn{1}{c|}{\begin{tabular}[c]{@{}c@{}}exact p-value \\ (binomial distribution)\end{tabular}} & \multicolumn{1}{c|}{2.856e-07} & \multicolumn{1}{c|}{6.092e-03} & \multicolumn{1}{c|}{8.220e-06} & \multicolumn{1}{c|}{7.316e-06} & 5.923e-03 \\ 
    \cline{2-7} 
    \multicolumn{1}{|c|}{} & \multicolumn{1}{c|}{\begin{tabular}[c]{@{}c@{}}p-value with \\ continuity correction \\ (chisquare distribution)\end{tabular}} & \multicolumn{1}{c|}{3.502e-07} & \multicolumn{1}{c|}{6.383e-03} & \multicolumn{1}{c|}{9.096e-06} & \multicolumn{1}{c|}{8.010e-06} & 5.997e-03 \\ 
    \cline{2-7} 
    \multicolumn{1}{|c|}{} & \multicolumn{1}{c|}{Holm-adjusted p-value} & \multicolumn{1}{c|}{\cellcolor{grannysmithapple} 5.997e-06} & \multicolumn{1}{c|}{\cellcolor{lightsalmonpink} 5.923e-02} & \multicolumn{1}{c|}{\cellcolor{grannysmithapple} 1.479e-04} & \multicolumn{1}{c|}{\cellcolor{grannysmithapple}1.390e-04} & \cellcolor{lightsalmonpink} 5.923e-02 \\ 
    \hline
    \end{tabular}}
    \renewcommand{\arraystretch}{1}
	\label{tab:detailed_mcnemar_test_hr2_vs_pai2}
\end{table*}


The statistical significance analysis validates the main RQ1 conclusions. First, PAI-2 achieves significantly superior performance on complex multi-hop reasoning benchmarks (2WikiMultihopQA and MuSiQue) compared to all baselines, including HippoRAG 2. Second, PAI-2 attains statistically comparable performance to HippoRAG 2 on factoid benchmarks (TriviaQA and HotpotQA), with non-significant differences after multiple comparison correction. Third, the 18 pp average improvement over PAI-1 is  statistically significant across all benchmarks ($p < 10^{-3}$ to $p < 10^{-59}$), validating the search planning mechanism.

These findings confirm that performance differences reported in RQ1 reflect true methodological advantages rather than random variation, strengthening the validity of our comparative conclusions about PAI-2's effectiveness.

\section{PAI-2 evaluation on MINE-1}
\label{app:mine_comparison}

\begin{figure}[H]
	\centering
	\includegraphics[width=.8\textwidth]{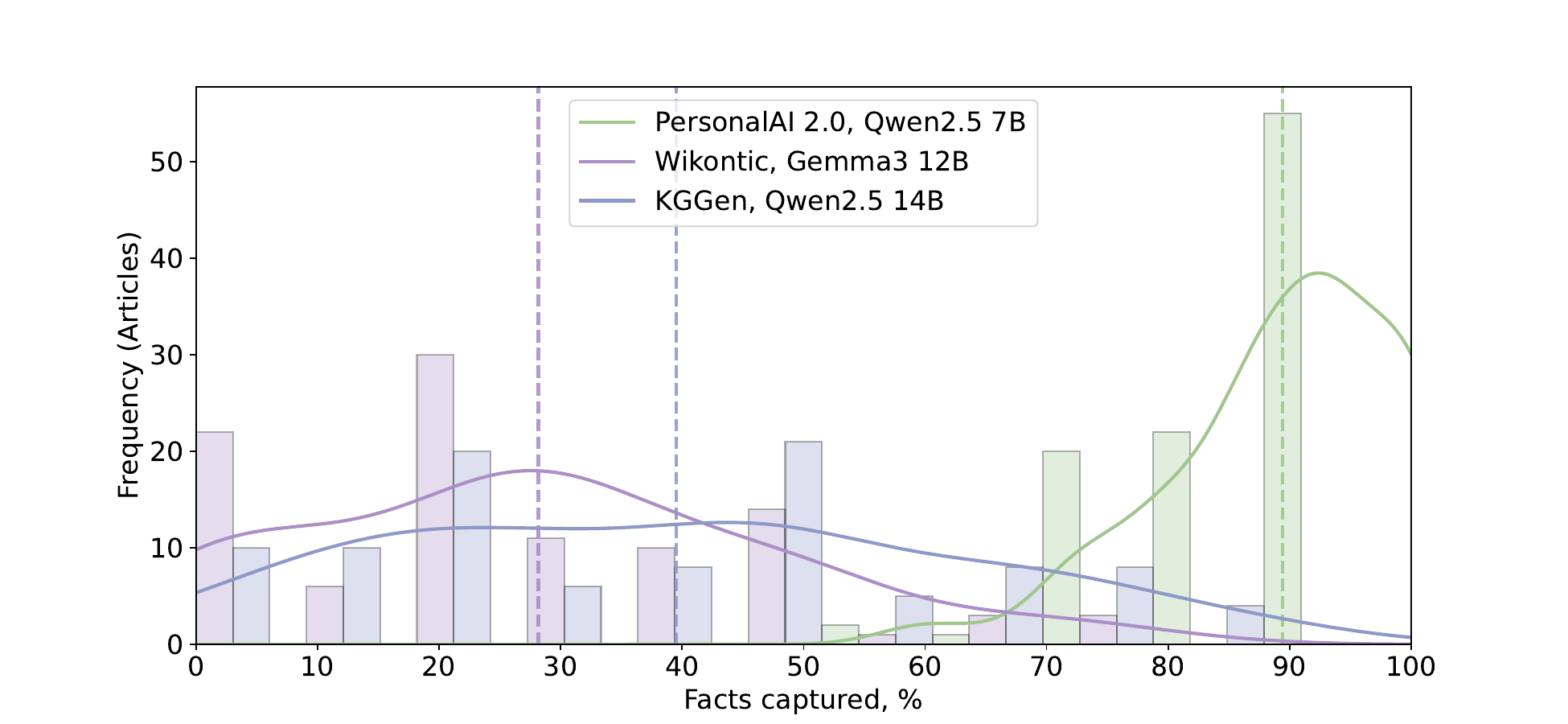}
	\caption{Distribution of MINE-1 scores across 100 articles for PAI-2, Wikontic and KGGen. Dotted vertical lines are averaged scores. PAI-2 scored 89\% on average, substantially outperforming Wikontic 28\% and KGGen 39\%.} 
	\label{fig:mine_diagram}
\end{figure}

We evaluated PAI-2 on MINE-1 benchmark, which measures how much factual information from the source text is retained in the constructed KG using an LLM-as-a-judge protocol from the original study~\cite{mo2025kggenextractingknowledgegraphs}. Figure~\ref{fig:mine_diagram} displays the retention scores distribution in articles of MINE-1 for PAI-2, Wikontick~\cite{chepurova-etal-2026-wikontic} and KGGen~\cite{mo2025kggenextractingknowledgegraphs}. Table ~\ref{tab:mine_comparison} demonstrates the results for KGGen~\cite{mo2025kggenextractingknowledgegraphs}, Wikontic~\cite{chepurova-etal-2026-wikontic}, GraphRAG~\cite{edge2025localglobalgraphrag} and PAI-2 with different LLM backbones. PAI-2 consistently outperforms other methods, reaching 89\% with Qwen2.5 7B and multilingual-e5-large compared to Wikontick's best score of 86\% (gpt4.1-mini and Contriever). These results demonstrate that PAI effectively preserves factual information during the construction of memory graphs.

\begin{table}[H]
    \centering
	\renewcommand{\arraystretch}{1.3}
    \caption{MINE-1 information-retention scores for KGGen, Wikontic, GraphRAG and PAI-2. PAI-2 achieves the highest retention performance. For PAI-2 evaluation, during triples retrieving (according to MINE setup) we only accept object and thesis vertices from constructed memory graph. Bold value indicate best performance.}
    \begin{tabular}{|c|c|c|c|}
    \hline
    \textbf{Method} & \textbf{LLM} & \textbf{Embedder} & \textbf{MINE-1 Score (\%)} \\ \hline \hline
    \multirow{5}{*}{KGGen} & Claude Sonnet 3.5 & \multirow{5}{*}{all-MiniLM-L6-v2} & 73 \\ \cline{2-2} \cline{4-4} 
     & GPT-4o & & 66 \\ \cline{2-2} \cline{4-4} 
     & Gemini 2.0 Flash & & 44 \\ \cline{2-2} \cline{4-4}
     & Qwen2.5 14B & & 39 \\ \cline{2-2} \cline{4-4}
     & Gemma3 12B & & 14 \\ \hline
    \multirow{4}{*}{Wikontic} & gpt4.1-mini & \multirow{4}{*}{contriever} & 86 \\ \cline{2-2} \cline{4-4}
     & gpt4o & & 84 \\ \cline{2-2} \cline{4-4}
     & Gemma3 12B & & 28 \\  \cline{2-2} \cline{4-4}
     & Qwen2.5 14B & & 19 \\ \hline
    GraphRAG & gpt4o & text-embedding-ada-002 & 44 \\ \hline
    PAI-2 & Qwen2.5 7B & multilingual-e5-large & \textbf{89} \\ \hline
    \end{tabular}
	\label{tab:mine_comparison}
\end{table}

Additionally, we perform ablation experiments for PAI-2 to understand how MINE-1 score relates to LLM backbone and accepted vertex types: see Table~\ref{tab:mine_acceptedntypes_ablation}.

\begin{table}[H]
    \centering
	\renewcommand{\arraystretch}{1.3}
    \caption{Dependence of MINE-1 information-retention score on accepted vertex types for PAI-2 across six LLMs. Bold values indicate best performance per LLM.}
    \begin{tabular}{|c|c|c|c|c|c|c||c}
    \hhline{|-|------||-|}
    \multirow{2}{*}{\begin{tabular}[c]{@{}c@{}}\textbf{Accepted}\\ \textbf{Vertex Types}\end{tabular}} & \multicolumn{6}{c||}{\textbf{LLM}} & \multicolumn{1}{c|}{\multirow{2}{*}{Mean}} \\ 
    \hhline{|~|------||~|}
    & \multicolumn{1}{c|}{\textbf{Qwen2.5 7B}} & \multicolumn{1}{c|}{\textbf{Llama3.1 8B}} & \multicolumn{1}{c|}{\textbf{Granite3.3 8B}} & \multicolumn{1}{c|}{\textbf{Gemma2 9B}} & \multicolumn{1}{c|}{\textbf{Gemma3 12B}} & \textbf{Qwen2.5 14B} & \multicolumn{1}{c|}{} \\ 
    \hhline{=======::=}
    object & 67 & 38 & 52 & 61 & 77 & 64 & \multicolumn{1}{c|}{60} \\ 
    \hhline{-------||-|}
    thesis & 81 & 76 & 66 & 81 & 76 & 80 & \multicolumn{1}{c|}{77} \\ 
    \hhline{-------||-|}
    episodic & 93 & \textbf{94} & \textbf{95} & 94 & 93 & 93 & \multicolumn{1}{c|}{94} \\ 
    \hhline{-------||-|}
    object, thesis & 89 & 80 & 78 & 89 & 85 & 88 & \multicolumn{1}{c|}{85} \\ 
    \hhline{-------||-|}
    object, thesis, episodic & \textbf{96} & \textbf{94} & 94 & \textbf{96} & \textbf{97} & \textbf{95} & \multicolumn{1}{c|}{\textbf{95}} \\
    \hhline{=======::=}
    Mean & 85 & 76 & 77 & 84 & \textbf{86} & 84 & \multicolumn{1}{c|}{82} \\
    \hhline{|-|------||-|}
    \end{tabular}
    \renewcommand{\arraystretch}{1}
	\label{tab:mine_acceptedntypes_ablation}
\end{table}

From Table \ref{tab:mine_acceptedntypes_ablation} it can be observed that the highest MINE-1 score is achieved when we accept triples, that is incident to all vertex types: object, thesis and episodic. However, when retrieving only episodic triples, quality degrades by only 1 pp. This means that object vertices (that are matched to the queries entities) are adjacent to episodic vertices, containing required knowledge to generate relevant responses. In other words, the set of object vertices, extracted from episodic memories, is sufficient to find a relevant, but redundant source document. Conversely, when generating responses based on triples, that incident only to object and thesis vertices, significant degradation is observed: on average 10 pp. This may indicate that the number of thesis and simple triples, extracted from episodic memories, is insufficient to cover the entire amount of knowledge they contain, which may be required to generate correct responses. Thus, to improve graph construction algorithm, it is necessary to include additional mechanics that (1) evaluate the knowledge coverage degree of original document (episodic vertex) with corresponding set of extracted/generated triples, (2) localize missing units of knowledge and (3) perform an additional extracting/generating round to get missing triples.

\section{Computational Cost Analysis}
\label{app:qapipe_execution_characteristics}

We analyze the computational requirements of PAI-2 across three dimensions: execution latency, LLM call frequency, and token consumption. Understanding these metrics is essential for practitioners considering deployment in production environments and for researchers seeking to reproduce our experimental results. Table~\ref{tab:latency_evaluation} presents end-to-end latency for processing a single query across all methods and benchmarks.

\begin{table*}[ht!]
    \centering
	\renewcommand{\arraystretch}{1.3}
    \caption{Latency (in seconds) of LightRAG, RAPTOR, HippoRAG, PAI-1 and PAI-2 QA pipelines across five benchmarks}
    \begin{tabular}{|c|c|c|c|c|c||c|}
    \hhline{------||-}
    \multirow{2}{*}{\textbf{Method}} & \multicolumn{5}{c||}{\textbf{Benchmark}} & \multirow{2}{*}{Mean} \\ 
    \hhline{~-----||~}
     & \textbf{Natural Questions} & \textbf{TriviaQA} & \textbf{HotpotQA} & \textbf{2WikiMultihopQA} & \textbf{MuSiQue} &  \\ 
     \hhline{======::=}
    LightRAG & 15 & 16 & 13 & 15 & 14 & 14 \\ 
    \hhline{------||-}
    RAPTOR & 3 & 3 & 2 & 3 & 3 & 3 \\ 
    \hhline{------||-}
    HippoRAG 2 & 5 & 6 & 6 & 6 & 6 & 6 \\ 
    \hhline{------||-}
    PAI-1 & 18 & 28 & 27 & 24 & 23 & 24 \\ 
    \hhline{------||-}
    PAI-2 & 85 & 124 & 115 & 200 & 158 & 136 \\ 
    \hhline{------||-}
    \end{tabular}
    \renewcommand{\arraystretch}{1}
	\label{tab:latency_evaluation}
\end{table*}



Table~\ref{tab:latency_evaluation} shows that RAPTOR achieves the lowest latency (3s mean), followed by HippoRAG 2 (6s) and LightRAG (14s). PAI-1 requires 24s on average, while PAI-2 requires 136s (5.7$\times$ slower than PAI-1). This substantial increase reflects PAI-2's iterative search planning mechanism: PAI-1 performs a single retrieval iteration, whereas PAI-2 executes multiple search steps with adaptive clue-queries generation. Latency varies across benchmarks for PAI-2, ranging from 85s (Natural Questions) to 200s (2WikiMultihopQA). This variation correlates with question complexity: multi-hop benchmarks require more search iterations and clue-query generations, increasing total execution time. The primary bottlenecks in PAI-2's pipeline are: (1) LLM inference for clue-query generation and response synthesis; (2) vector search for semantic similarity matching; (3) knowledge graph traversal for multi-hop retrieval. To mitigate these factors, practitioners can: (1) cache and reuse LLM inference results; (2) partition vector stores by element types to reduce search space; and (3) constrain the maximum search steps ($S_m$) and clue-queries per step ($C_m$).

We also analyze the token consumption and LM call count. Table~\ref{tab:pai2_llm_calls_by_stage} shows the number of LLM calls, required for each stage of the PAI-2's QA pipeline.

\begin{table}[H]
    \centering
    \renewcommand{\arraystretch}{1.3}
    \caption{LLM call breakdown for PAI-2 QA pipeline stages. The amount of LLM calls in Memory Graph Exploration stage depend on $S$ (number of the search steps in a plan) and $C$ (number of clue-queries, generated per search step) hyperparameters.}
    \begin{tabular}{|cc|c|}
    \hline
    \multicolumn{2}{|c|}{\textbf{PAI-2 Stages}} & \textbf{LLM Calls} \\ 
    \hhline{===}
    \multicolumn{1}{|c|}{\multirow{3}{*}{\begin{tabular}[c]{@{}c@{}}Question\\ Preprocessing\end{tabular}}} & Denoise & 2 \\ 
    \cline{2-3} 
    \multicolumn{1}{|c|}{} & Enhance & 3 \\ 
    \cline{2-3} 
    \multicolumn{1}{|c|}{} & Decompose & 2 \\ 
    \hline
    \multicolumn{2}{|c|}{\begin{tabular}[c]{@{}c@{}}Memory Graph\\ Exploration\end{tabular}} & $2 + S * (5 + C * 2)$ \\ 
    \hline
    \multicolumn{2}{|c|}{\begin{tabular}[c]{@{}c@{}}Answer\\ Aggregation\end{tabular}} & 1 \\ 
    \hline
    \end{tabular}
    \renewcommand{\arraystretch}{1}
    \label{tab:pai2_llm_calls_by_stage}
\end{table}

From Table~\ref{tab:pai2_llm_calls_by_stage} it can be observed that Question Preprocessing and Answer Aggregation together require 8 LLM calls per query. Memory Graph Exploration dominates the total call count, ranging from 9 calls ($S=1, C=1$) to 170 calls ($S=8, C=8$). Table~\ref{tab:pai2_llm_calls_hyperparams} quantifies the total LLM calls across different configurations of $S_m$ (max search steps) and $C_m$ (max clue queries) hyperparameters in a worst-case scenarios.

\begin{table}[H]
    \centering
    \renewcommand{\arraystretch}{1.3}
    \caption{Total LLM calls for PAI's Memory Graph Exploration Stage across different hyperparameter configurations in a worst-case scenarios, then the maximum amount of search steps in a plan ($S_m$) were reached and for every search step the maximum number of clue-queries ($C_m$) were generated.}
    \begin{tabular}{|cccccc|}
    \hline
    \multicolumn{6}{|c|}{\textbf{Hyperparameters of PAI-2's Memory Graph Exploration Stage}} \\
    \hline
    \multicolumn{1}{|c|}{\multirow{2}{*}{\begin{tabular}[c]{@{}c@{}}\textbf{Max Search} \\ \textbf{Steps ($S_m$)}\end{tabular}}} & \multicolumn{5}{c|}{\textbf{Max Clue-Queries ($C_m$)}} \\
    \cline{2-6}
    \multicolumn{1}{|c|}{} & \multicolumn{1}{c|}{\textbf{1}} & \multicolumn{1}{c|}{\textbf{2}} & \multicolumn{1}{c|}{\textbf{4}} & \multicolumn{1}{c|}{\textbf{6}} & \textbf{8} \\
    \hhline{======}
    \multicolumn{1}{|c|}{1} & \multicolumn{1}{c|}{9} & \multicolumn{1}{c|}{11} & \multicolumn{1}{c|}{15} & \multicolumn{1}{c|}{19} & 23 \\
    \hline
    \multicolumn{1}{|c|}{2} & \multicolumn{1}{c|}{16} & \multicolumn{1}{c|}{20} & \multicolumn{1}{c|}{28} & \multicolumn{1}{c|}{36} & 44 \\
    \hline
    \multicolumn{1}{|c|}{4} & \multicolumn{1}{c|}{30} & \multicolumn{1}{c|}{38} & \multicolumn{1}{c|}{54} & \multicolumn{1}{c|}{70} & 86 \\
    \hline
    \multicolumn{1}{|c|}{6} & \multicolumn{1}{c|}{44} & \multicolumn{1}{c|}{56} & \multicolumn{1}{c|}{80} & \multicolumn{1}{c|}{104} & 128 \\
    \hline
    \multicolumn{1}{|c|}{8} & \multicolumn{1}{c|}{58} & \multicolumn{1}{c|}{74} & \multicolumn{1}{c|}{106} & \multicolumn{1}{c|}{138} & 170 \\
    \hline
    \end{tabular}
    \renewcommand{\arraystretch}{1}
    \label{tab:pai2_llm_calls_hyperparams}
\end{table}

From Table~\ref{tab:pai2_llm_calls_hyperparams} it can be seen that increasing $S_m$ from 1 to 8 results in a 7.0$\times$ increase in LLM calls, while increasing $C_m$ from 1 to 8 results in a 2.8$\times$ increase. The combined effect at $S_m = 8$ and $C_m = 8$ yields 18.9$\times$ more calls than the minimum configuration ($S_m = 1$, $C_m = 1$).

Table~\ref{tab:pai2_consumed_tokens_in_best_setting} presents the average token consumption per query across PAI-2's three main QA pipeline stages. It can be observed that Memory Graph Exploration dominates token consumption. This stage accounts for approximately 94\% of total token usage across all benchmarks.

\begin{table}[H]
    \centering
    \renewcommand{\arraystretch}{1.3}
    \caption{Average token consumption per query for PAI-2 across QA pipeline stages and benchmarks. Mentioned statistics about Question Preprocessing stage includes tokens for $Decompose$, $Enhance$ and $Denoise$ operations. Mentioned statistics about Memory Graph Exploration stage shows worst-case scenario, then the maximum amount of search steps in a plan ($S_m$) were reached and for every search step the maximum number of clue-queries ($C_m$) were generated: $S_m = 8$ and $C_m = 8$.}
    \begin{tabular}{|c|c|ccccc||c|}
    \hhline{|--|-|-|-|-|-||-|}
    \multicolumn{1}{|c|}{\multirow{2}{*}{\begin{tabular}[c]{@{}c@{}}\textbf{PAI-2} \\ \textbf{Stages}\end{tabular}}} & \multirow{2}{*}{\begin{tabular}[c]{@{}c@{}}\textbf{Tokens}\\ \textbf{Category}\end{tabular}} & \multicolumn{5}{c||}{\textbf{Benchmark}} & \multirow{2}{*}{Mean} \\ 
    \hhline{|~~|-|-|-|-|-||~|}
    \multicolumn{1}{|c|}{} &  & \multicolumn{1}{c|}{\textbf{Natural Questions}} & \multicolumn{1}{c|}{\textbf{TriviaQA}} & \multicolumn{1}{c|}{\textbf{HotpotQA}} & \multicolumn{1}{c|}{\textbf{2WikiMultihopQA}} & \multicolumn{1}{c||}{\textbf{MuSiQue}} &  \\ 
    \hhline{=======::=}
    \multicolumn{1}{|c|}{\multirow{2}{*}{\begin{tabular}[c]{@{}c@{}}Question \\ Preprocessing\end{tabular}}} & prompt & \multicolumn{1}{c|}{2 721} & \multicolumn{1}{c|}{2 722} & \multicolumn{1}{c|}{2 913} & \multicolumn{1}{c|}{3 005} & 2 808 & 2 834 \\ 
    \hhline{|~|-|-|-|-|-|-||-|} 
    \multicolumn{1}{|c|}{} & completion & \multicolumn{1}{c|}{152} & \multicolumn{1}{c|}{183} & \multicolumn{1}{c|}{205} & \multicolumn{1}{c|}{191} & 174 & 181 \\ 
    \hhline{|-|-|-|-|-|-|-||-|}
    \multicolumn{1}{|c|}{\multirow{2}{*}{\begin{tabular}[c]{@{}c@{}}Memory Graph \\ Exploration\end{tabular}}} & prompt & \multicolumn{1}{c|}{156 018} & \multicolumn{1}{c|}{167 084} & \multicolumn{1}{c|}{154 502} & \multicolumn{1}{c|}{153 381} & 153 276 & 156 852 \\ 
    \hhline{|~|-|-|-|-|-|-||-|}
    \multicolumn{1}{|c|}{} & completion & \multicolumn{1}{c|}{5 628} & \multicolumn{1}{c|}{5 753} & \multicolumn{1}{c|}{5 288} & \multicolumn{1}{c|}{4 661} & 4 977 & 5 262 \\ 
    \hhline{|-|-|-|-|-|-|-||-|}
    \multicolumn{1}{|c|}{\multirow{2}{*}{\begin{tabular}[c]{@{}c@{}}Answer \\ Aggregation\end{tabular}}} & prompt & \multicolumn{1}{c|}{889} & \multicolumn{1}{c|}{887} & \multicolumn{1}{c|}{851} & \multicolumn{1}{c|}{860} & 844 & 866 \\ 
    \hhline{|~|-|-|-|-|-|-||-|}
    \multicolumn{1}{|c|}{} & completion & \multicolumn{1}{c|}{103} & \multicolumn{1}{c|}{82} & \multicolumn{1}{c|}{69} & \multicolumn{1}{c|}{81} & 64 & 80 \\ 
    \hhline{=======::=}
    \multicolumn{2}{|c|}{Sum} & \multicolumn{1}{c|}{165 510} & \multicolumn{1}{c|}{176 711} & \multicolumn{1}{c|}{163 827} & \multicolumn{1}{c|}{162 179} & 162 144 & 166 075 \\ 
    \hhline{|--|-|-|-|-|-||-|}
    \end{tabular}
    \renewcommand{\arraystretch}{1}
	\label{tab:pai2_consumed_tokens_in_best_setting}
\end{table}

The Memory Graph Exploration stage includes two critical hyperparameters that directly influence token consumption: $S_m$ and $C_m$. Table~\ref{tab:pai2_consumed_tokens_in_ablation_settings} quantifies the token consumption across different configurations of these hyperparameters.

\begin{table}[H]
    \centering
    \renewcommand{\arraystretch}{1.3}
    \caption{Average token consumption across hyperparameters configuration for PAI-2's Memory Graph Exploration Stage. Values indicate total tokens (prompt + completion) consumed for each combination of $S_m$ and $C_m$. Values are averaged across five benchmarks.}
    \begin{tabular}{|cccccc|}
    \hline
    \multicolumn{6}{|c|}{\textbf{Hyperparameters of PAI-2's Memory Graph Exploration Stage}} \\ 
    \hline
    \multicolumn{1}{|c|}{\multirow{2}{*}{\begin{tabular}[c]{@{}c@{}}\textbf{Max Search} \\ \textbf{Steps ($S_m$)}\end{tabular}}} & \multicolumn{5}{c|}{\textbf{Max Clue Queries ($C_m$)}} \\ 
    \cline{2-6}
    \multicolumn{1}{|c|}{} & \multicolumn{1}{c|}{\textbf{1}} & \multicolumn{1}{c|}{\textbf{2}} & \multicolumn{1}{c|}{\textbf{4}} & \multicolumn{1}{c|}{\textbf{6}} & \textbf{8} \\ 
    \hhline{======}
    \multicolumn{1}{|c|}{1} & \multicolumn{1}{c|}{7 984} & \multicolumn{1}{c|}{9 972} & \multicolumn{1}{c|}{13 947} & \multicolumn{1}{c|}{17 923} & 21 898 \\ 
    \hline
    \multicolumn{1}{|c|}{2} & \multicolumn{1}{c|}{14 101} & \multicolumn{1}{c|}{18 077} & \multicolumn{1}{c|}{26 027} & \multicolumn{1}{c|}{33 978} & 41 929 \\ 
    \hline
    \multicolumn{1}{|c|}{4} & \multicolumn{1}{c|}{26 335} & \multicolumn{1}{c|}{34 286} & \multicolumn{1}{c|}{50 187} & \multicolumn{1}{c|}{66 089} & 81 990 \\ 
    \hline
    \multicolumn{1}{|c|}{6} & \multicolumn{1}{c|}{38 569} & \multicolumn{1}{c|}{50 495} & \multicolumn{1}{c|}{74 348} & \multicolumn{1}{c|}{98 200} & 122 052 \\ 
    \hline
    \multicolumn{1}{|c|}{8} & \multicolumn{1}{c|}{50 803} & \multicolumn{1}{c|}{66 705} & \multicolumn{1}{c|}{98 508} & \multicolumn{1}{c|}{130 311} & 162 114 \\ 
    \hline
    \end{tabular}
    \renewcommand{\arraystretch}{1}
	\label{tab:pai2_consumed_tokens_in_ablation_settings}
\end{table}

From Table~\ref{tab:pai2_consumed_tokens_in_ablation_settings} it can be seen that increasing $S_m$ from 1 to 8 results in a 7.0$\times$ increase in token consumption, while increasing $C_m$ from 1 to 8 results in a 3.0$\times$ increase. The combined effect at $S_m = 8$ and $C_m = 8$ yields 20.3$\times$ more tokens consume than the minimum configuration ($S_m = 1$, $C_m = 1$).

Table~\ref{tab:llmcalls_comparison_evaluation} compares average and maximum LLM calls per query across all methods resulted during RQ1 experiments.

\begin{table*}[ht!]
    \centering
	\renewcommand{\arraystretch}{1.3}
    \caption{Average / maximum LLM calls per query for QA-pipeline inference of LightRAG, RAPTOR, HippoRAG, PAI-1 and PAI-2 across five benchmarks}
    \begin{tabular}{|c|c|c|c|c|c||c|}
    \hhline{------||-}
    \multirow{2}{*}{\textbf{Method}} & \multicolumn{5}{c||}{\textbf{Benchmark}} & \multirow{2}{*}{Mean} \\ 
    \hhline{~-----||~}
     & \textbf{Natural Questions} & \textbf{TriviaQA} & \textbf{HotpotQA} & \textbf{2WikiMultihopQA} & \multicolumn{1}{c||}{\textbf{MuSiQue}} &  \\ 
     \hhline{======::=}
    LightRAG & \multicolumn{5}{c||}{2 / 2} & - \\ 
    \hhline{------||-}
    RAPTOR & \multicolumn{5}{c||}{1 / 1} & - \\ 
    \hhline{------||-}
    HippoRAG 2 & \multicolumn{5}{c||}{2 / 2} & - \\ 
    \hhline{------||-}
    PAI-1 & \multicolumn{5}{c||}{2 / 2} & - \\ 
    \hhline{------||-}
    PAI-2 & 28 / 168 & 32 / 168 & 44 / 168 & 57 / 168 & 51 / 168 & 42 / 168 \\ 
    \hhline{------||-}
    \end{tabular}
    \renewcommand{\arraystretch}{1}
	\label{tab:llmcalls_comparison_evaluation}
\end{table*}

From Table~\ref{tab:llmcalls_comparison_evaluation} it can be observed that baseline methods require minimal calls: RAPTOR (1), LightRAG (2), HippoRAG 2 (2) and PAI-1 (2). In contrast, PAI-2 averages 42 calls per query, with a maximum of 168. This increase results from the iterative search planning mechanism, where each search step generates multiple clue-queries requiring LLM processing. Average calls vary by benchmark complexity: Natural Questions requires fewer iterations (28 calls) than 2WikiMultihopQA (57 calls), reflecting adaptive termination based on answer confidence. Table~\ref{tab:llmtokens_comparison_evaluation} presents average and maximum token consumption per query.

\begin{table*}[ht!]
    \centering
	\renewcommand{\arraystretch}{1.3}
    \caption{Average / maximum tokens per query for QA-pipeline inference of LightRAG, RAPTOR, HippoRAG, PAI-1 and PAI-2 across five benchmarks}
    \begin{tabular}{|c|c|c|c|c|c|c||c|}
    \hhline{-------||-}
    \multirow{2}{*}{\textbf{Method}} & \multirow{2}{*}{\begin{tabular}[c]{@{}c@{}}\textbf{Tokens} \\ \textbf{Category}\end{tabular}} & \multicolumn{5}{c||}{\textbf{Benchmark}} & \multirow{2}{*}{Mean} \\ 
    \hhline{~~-----||~}
     &  & \textbf{Natural Questions} & \textbf{TriviaQA} & \textbf{HotpotQA} & \textbf{2WikiMultihopQA} & \textbf{MuSiQue} &  \\ 
     \hhline{=======::=}
    \multirow{2}{*}{LightRAG} & prompt & 13114 / 18739 & 14752 / 20573 & 12538 / 17502 & 13392 / 17072 & 12771 / 18506 & 13313 / 18478 \\ 
    \hhline{~------||-} 
     & completion & 210 / 1413 & 145 / 715 & 146 / 376 & 190 / 660 & 154 / 473 & 169 / 727 \\ 
     \hhline{-------||-} 
    \multirow{2}{*}{RAPTOR} & prompt & 927 / 1084 & 964 / 1110 & 955 / 1081 & 969 / 1102 & 952 / 1175 & 953 / 1110 \\ 
    \hhline{~------||-} 
     & completion & 113 / 150 & 114 / 150 & 108 / 150 & 113 / 150 & 118 / 150 & 113 / 150 \\ 
     \hhline{-------||-} 
    \multirow{2}{*}{HippoRAG 2} & prompt & 4238 / 4869 & 4709 / 5440 & 4195 / 4720 & 4135 / 4699 & 4249 / 4783 & 4305 / 4902 \\ 
    \hhline{~------||-}
     & completion & 132 / 463 & 119 / 430 & 155 / 409 & 186 / 474 & 154 / 550 & 149 / 465 \\ 
     \hhline{-------||-} 
    \multirow{2}{*}{PAI-1} & prompt & 2443 / 4379 & 2574 / 4478 & 2355 / 4469 & 2345 / 4441 & 2356 / 4486 & 2415 / 4451 \\ 
    \hhline{~------||-}
     & completion & 110 / 530 & 118 / 1150 & 128 / 414 & 129 / 541 & 116 / 408 & 120 / 609 \\ 
     \hhline{-------||-} 
    \multirow{2}{*}{PAI-2} & prompt & 24853 / 154149 & 30485 / 165184 & 38765 / 152577 & 50282 / 151693 & 44780 / 151338 & 37833 / 154988 \\ 
    \hhline{~------||-}
     & completion & 947 / 5530 & 1093 / 5631 & 1416 / 5155 & 1681 / 4555 & 1569 / 4842 & 1341 / 5143 \\ 
     \hhline{-------||-}
    \end{tabular}
    \renewcommand{\arraystretch}{1}
	\label{tab:llmtokens_comparison_evaluation}
\end{table*}

From Table~\ref{tab:llmtokens_comparison_evaluation} it can be seen that PAI-2 consumes substantially more prompt tokens (mean 37833) compared to baselines: LightRAG (13313), HippoRAG 2 (4305), PAI-1 (2415), and RAPTOR (953). This represents a 15.7$\times$ increase over PAI-1, reflecting the cumulative cost of multiple search iterations. Completion tokens also increase (1341 mean for PAI-2 vs. 120 for PAI-1), corresponding to additional LLM outputs generated during iterative search steps.

Token consumption varies across benchmarks for PAI-2, ranging from 25800 tokens (Natural Questions) to 51963 tokens (2WikiMultihopQA). This variation aligns with question complexity, as multi-hop benchmarks require more retrieval iterations and generate longer intermediate contexts. Table~\ref{tab:pai2_consumed_tokens_in_ablation_settings} quantifies the impact of hyperparameters, showing that reducing $S_m$ from 8 to 4 decreases token consumption by approximately 2.0$\times$.

The computational cost analysis reveals distinct trade-offs between efficiency and answer quality:
\begin{itemize}
    \item \textbf{Latency vs. Accuracy.} PAI-2's 5.7$\times$ latency increase over PAI-1 yields an 18 pp improvement in LLM-as-a-Judge scores (Section~\ref{subsec:rq1_comparison}). For applications prioritizing accuracy over response time, this trade-off is favorable.
    \item \textbf{Token Cost vs. Reasoning Capability.} PAI-2's higher token consumption enables complex multi-hop reasoning. On 2WikiMultihopQA and MuSiQue, PAI-2 achieves a 9 pp average gain over all baselines, justifying the additional cost for complex reasoning tasks.
    \item \textbf{Configurable Cost.} Computational cost is configurable via $S_m$ and $C_m$ hyperparameters. Practitioners can reduce costs by limiting search depth, accepting accuracy decrease trade-offs.
    \item \textbf{Benchmark-Dependent Utility.} For simpler benchmarks (Natural Questions, TriviaQA), PAI-2's cost may be unjustified given comparable performance to HippoRAG 2. For complex multi-hop benchmarks, the cost is justified by significant accuracy improvements.
\end{itemize}

Based on these findings, we suggest the following considerations for deploying PAI-2. First, cache LLM inference results for repeated queries and intermediate search steps to reduce redundant computation. Second, for latency-sensitive applications, reduce $S_m$ to 4 and $C_m$ to 4, achieving 3.0$\times$ cost reduction. Third, segment vector databases by element types (objects, theses, episodes) to reduce search space and improve retrieval latency. Fourth, implement early stopping when answer confidence exceeds a threshold, reducing average LLM calls. Finally, deploy PAI-2 primarily for complex multi-hop reasoning tasks where accuracy improvements justify computational costs. For simple factoid queries, lighter baselines (e.g., HippoRAG 2) may be preferable.

These findings demonstrate that while PAI-2 incurs higher computational costs than baselines, the cost is configurable and justified for applications requiring complex reasoning capabilities. The iterative search planning mechanism that drives these costs is also the primary source of PAI-2's accuracy improvements on multi-hop benchmarks.

\clearpage
\twocolumn
\bibliographystyle{unsrt}
\bibliography{references}
\vspace*{-15mm}

\begin{IEEEbiography}[{\includegraphics[width=1in,height=1.25in,clip,keepaspectratio]{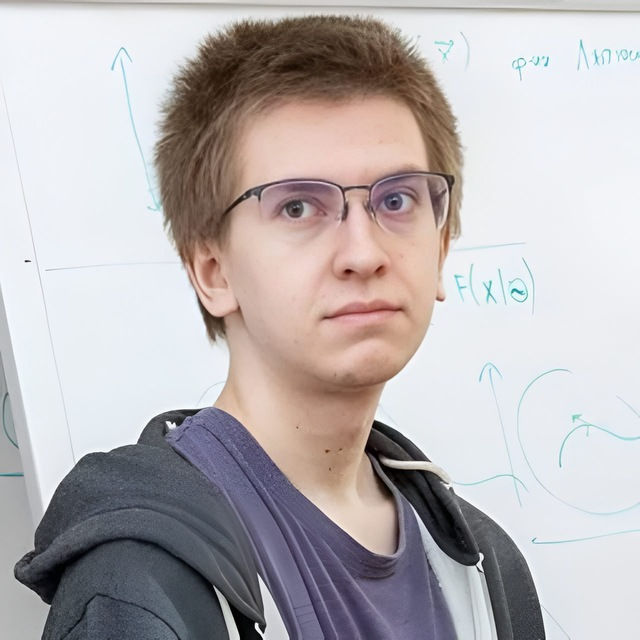}}]{M. Menschikov} received a B.Sc. degree in Software Engineering from Petrozavodsk State University in 2023 and an M.Sc. degree in Machine Learning Engineering from ITMO University in 2025. He is currently a Software Engineer at the Skoltech AI Center, where he has contributed to a projects on developing working memory for LLM agents based on a graph data structures: knowledge graphs. His research interests include generative modeling, GraphRAG, LLM-based knowledge-graph reasoning, LLM-based knowledge-graph construction, multi-agent systems, and dialogue systems.
\end{IEEEbiography}

\vspace*{-15mm}

\begin{IEEEbiography}[{\includegraphics[width=1in,height=1.25in,clip,keepaspectratio]{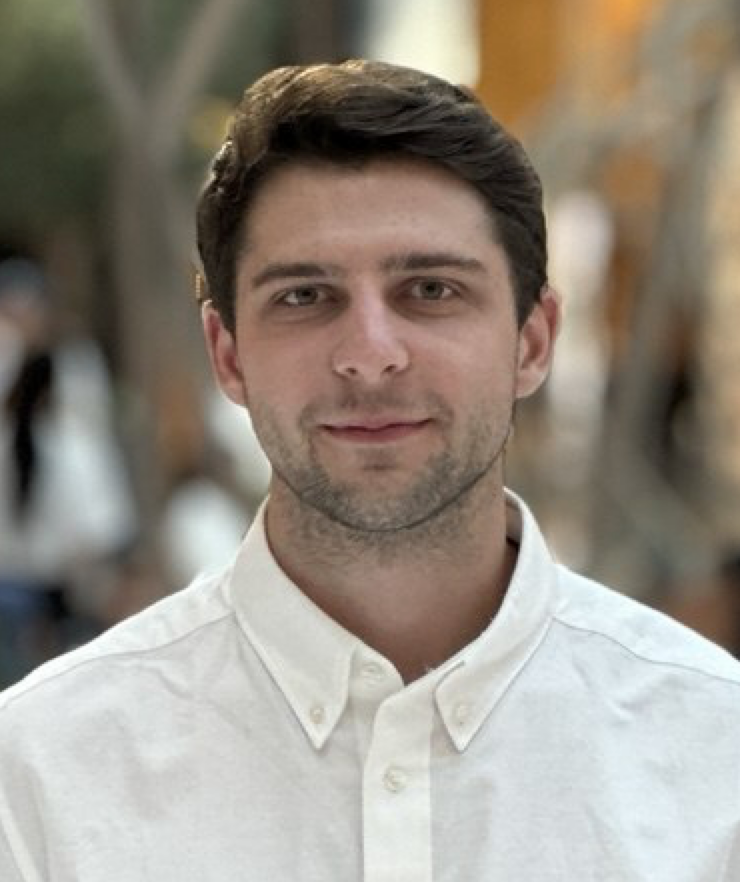}}]{M. Iskornev} received the Specialist degree in mathematics from Lomonosov Moscow State University, Faculty of Mechanics and Mathematics, graduating with honors. He is currently an ML Engineer at the Skoltech AI Center, where he works on methods for building knowledge-graph-based memory for LLM agents and improving contextual learning. He has several years of experience developing production AI systems for NLP, search, and semantic matching. His research interests include LLM-based knowledge graph reasoning and construction, memory and reflection mechanisms for LLM agents, multi-agent systems and long-context compression for in-context learning.
\end{IEEEbiography}

\vspace*{-15mm}

\begin{IEEEbiography}[{\includegraphics[width=1in,height=1.25in,clip,keepaspectratio]{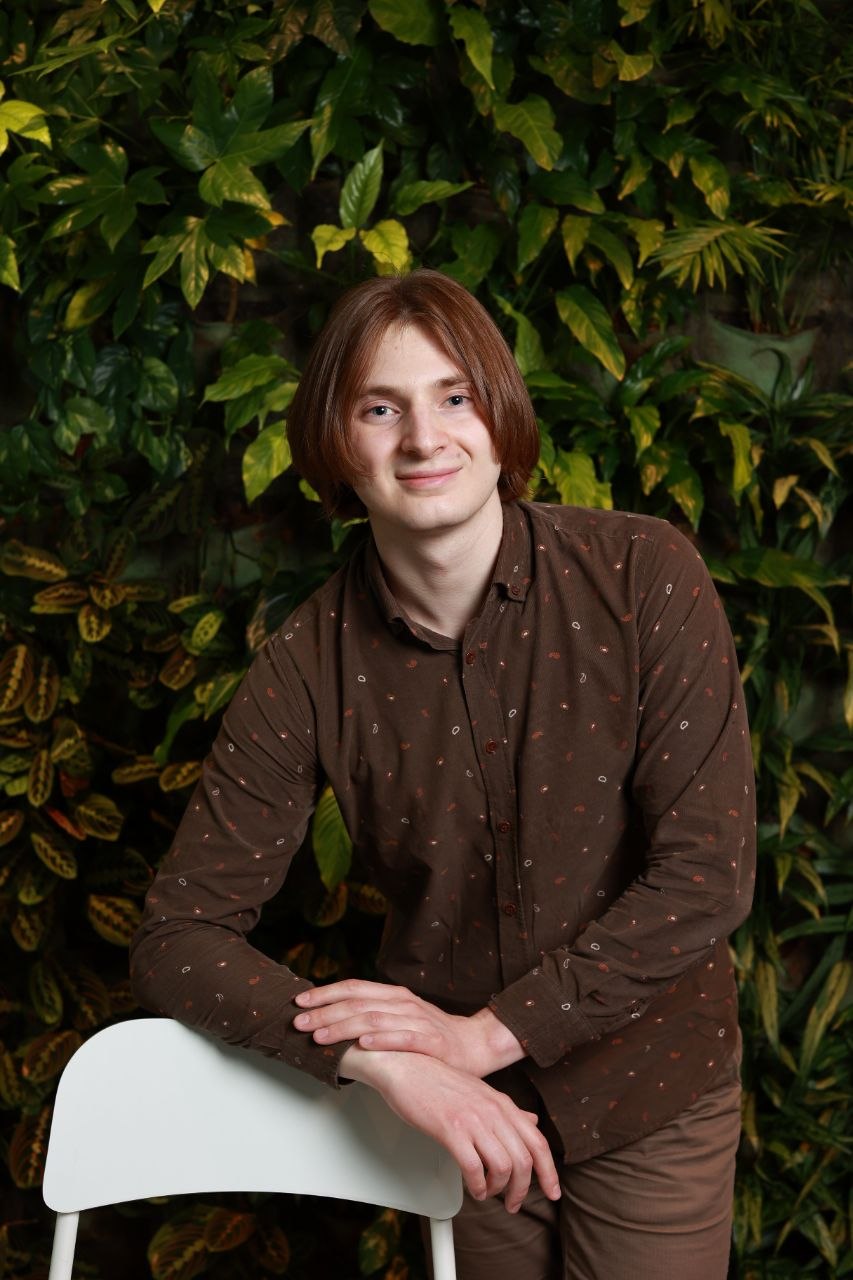}}]{A. Kharitonov} holds a Master’s degree in Data Science from the Skolkovo Institute of Science and Technology. He began his career at Huawei, working pretraining and distillation of Large Language Models. He currently works at SberAI, focusing on the evaluation and validation of AI algorithms. His interests include model evaluation, trustworthy AI, and deploying machine learning benchmarks.
\end{IEEEbiography}

\vspace*{-15mm}

\begin{IEEEbiography}[{\includegraphics[width=1in,height=1.25in,clip,keepaspectratio]{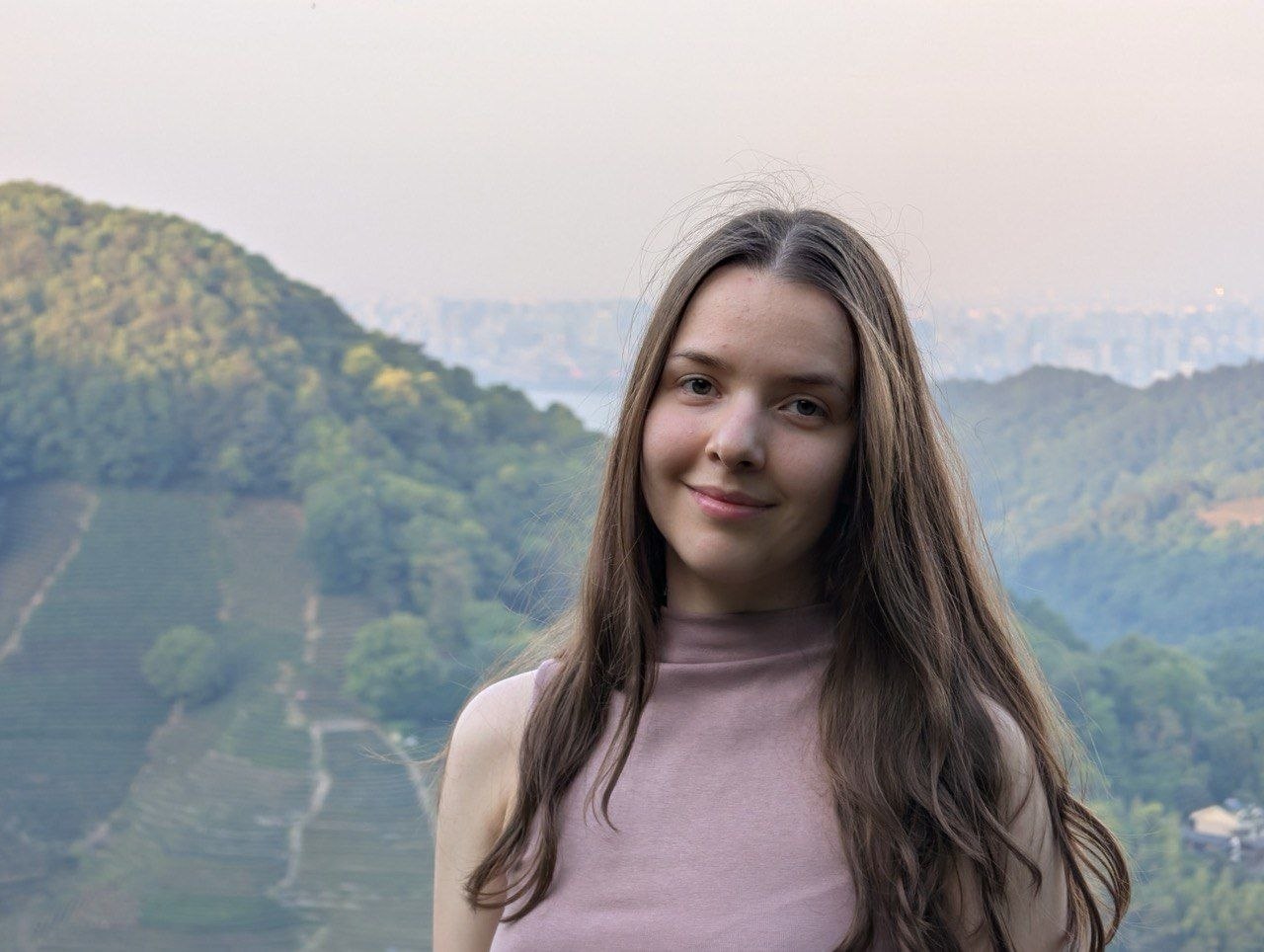}}]{A. Bogdanova} is a researcher specializing in large language model quantization at Huawei. She earned her bachelor’s degree in Mathematics from the Higher School of Economics. She later obtained a master’s degree in Data Science from Skolkovo Institute of Science and Technology (Skoltech). Her current work centers on improving the efficiency and deployment of large language models through quantization techniques, enabling faster and more resource-efficient AI systems. Her research interests include deep learning optimization, model compression, and scalable artificial intelligence systems.
\end{IEEEbiography}

\vspace*{-15mm}

\begin{IEEEbiography}[{\includegraphics[width=1in,height=1.25in,clip,keepaspectratio]{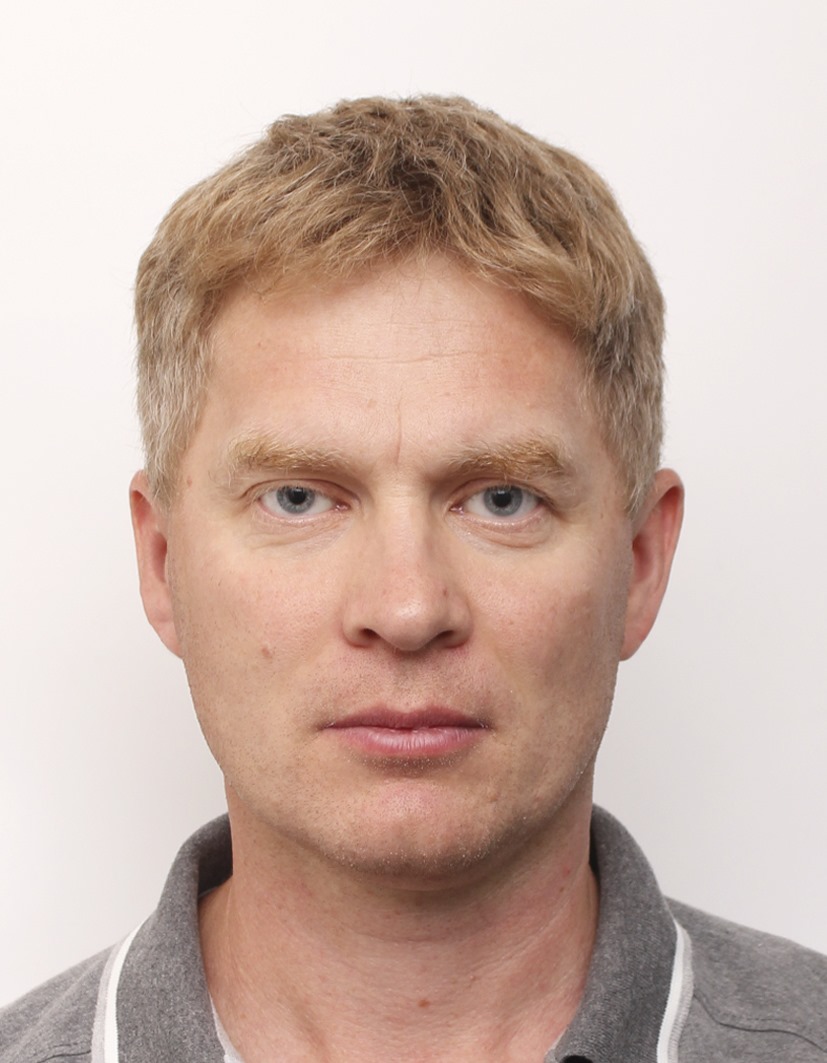}}]{M. Belkin} is a product owner. He has two degrees in technical sciences and economics. Currently, he is a product owner at Sber, where he manages the integration of GenAI into business processes to improve customer service and increase sales conversions. His career in NLP began in 2014 with the development and implementation of models in industrial customer service processes. His experience includes GenAI, NLP, product management, and AI-driven business optimization. Research interests: large language models, NLP, and big data.
\end{IEEEbiography}

\vspace*{-15mm}

\begin{IEEEbiography}[{\includegraphics[width=1in,height=1.25in,clip,keepaspectratio]{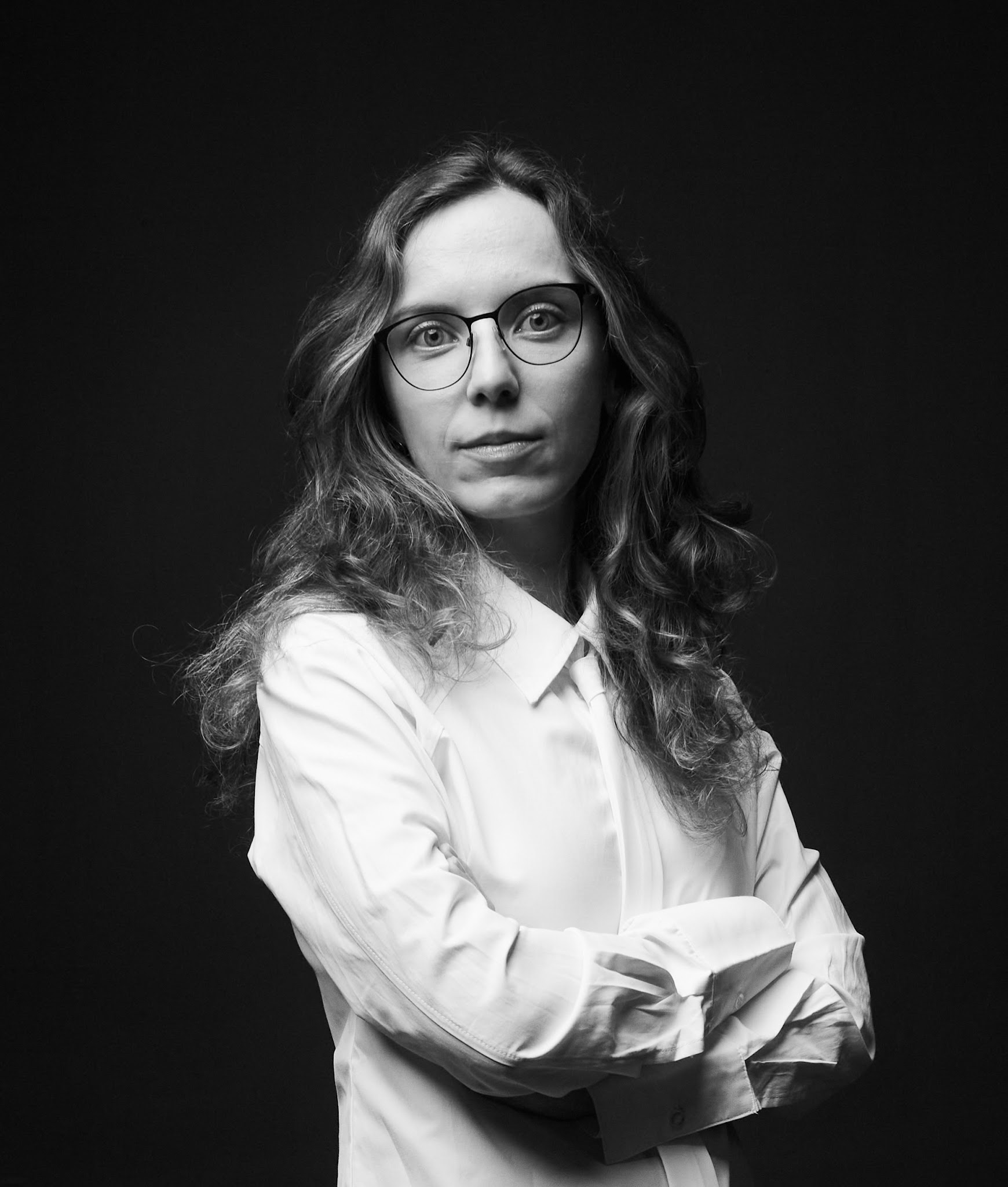}}]{E. Lisitsyna} is a Lead Data Scientist at Sber. Graduated from the Computer Linguistics program at the Higher School of Economics (2020). Currently working on building AI models and agents for speech analytics and knowledge extraction from communications with corporate clients. The created AI solutions help to increase call-to-deal conversion for sales, as well as to automate the quality control of services and support provided by the corporate call center.
\end{IEEEbiography}

\vspace*{-15mm}

\begin{IEEEbiography}[{\includegraphics[width=1in,height=1.25in,clip,keepaspectratio]{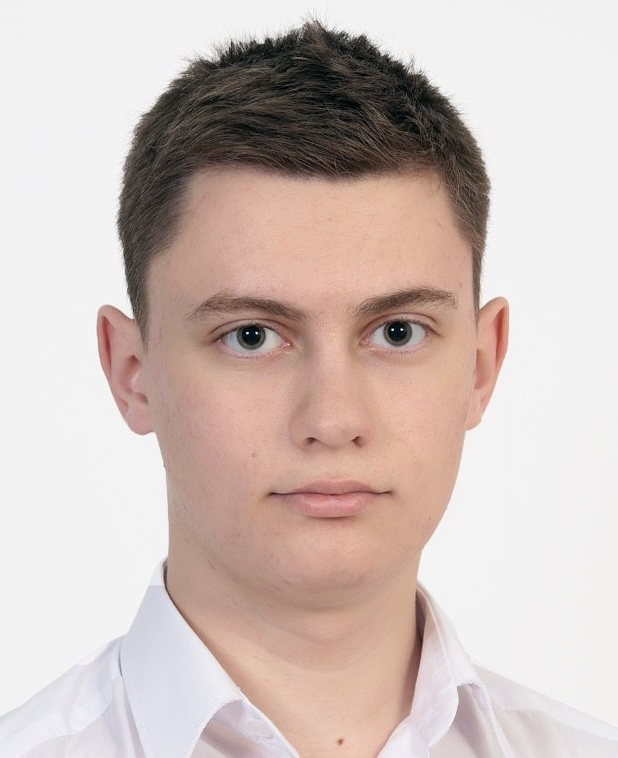}}]{A. Sosedka} received the B.Sc. degree in Applied Mathematics and M.Sc degree in Applied Informatics from National University of Science and Technology MISIS in 2022 and 2024 respectively. He is currently a Senior AI Researcher at Sber, working on building Graph-first Multi-Agent System framework as well as Knowledge Graph memory extensions for LLM Agents. His research interests include GraphRAG, Graph-based Multi Agent Systems, LLM-enhanced Graph Reasoning and Reasoning on Graphs.
\end{IEEEbiography}

\vspace*{-15mm}

\begin{IEEEbiography}[{\includegraphics[width=1in,height=1.25in,clip,keepaspectratio]{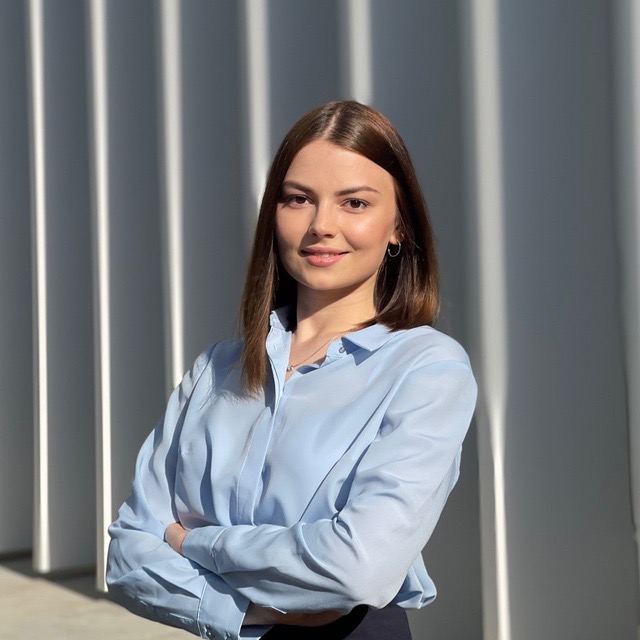}}]{V. Dochkina} is the Director of the AI and Data Center, Strategy and Development Block, Sber. She leads the block’s strategic AI implementation and digital transformation initiatives.
Education: Bachelor’s and Master’s degrees with honors from the Moscow Institute of Physics and Technology (MIPT); Master’s degree from Skoltech, recipient of the 2021 Best Thesis Award; ongoing PhD at MIPT focused on multiagent AI systems and foundation model architectures.
Expertise: Development and deployment of enterprise-scale AI solutions; AI governance frameworks; Agentic AI.
Research interests: Foundation models; multimodal expansion; agentic LLM capability development; scaling AI agents for process automation; autonomous AI systems; mixture-of-experts architectures; coordination frameworks for enterprise-wide autonomization.
\end{IEEEbiography}

\vspace*{-15mm}

\begin{IEEEbiography}[{\includegraphics[width=1in,height=1.25in,clip,keepaspectratio]{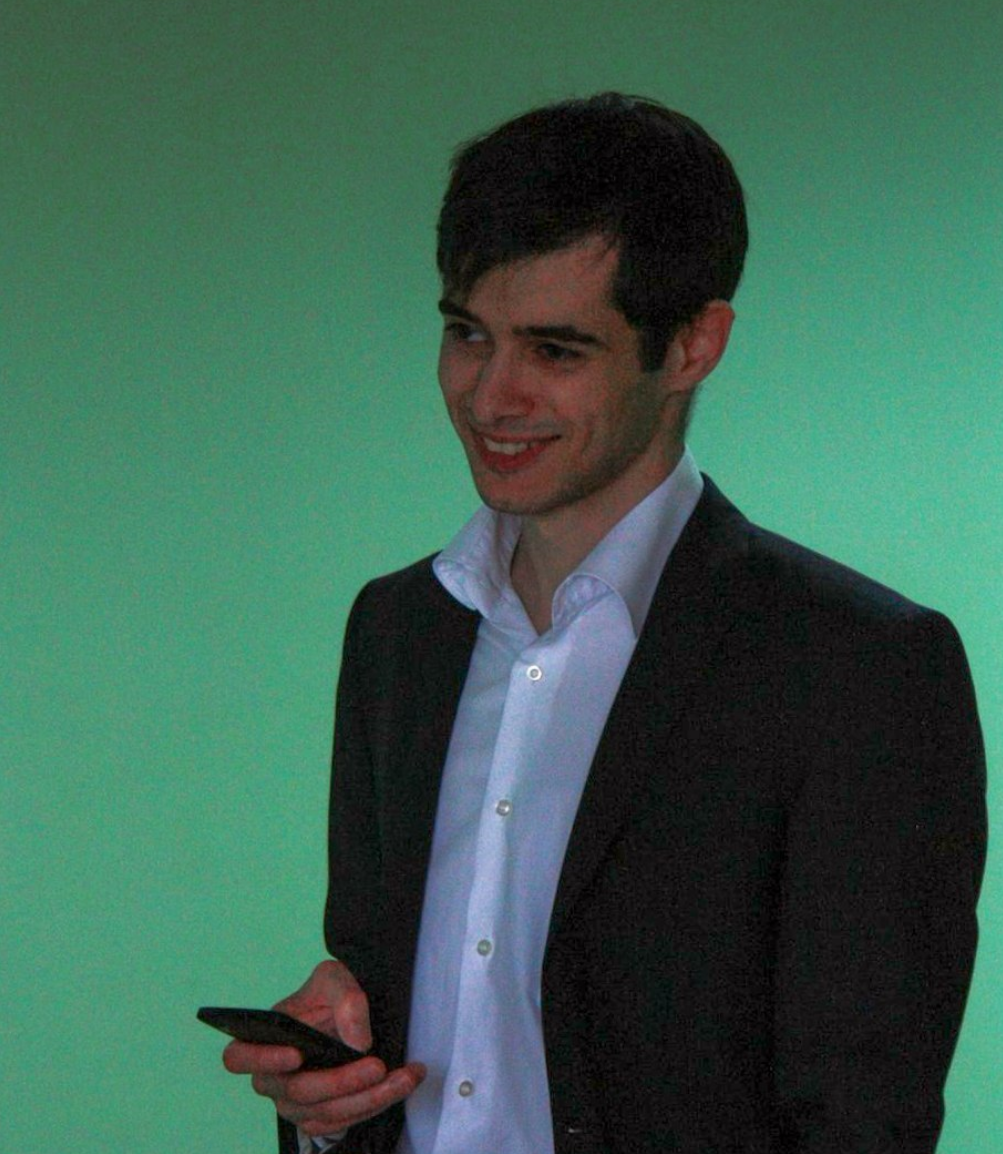}}]{R. Kostoev} got M.Sc. degree in applied mathematics and computer science from Lomonosov Moscow State University, and built an impressive career spanning technology, innovation, and leadership roles. His professional journey includes experience at major companies such as Philips and Google, where he contributed to significant projects and initiatives. 
\end{IEEEbiography}

\vspace*{-15mm}

\begin{IEEEbiography}[{\includegraphics[width=1in,height=1.25in,clip,keepaspectratio]{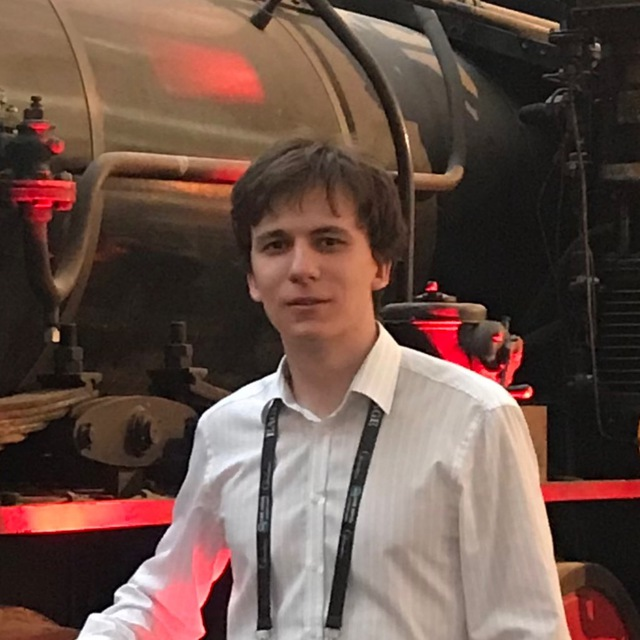}}]{I. Perepechkin} got M.Sc. degree in Applied Mathematics and Physics from Moscow Institute of Physics and Technology in 2017. He has experience developing enterprise-level AI solutions. He is currently a team lead data scientist at Sberbank, developing multi-agent systems.
\end{IEEEbiography}

\vspace*{-15mm}

\begin{IEEEbiography}[{\includegraphics[width=1in,height=1.25in,clip,keepaspectratio]{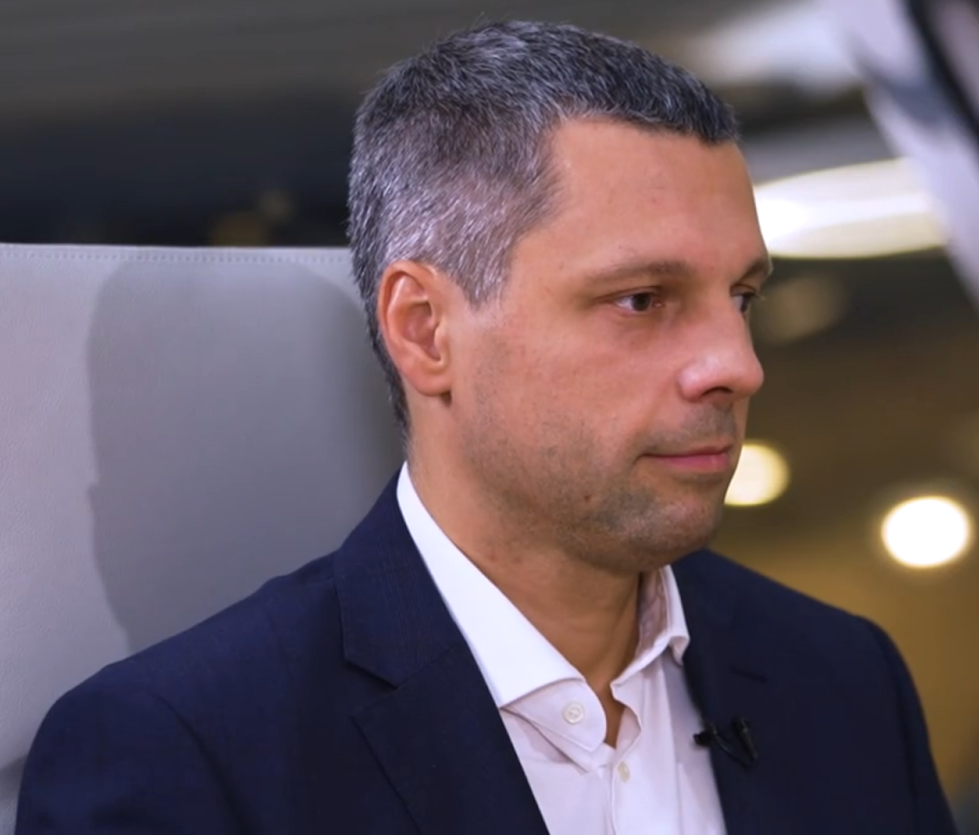}}]{E. Burnaev} received the M.Sc. degree in applied physics and mathematics from Moscow Institute of Physics and Technology, in 2006, the Ph.D. degree in foundations of computer science from the Institute for Information Transmission Problem RAS, in 2008, and the Dr.Sci. degree in mathematical modeling and numerical methods from Moscow Institute of Physics and Technology, in 2022. 
He is currently the Director of the AI Center, Skolkovo Institute of Science and Technology, and a Full Professor. His research interests include generative modeling, manifold learning, deep learning for 3D data analysis, multi-agent systems, and industrial applications.
\end{IEEEbiography}

\EOD

\end{document}